\documentclass[10pt,twocolumn,letterpaper]{article}

\usepackage{cvpr}
\usepackage{color}
\usepackage{times}
\usepackage{epsfig}
\usepackage{graphicx}
\usepackage{amsmath}
\usepackage{amssymb}
\usepackage{mathrsfs}
\usepackage{xcolor}
\usepackage[full,warn]{textcomp}
\usepackage{color}
\usepackage{colortbl}
\usepackage{subfigure}
\usepackage{censor}
\usepackage[sort, nocompress]{cite}

\definecolor{yellow}{rgb}{1,1, 0.6}
\definecolor{orange}{rgb}{1, 0.8, 0.6}
\definecolor{red}{rgb}{1, 0.6, 0.6}
\definecolor{darkred}{rgb}{0.8, 0.0, 0.0}
\definecolor{blue}{rgb}{0, 0, 1.0}
\definecolor{green}{rgb}{0, 1.0, 0}
\definecolor{darkgreen}{rgb}{0, 0.4, 0}
\definecolor{darkorange}{rgb}{1, 0.6, 0}

\newcommand{\contrastone}{I1\,}
\newcommand{\contrastall}{I*\,}
\newcommand{\dualpixelall}{D*\,}
\newcommand{\dualpixelone}{D1\,}

\newcommand{\norm}[1]{\left\lVert#1\right\rVert}

\long\def\ignore#1{}
\def\med(#1){\mathrm{med}(#1)} %

\usepackage[breaklinks=true,bookmarks=false]{hyperref}

\cvprfinalcopy %

\def\cvprPaperID{5896} %

\DeclareMathOperator*{\argmax}{argmax}

\ifcvprfinal\fi
\begin{document}

\title{Learning to Autofocus}

\author{
   {Charles Herrmann}\textsuperscript{\rm 1} \quad
   {Richard Strong Bowen}\textsuperscript{\rm 1}\quad
   {Neal Wadhwa}\textsuperscript{\rm 2}\quad
   {Rahul Garg}\textsuperscript{\rm 2} \\
   {Qiurui He}\textsuperscript{\rm 2} \quad
   {Jonathan T. Barron}\textsuperscript{2} \quad
   {Ramin Zabih}\textsuperscript{1,2} \\
  {\small \textsuperscript{1}Cornell Tech \quad
  \textsuperscript{2}Google Research} \\
  {\tt\small\{cih, rsb, rdz\}@cs.cornell.edu} \quad
  {\tt\small\{nealw, rahulgarg, qiurui, barron, raminz\}@google.com} \\
}
\ignore{
\author{Charles Herrmann\\
Cornell Tech\\
{\tt\small cih@cs.cornell.edu}
\and
Richard Strong Bowen\\
Cornell Tech\\
{\tt\small rsb@cs.cornell.edu}
\and
Neal Wadhwa\\
Google Research\\
{\tt\small nealw@google.com}
\and
Rahul Garg\\
Google Research\\
{\tt\small rahulgarg@google.com}\and
Qiurui He\\
Google Research\\
{\tt\small qiurui@google.com}
\and
Jon Barron\\
Google Research\\
{\tt\small barron@google.com}
\and
Ramin Zabih\\
Cornell Tech, Google Research\\
{\tt\small rdz@cs.cornell.edu, raminz@google.com}

}

\author{Charles Herrmann\inst{1} \and Richard Strong Bowen\inst{1} \and Neal Wadwha\inst{2} \and Rahul Garg\inst{2}\and Charles He\inst{2}\and \\ Jon Barron\inst{2} \and Ramin Zabih\inst{1,2}}
\institute{
    Cornell Tech, New York, NY 10044, USA \and
    Google Research, Mountain View, CA 94043, USA\\
    \email{\{cih,rsb,rdz\}@cs.cornell.edu, \{nealw,rahulgarg,qiurui,barron,raminz\}@google.com}
}
}
\maketitle

\begin{abstract}

Autofocus is an important task for digital cameras, yet current approaches often exhibit poor performance. We propose a learning-based approach to this problem, and provide a realistic dataset of sufficient size for effective learning. Our dataset is labeled with per-pixel depths obtained from multi-view stereo, following \cite{GargICCV2019}.
Using this dataset, we apply modern deep classification models and an ordinal regression loss to obtain an efficient learning-based autofocus technique. 
We demonstrate that our approach provides a significant improvement compared with previous learned and non-learned methods: our model reduces the mean absolute error by a factor of 3.6 over the best comparable baseline algorithm. 
Our dataset and code are publicly available.
\end{abstract}

\section{Introduction}

In a scene with variable depth, any camera lens with a finite-size aperture can only focus at one scene depth (the focus distance), and the rest of the scene will contain blur. This blur is difficult to remove via post-processing, and so selecting an appropriate focus distance is crucial for image quality.

There are two main, independent tasks that a camera must address when focusing. First, the camera must determine the salient region that should be in focus. The user may choose such a region explicitly, e.g., by tapping on the screen of a smartphone, or it may be detected automatically by, for example, a face detector. Second, given a salient region (which camera manufacturers often refer to as ``autofocus points'') and one or more possibly out-of-focus observations, the camera must predict the most suitable focus distance for the lens that brings that particular region into focus. This second task is called \emph{autofocus}.

Conventional autofocus algorithms generally fall into two major categories: contrast-based and phase-based methods. Contrast-based methods define a sharpness metric, and identify the ideal focus distance by maximizing the sharpness metric across a range of focus distances. Such methods are necessarily slow in practice, as they must make a large number of observations, each of which requires physical lens movement. In addition, they suffer from a few important weaknesses, which we discuss in Section~\ref{sec:challenges}.

Modern phase-based methods leverage disparity from the dual-pixel sensors that are increasingly available on smartphones and DSLR cameras. These sensors are essentially two-view plenoptic cameras~\cite{Ng2005} with left and right sub-images that receive light from the two halves of the aperture. These methods operate under the assumption that in-focus objects will produce similar left and right sub-images, whereas out-of-focus objects will produce sub-images with a displacement or disparity that is proportional to the degree of defocus. Naively, one could search for the focus distance that minimizes the left/right mismatch, like the contrast-based methods. Alternatively, some methods use calibration to model the relationship between disparity and depth, and make a prediction with just one input. However, accurate estimation of disparity between the dual-pixel sub-images is challenging due to the small effective baseline. Further, it is difficult to characterize the relationship between disparity and depth accurately due to optical effects that are hard to model, resulting in errors \cite{GargICCV2019}.

In this paper, we introduce a novel learning-based approach to autofocus: a ConvNet that takes as input raw sensor data, optionally including the dual-pixel data, and predicts the ideal focus distance. Deep learning is well-suited to this task, as modern ConvNets are able to utilize subtle defocus clues (such as irregularly-shaped point spread functions) in the data that often mislead heuristic contrast-based autofocus methods. Unlike phase-based methods, a learned model can also directly estimate the position the lens should be moved to, instead of determining it from disparity using a hand-crafted model and calibration---a strategy which may be prone to errors.

In order to train and evaluate our network, we also introduce a large and realistic dataset captured using a smartphone camera and labeled with per-pixel depth computed using multi-view stereo.
The dataset consists of \emph{focal stacks}: a sequence of image patches of the same scene, varying only in focus distance.  We will formulate the autofocus problem precisely in section~\ref{sec:formulation}, but note that the output of autofocus is a focal index which specifies one of the patches in the focal stack.
Both regular and dual-pixel raw image data are included, allowing evaluation of both contrast- and phase-based methods.
Our dataset is larger than most previous efforts \cite{carvalho2018deep, DeepDDF, mir2015autofocus}, and contains a wider range of realistic scenes. Notably, we include outdoors scenes (which are particularly difficult to capture with a depth sensor like Kinect) as well as scenes with different levels of illumination.

We show that our models achieve a significant improvement in accuracy on all versions of the autofocus problem, especially on challenging imagery. On our test set, the best baseline algorithm that takes one frame as input produces a mean absolute error of 11.3 (out of 49 possible focal indices). Our model with the same input has an error of 3.1, and thus reduces the mean absolute error by a factor of 3.6.

\section{Related Work}

There has been surprisingly little work in the computer vision community on autofocus algorithms. There are a number of non-learning techniques in the image processing literature \cite{Chan:ICIP17,Yang:TIP18,Yang:ICIP16,Lee2008,Lee2009}, but the only learning approach  \cite{mir2015autofocus} uses classical instead of deep learning.

A natural way to use computer vision techniques for autofocus would be to first compute metric depth. Within the vast body of literature on depth estimation, the most closely related work of course relies on focus. 

Most monocular depth techniques that use focus take a complete focal stack as input and then estimate depth by scoring each focal slice according to some measure of sharpness \cite{horn1968focusing, Nayar1994, Tenenbaum1971}. Though acquiring a complete focal stack of a static scene with a static camera is onerous, these techniques can be made tractable by accounting for parallax \cite{Suwajanakorn2015}. More recently, deep learning-based methods \cite{DeepDDF} have yielded improved results with a full focal stack approach.

Instead of using a full focal stack, some early work attempted to use the focal cues in just one or two images to estimate depth at each pixel, by relating the apparent blur of the image to its disparity \cite{grossmann1987depth, pentland1987new}, though these techniques are necessarily limited in their accuracy compared to those with access to a complete focal stack. Both energy minimization \cite{Tang_2017_CVPR} and deep learning \cite{carvalho2018deep, Srinivasan2018} have also been applied to single-image approaches for estimating depth from focus, with significantly improved accuracy. Similarly, much progress has been made in the more general problem of using learning for monocular depth estimation using depth cues besides focus~\cite{eigen2014depth, saxena2008make3d}, including dual-pixel cues \cite{WadhwaSIGGRAPH2018,GargICCV2019}.
 
In this work, we address the related problem of autofocus by applying deep learning. A key aspect of the autofocus problem is that commodity focus modules require a single focus estimate to guide them, that may have a tenuous connection with predicted depth map due to hardware issues (see Section~\ref{sec:challenges}). Many algorithms predict non-metric depth maps, making the task harder, e.g., scale invariant monocular depth prediction \cite{eigen2014depth} or affine invariant depth prediction using dual-pixel data \cite{GargICCV2019}. Hence, instead of predicting a dense depth map, we directly predict a single estimate of focal depth that can be used to guide the focus module. This prediction is done end to end with deep learning.

\section{Problem Formulation}
\label{sec:formulation}

\begin{figure}[t]
\centering
\subfigure[Single-Slice]{\includegraphics[width=0.4\linewidth]{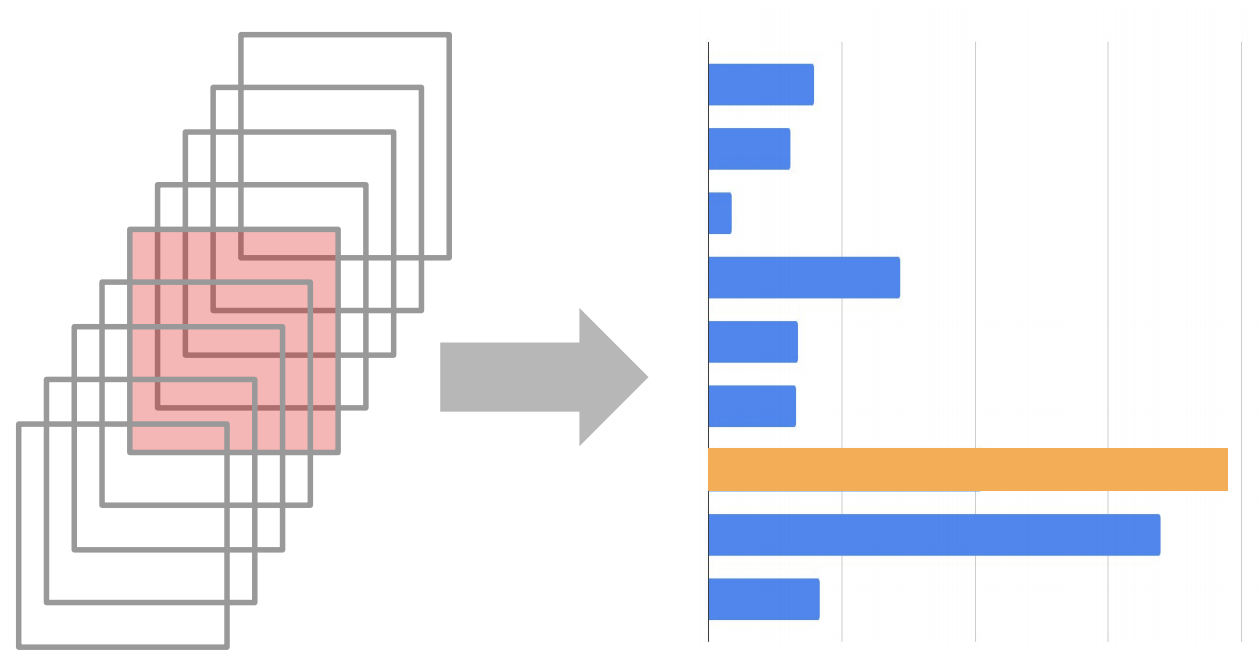}\label{subfig:single}}
\quad\quad
\subfigure[Focal Stack]{\includegraphics[width=0.4\linewidth]{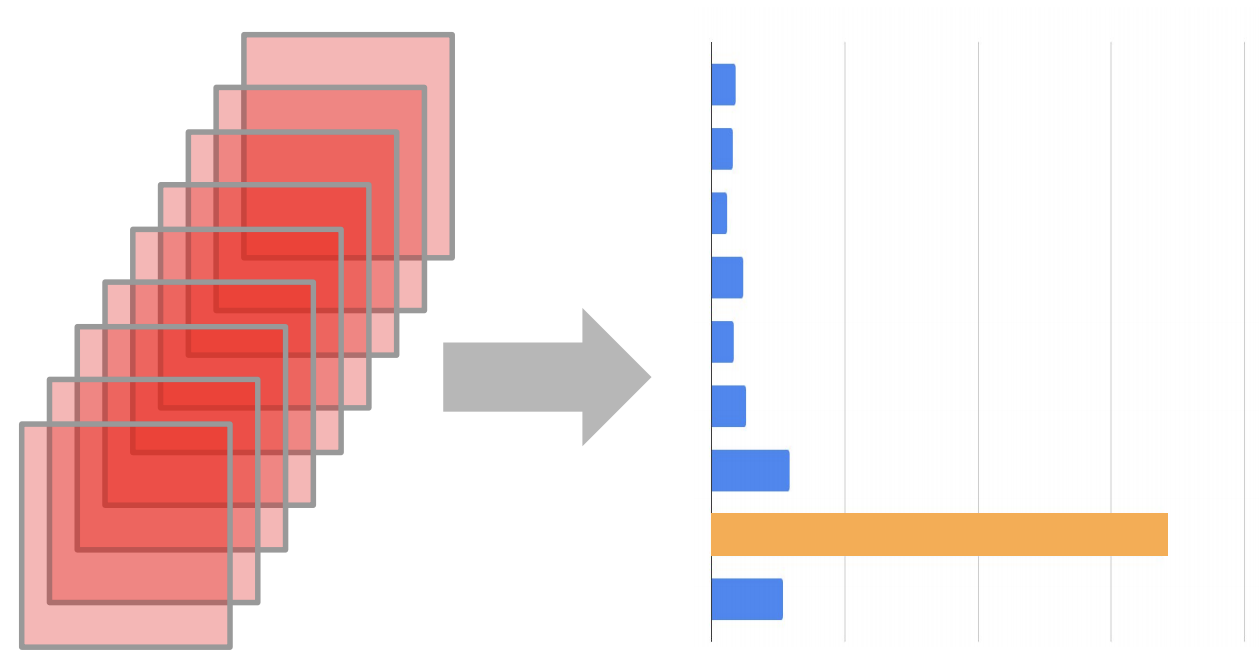}\label{subfig:dense}}
\newline
\subfigure[Two-Step]{\includegraphics[width=0.9\linewidth]{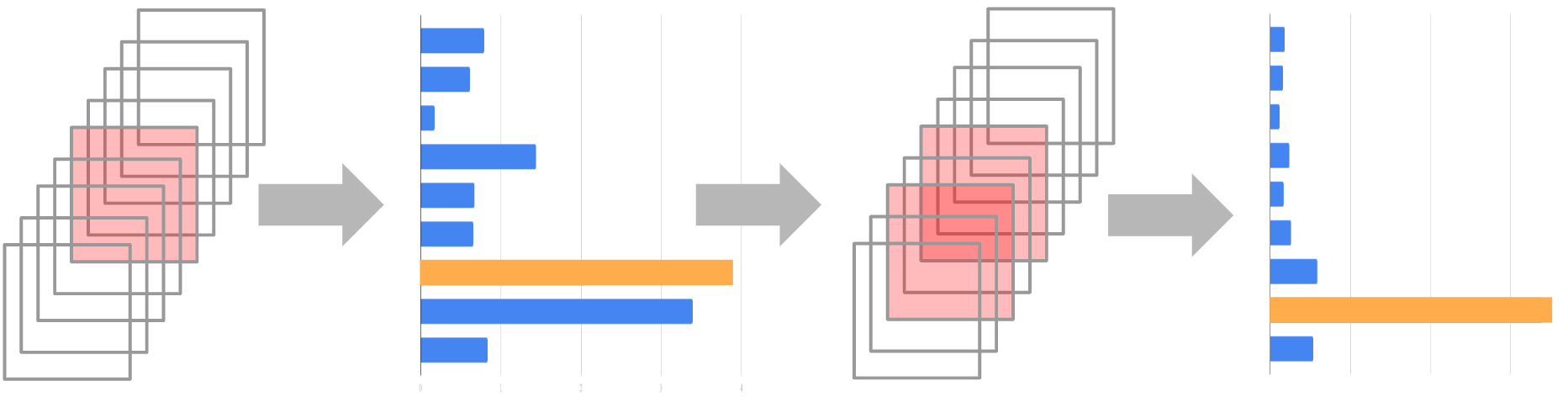}\label{subfig:coarsetofine}}
\caption{Three different autofocus subproblems; in each, the goal is to estimate the in-focus slice, by taking the argmax (orange) of a set of scores produced for each possible focal slice (blue). In the single-slice problem \subref{subfig:single}, the algorithm is given a single observed slice (red). In the focal stack problem \subref{subfig:dense}, the algorithm is given the entire stack. In the multi-step problem (here shown with just two steps) \subref{subfig:coarsetofine}, the problem is solved in stages; Given an initial lens position and image, we decide where to focus next, obtain a new observation, and then make a final estimate of the in-focus slice using both observed images.
\label{fig:af-forms}
}
\end{figure}

In the natural formulation of the autofocus problem, the lens can move continuously, producing an infinite set of possible focus distances corresponding to different focal planes. We discretize the continuous lens positions into $n$ focus distances, and from each position we extract an image patch $I_k, k \in \{ 1, \ldots, n \}$ corresponding to the region of interest. We assume the location of the patch $I_k$ has been determined by a user or some external saliency algorithm, and so we consider this image patch to be ``the image'' and will refer to it as such throughout the paper.
Further, the image can either contain the dual-pixel subimages as two channels or it can contain just the green channel based on the type of input being considered.
We refer to the set of images obtained at different focus distances $\{ I_k \}$ as a \emph{focal stack}, an individual image $I_k$ as a \emph{focal slice}, and $k$ as the \emph{focal index}. We assume each focal stack has exactly one focal index whose slice is in focus. 

Standard autofocus algorithms can be naturally partitioned according to the number of focal slices they require as input. For example, contrast-based methods often require the entire focal stack (or a large subset), whereas phase-based or depth-from-defocus algorithms can estimate a focus distance given just a single focal slice.
Motivated by the differences in input space among standard autofocus algorithms, we define three representative sub-problems (visualized in Figure~\ref{fig:af-forms}), which all try to predict the correct focal index but vary based primarily on their input.
\setlength{\abovedisplayskip}{6pt}
\setlength{\belowdisplayskip}{6pt}

\vspace{-.01cm}
\noindent\textbf{Focal Stack:}
\begin{equation}
f: \{ I_k \mid k = 1, \ldots, n \} \mapsto k^*
\end{equation}
This is the simplest formulation where the algorithm is given a completely observed focal stack. Algorithms for this type typically define a sharpness or contrast metric and pick the focal index which maximizes the chosen metric.

\vspace{-.01cm}
\noindent\textbf{Single Slice:}
\begin{equation}
f: I_k \mapsto k^*, \,\, \forall k \in \{ 1, \ldots, n \}
\end{equation}
This is the most challenging formulation, as the algorithm is given only a single, random focal slice, which can be thought of as the starting position of the lens. In this formulation, algorithms generally try to estimate blur size or use geometric cues to estimate a measure of depth that is then translated to a focal index.

\vspace{-.01cm}
\noindent\textbf{Multi-Step:}
\begin{align}
    f_1: I_{k_0}          & \mapsto k_1  \nonumber \\
    f_2: I_{k_0}, I_{k_1} & \mapsto k_2  \nonumber \\
                        & \ldots       \nonumber \\
    f_m: I_{k_0}, \ldots, I_{k_{m-1}} & \mapsto k_m
\end{align}
where $k_0 \in \{ 1, \ldots, n \}$, and $m$ is a predetermined constant controlling the total number of steps.
The multi-step problem is a mix between the previous two problems. The algorithm is given an initial focal index, acquires and analyzes the image at that focus distance, and then is permitted to move to an additional focal index of its choice, repeating the process at most $m$ times. This formulation approximates the online problem of moving the lens to the correct position with as few attempts as possible.
This multi-step formulation resembles the ``hybrid'' autofocus algorithms that are often used by camera manufacturers, in which a coarse focus estimate is produced by some phase-based system (or a direct depth sensor if available) which is then refined by a contrast-based solution that uses a constrained and abbreviated focal stack as input.

\section{Autofocus Challenges}
\label{sec:challenges}

\begin{figure}
\centering
\subfigure[Ordinary Sensor]{\includegraphics[width=0.46\linewidth]{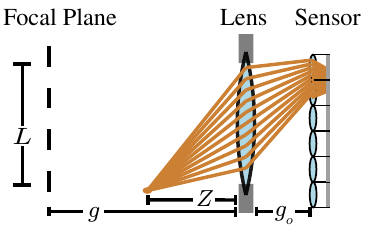}\label{fig:what_is_camera_a}}
\subfigure[Dual-Pixel Sensor]{\includegraphics[width=0.46\linewidth]{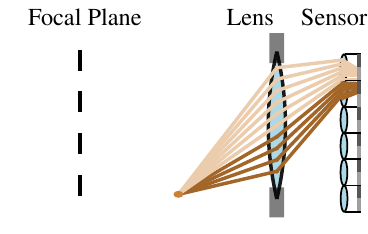}\label{fig:what_is_camera_b}}
\caption{Cameras (a) focus by moving the sensor or lens, and only produce sharp images at a single depth ($g$ in this case). Dual-pixel sensors (b) split each pixel into two halves that each collect light from the two halves of the lens, which aides autofocus.}
\label{fig:what_is_camera}
\end{figure}

We now describe the challenges in real cameras that make the autofocus problem hard in practice.
With the thin-lens and paraxial approximations, the amount of defocus blur is specified by
\begin{equation}\label{eq:blur_size}
\frac{Lf}{1-f/g} \left ( \left | \frac{1}{g} - \frac{1}{Z} \right| \right )
\end{equation}
where $L$ is the aperture size, $f$ the focal length, $Z$ the depth of a scene-point and $g$ the focus distance (Figure.~\ref{fig:what_is_camera_a}). $g$ is related to the distance $g_o$ between the lens and the sensor by the thin-lens equation. This implies that if the depth $Z$ is known, one can focus, i.e, reduce the defocus blur to zero by choosing an appropriate $g$, which can be achieved by physically adjusting the distance between the lens and the sensor $g_o$. This suggests that recovering depth ($Z$) is sufficient to focus. Dual-pixel sensors can aid in the  task of finding $Z$ as they produce two images, each of which sees a slightly different viewpoint of the scene (Figure~\ref{fig:what_is_camera_b}). The disparity $d$ between these viewpoints \cite{GargICCV2019} is 
\begin{equation}\label{eq:disparity}
d = \alpha\frac{Lf}{1-f/g} \left (\frac{1}{g} - \frac{1}{Z} \right )
\end{equation}
where $\alpha$ is a constant of proportionality.

This theoretical model is often used in the academic pursuit of autofocus (or more often, depth-from-defocus) algorithms.
However, the paraxial and thin lens approximations are significant simplifications of camera hardware design and of the physics of image formation. Here we detail some of the issues ignored by this model and existing approaches, and explain how they are of critical importance in the design of an effective, practical autofocus algorithm.

\begin{figure}
    \centering
    \subfigure[Im, $\norm{\nabla}=1.22$]{\includegraphics[width=0.325\linewidth]{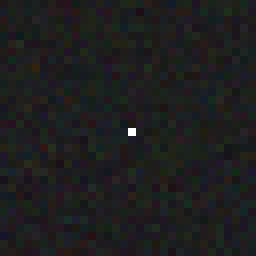}\label{subfig:psf1}}
    \subfigure[Blur, $\norm{\nabla} = 0.62$]{\includegraphics[width=0.325\linewidth]{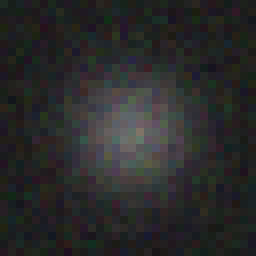}\label{subfig:psf2}}
    \subfigure[Disc, $\norm{\nabla} =2.45$]{\includegraphics[width=0.325\linewidth]{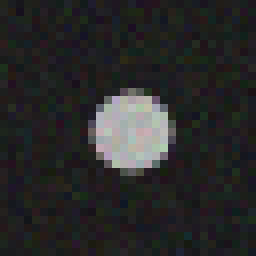}\label{subfig:psf3}}
    \caption{Many contrast-based autofocus algorithms return the focus distance that maximizes image sharpness, measured here as the norm of the image gradient $\norm{\nabla}$. This works well for some camera PSFs, as a sharp image (such as the saturated delta function in \subref{subfig:psf1}) will likely have more gradient energy than the same image seen out of focus under a Gaussian PSF (such as in \subref{subfig:psf2}). But actual cameras tend to have irregular PSFs that more closely resemble discs than Gaussians, and as a result an out-of-focus image may have a \emph{higher} gradient energy than an in-focus image (such as the delta function convolved with a disc filter in \subref{subfig:psf3}). This is one reason why simple contrast-based autofocus algorithms often fail in practice.
    }
    \label{fig:psf}
\end{figure}

\paragraph{Unrealistic PSF Models.} One core assumption underlying contrast-based algorithms is that, as the subject being imaged moves further out of focus, the high-frequency image content corresponding to the subject is reduced. The assumption that in-focus content results in sharp edges while out-of-focus content results in blurry edges has only been shown to be true for Gaussian point spread functions (PSF) \cite{lindeberg1990scale, yuille1986scaling}. However, this assumption can be broken by real-world PSFs, which may be disc- or hexagon-shaped with the goal of producing an aesthetically pleasing ``bokeh''. Or they may be some irregular shape that defies characterization as a side effect of hardware and cost constraints of modern smartphone camera construction. In the case of a disc-shaped PSF, for example, an out-of-focus delta function may actually have \emph{more} gradient energy than an in-focus delta function, especially when pixels are saturated (See Figure~\ref{fig:psf}).

\begin{figure}
    \centering
    \subfigure[Contrast metric]{\includegraphics[width=0.32\linewidth]{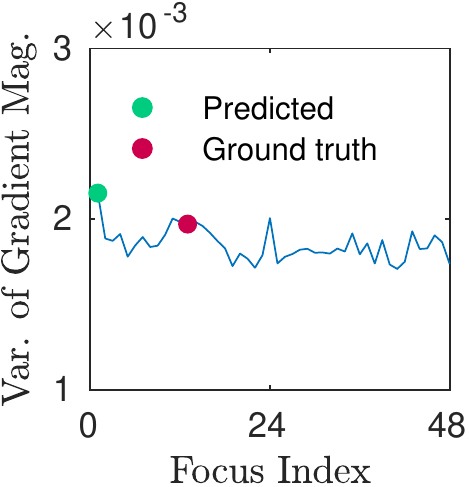}\label{subfig:noise-contrast}}
    \subfigure[Predicted]{\includegraphics[width=0.32\linewidth]{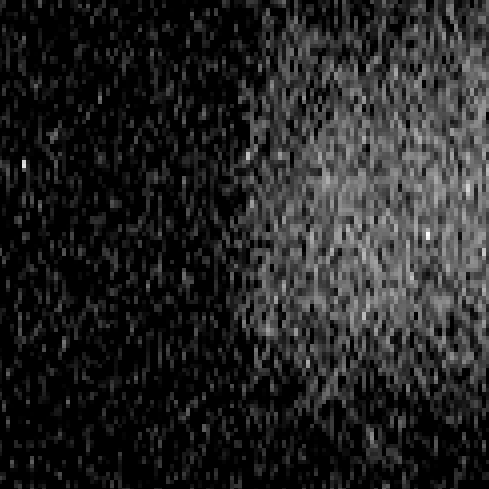}\label{subfig:noise-blurry}}
    \subfigure[Ground truth]{\includegraphics[width=0.32\linewidth]{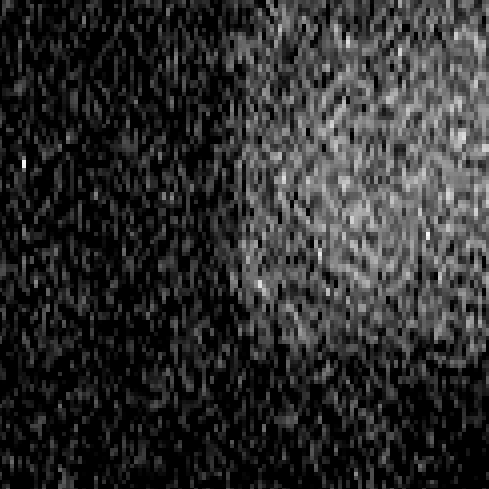}\label{subfig:noise-gt}}
    \caption{Image noise misleads contrast-based focus measures, making it difficult to focus in low-light.
    There is no obvious peak in a contrast measure \subref{subfig:noise-contrast} applied to the noisy patches in \subref{subfig:noise-blurry} and \subref{subfig:noise-gt}.
    As a result, the argmax index results in patch \subref{subfig:noise-blurry} that is out of focus, instead of the in-focus ground-truth patch \subref{subfig:noise-gt}, which contains subtle high-frequency texture.} %
    \label{fig:noise}
\end{figure}

\paragraph{Noise in Low Light Environments.} Images taken in dim environments often contain significant noise, a problem that is exacerbated by the small aperture sizes and small pixel pitch of consumer cameras~\cite{Hasinoff2016}. Prior work in low-light imaging has noted that conventional autofocus algorithms systematically break in such conditions \cite{liba2019handheld}.
This appears to be due to the gradient energy resulting from sensor noise randomly happening to exceed that of the actual structure in the image, which causes contrast-based autofocus algorithms (which seek to maximize contrast) to be misled. See Figure~\ref{fig:noise} for a visualization of this issue.

\begin{figure}
    \centering
    \subfigure[Optics]{\includegraphics[width=0.27\columnwidth]{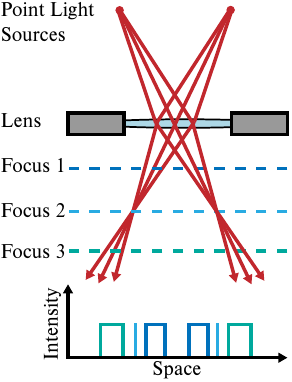}\label{subfig:focal_breathing_optics}}
    \subfigure[Focused, $\norm{\nabla}$\textdblhyphen 0.88]{\includegraphics[width=0.35\columnwidth]{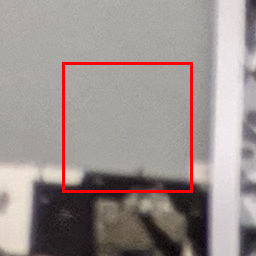}\label{subfig:focal_breathing_focused}}
    \subfigure[Unfocused, $\norm{\nabla}$\textdblhyphen 1.02]{\includegraphics[width=0.35\columnwidth]{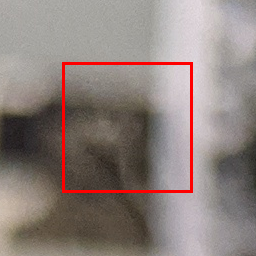}\label{subfig:focal_breathing_unfocused}}
    
    \caption{The optics of image formation mean that modifying the focus of a lens causes ``focal breathing'': a change of the camera's field of view. Consider light from two points that is being imaged at three different focus distances, as in the top of \subref{subfig:focal_breathing_optics}. Because the light is spreading away from the center of the sensor, focusing therefore causes the positions of the points on the imaging plane to shift inwards as the distance between the imaging plane and lens (i.e., focus distance) decreases. 
    This occurs in real image patches and can mislead contrast-based metrics: the in-focus image patch \subref{subfig:focal_breathing_focused} has less gradient energy than the out-of-focus image patch \subref{subfig:focal_breathing_unfocused} because edges move in and out of the patch when focusing. (Gradient energy is only computed within the red rectangles of \subref{subfig:focal_breathing_focused} and \subref{subfig:focal_breathing_unfocused}.)
    }
    
    \label{fig:focal_breathing}
\end{figure}

\paragraph{Focal Breathing.} A camera's field of view depends on its focus distance, a phenomenon called focal breathing.\footnote{Sometimes also referred to as focus breathing or lens breathing.}
This occurs because conventional cameras focus by changing the distance between the image plane and the lens, which induces a zoom-like effect as shown in Figure~\ref{fig:focal_breathing}. 
This effect can be problematic for contrast-based autofocus algorithms, as edges and gradients can leave or enter the field of view of the camera over the course of a focal sweep, even when the camera and scene are stationary. While it is possible to calibrate for focal breathing by modeling it as a zoom and crop, applying such a calibration increases latency, may be inaccurate due to unknown radial distortion, and may introduce resampling artifacts that interfere with contrast-based metrics.

\paragraph{Hardware Support.}
Nearly all smartphone cameras use voice coil motors (VCMs) to focus: the lens sits within a barrel, where it is attached to a coil spring and positioned near an electromagnet, and the electromagnet's voltage is adjusted to move the camera along the 1D axis of the spring and barrel, thereby changing the focus distance of the camera. Though VCMs are inexpensive and ubiquitous, they pose a number of issues for the design of an autofocus or depth-from-defocus algorithm.
1) Most VCM autofocus modules are ``open loop'': a voltage can be specified, but it is not possible to determine the actual metric focus distance that is then induced by this voltage. 
2) Due to variation in temperature, the orientation of the lens relative to gravity, cross talk with other components (e.g., the coils and magnets in optical image stabilization (OIS) module), and simple wear-and-tear on the VCM's spring, the mapping from a specified voltage to its resulting metric focus distance be grossly inaccurate.
3) The lens may move ``off-axis'' (perpendicular to the spring) during autofocus due to OIS, changing both the lens's focus distance and its principal point.

Unknown and uncalibrated PSFs, noise, focal breathing, and the large uncertainty in how the VCM behaves make it difficult to manually engineer a reliable solution to the autofocus problem. This suggests a learning-based approach using a modern neural network.

\section{Dataset}

\begin{figure}
    \centering
    \subfigure[Our capture rig]{\includegraphics[width=0.15\textwidth]{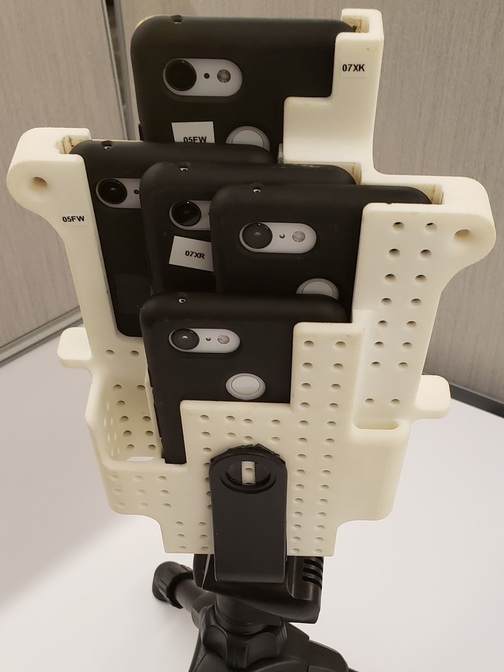}\label{subfig:rig_hardware}}
    \subfigure[RGB]{\includegraphics[width=0.15\textwidth]{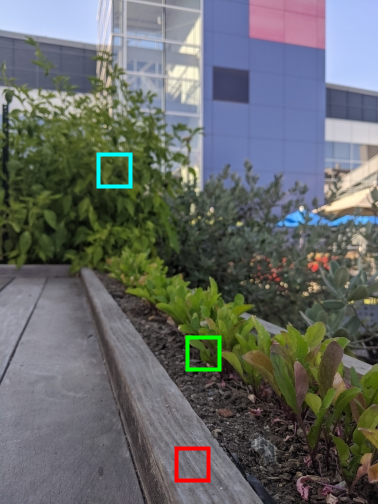}\label{subfig:rig_rgb}}
    \subfigure[Depth]{\includegraphics[width=0.15\textwidth]{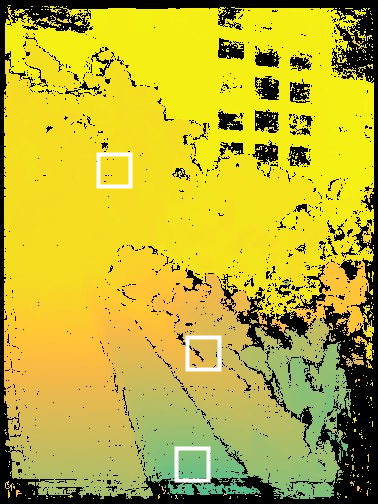}\label{subfig:rig_depth}}
    \subfigure[Example focal stacks]{\includegraphics[width=0.47\textwidth]{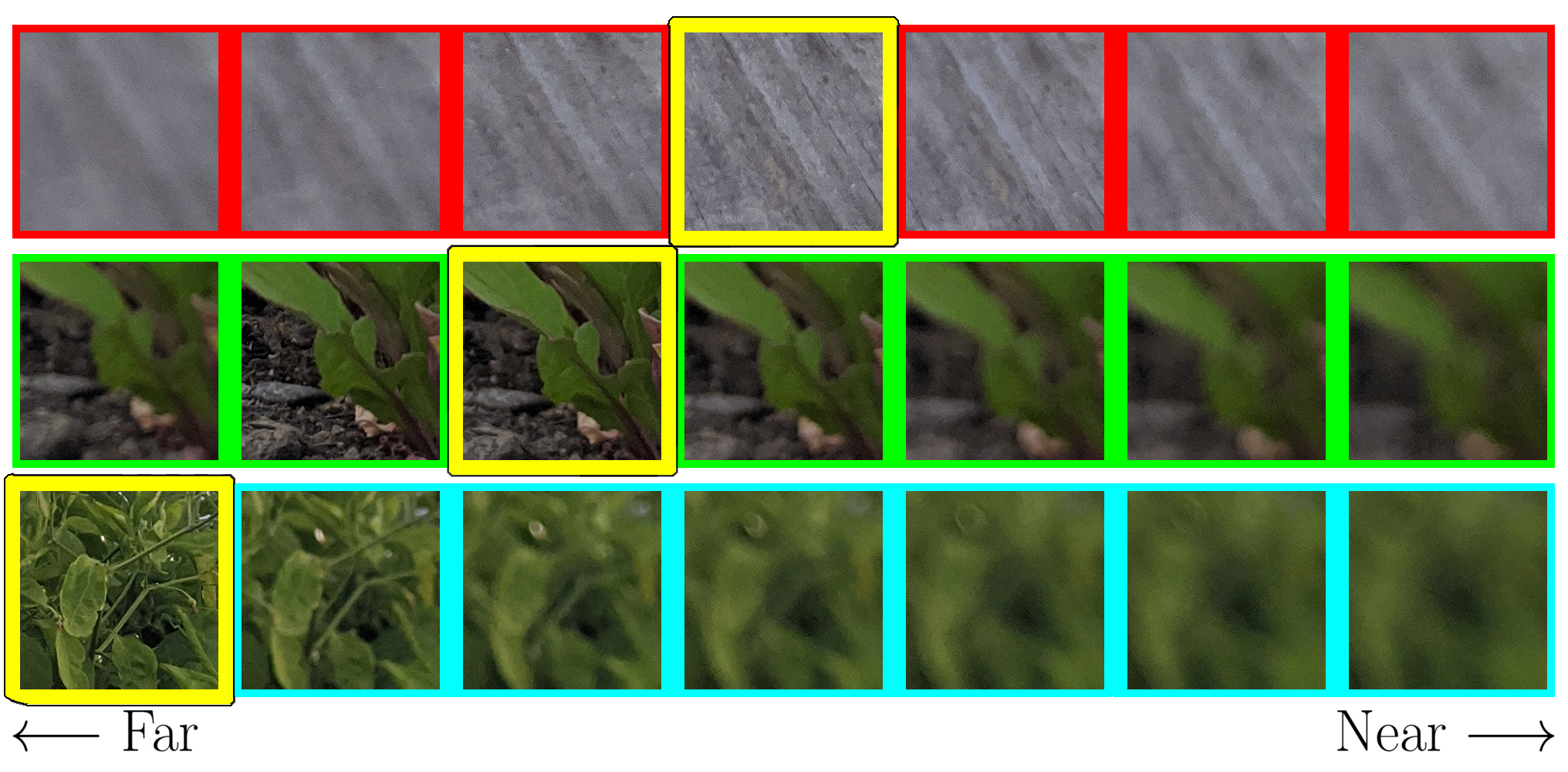}\label{subfig:rig_focal_stacks}} 
    \caption{Our portable rig \subref{subfig:rig_hardware} with 5 synchronized cameras similar to the one in \cite{GargICCV2019} allows us to capture outdoor scenes \subref{subfig:rig_rgb} and compute ground truth depth \subref{subfig:rig_depth} using multi-view stereo. In \subref{subfig:rig_focal_stacks} we show 7 of the 49 slices from three focal stacks at different depths corresponding to the patches marked in \subref{subfig:rig_rgb}. The ground truth patches (the in-focus patches according to our estimated depth) are marked in yellow.}
    \label{fig:rig}
\end{figure}

Our data capture procedure generally follows the approach of \cite{GargICCV2019}, with the main difference being that we capture and process focal stacks instead of individual in-focus captures. Specifically, we use the smartphone camera synchronization system of \cite{ansari2019wireless} to synchronize captures from five Google Pixel 3 devices arranged in a cross pattern (Figure~\ref{subfig:rig_hardware}). We capture a static scene with all five cameras at 49 focal depths sampled uniformly in inverse depth space from 0.102 meters to 3.91 meters. We jointly estimate intrinsics and extrinsics of all cameras using structure from motion \cite{hartley2003multiple}, and then compute depth (Figure~\ref{subfig:rig_depth}) for each image using a modified form of the multi-view stereo pipeline of \cite{GargICCV2019}. We sample $128 \times 128$ patches with a stride of $40$ from the central camera capture yielding focal stacks of dimensions $128 \times 128 \times 49$. We then calculate the ground-truth index for each stack by taking the median of the corresponding stack in the associated depth maps and finding the focal index with the closest focus distance in inverse-depth space. The median is robust to errors in depth and a reasonable proxy for other sources of ground truth that might require more effort, e.g., manual annotation.
We then filter these patches by the median confidence of the depth maps. Please see the supplemental material for more details.

Our dataset has 51 scenes, with 10 stacks per scene containing different compositions, for a total of 443,800 patches. These devices capture both RGB and dual-pixel data. Since autofocus is usually performed on raw sensor data (and not a demosaiced RGB image), we use only the raw dual-pixel data and their sum, which is equivalent to the raw green channel. To generate a train and test set, we randomly selected 5 scenes out of the 51 to be the test set; as such, our train set contains 460 focal stacks (387,000 patches) and our test set contains 50 (56,800 patches).

Our portable capture rig allows us to capture a semantically diverse dataset with focal stacks from both indoor and outdoor scenes using a consumer camera (Figure~\ref{fig:rig}), making the dataset one of the first of its kind.
Compared to other datasets primarily intended for autofocus \cite{carvalho2018deep,mir2015autofocus}, our dataset is substantially larger, a key requirement for deep learning techniques.
Our dataset is comparable in size to \cite{DeepDDF}, which uses a Lytro for lightfield capture and a Kinect for metric depth. However, we have significantly more scenes (51 vs 12) and use a standard phone camera instead of a plenoptic camera.
The latter has a lower resolution ($383\times 552$ for the Lytro used in \cite{DeepDDF} vs $1512 \times 2016$ for our dual-pixel data) and ``focal stacks'' generated by algorithmic refocusing do not exhibit issues such as focal breathing, hardware control, noise, PSFs, etc, which are present upon focusing a standard camera. These issues are some of the core challenges of autofocus, as described above in Section~\ref{sec:challenges}.

\section{Our Model}

We build our model upon the MobileNetV2 architecture~\cite{sandler2018mobilenetv2}, which has been designed to take as input a conventional 3-channel RGB image.
In our use case, we need to represent a complete focal stack, which contains 49 images.
We encode each slice of the focal stack as a separate channel, so the model can reason about each image in the focal stack.
In our experiments where we give the model access to dual pixel data, each image in the focal stack is a 2-channel image where the channels correspond to the left and right dual-pixel images respectively.
In our ablations where the model is deprived of dual-pixel data, each image in the focal stack is a 1-channel image that contains the sum of the left and right views (which is equivalent to the green channel of the raw RGB image).
To accommodate this much ``wider'' number of channels in the input to our network, we increase the number of channels by 4 times the original amount (width multiplier of 4) to prevent a contraction in the number of channels between the input and the first layer. In practice, the network runs quickly: 32.5 ms on a flagship smartphone.

In the setup where the full focal stack is available as input, the model is given a $128 \times 128 \times 98$ tensor for dual-pixel data, and a $128 \times 128 \times 49$ tensor for traditional green-channel sensor data.
In the task where only one focal slice is observable, we use one-hot encoding along the channel dimension as input: the input is a 98-channel tensor (or 49 for green-channel only input) where the channels that correspond to unobserved slices in the focal stack are all zeros.
We use this same encoding in the first step of our multi-step model, but we add an additional one-hot encoding for each subsequent step of the model, thereby giving the model access to all previously-observed images in the focal stack. We train this network by taking a completed single-slice network and evaluate it on all possible focal stacks and input indices. We then feed a new network this one-hot encoding, so the new network sees the first input index and the prediction of the single-slice network.

We model autofocus as an ordinal regression problem: we treat each focal index as its own discrete distinct class, but we assume that there is an ordinal relationship between the class labels corresponding to each focal index  (e.g., index 6 is closer to index 7 than it is to index 15).
The output of all versions of our network is 49 logits.
We train our model by minimizing the ordinal regression loss of \cite{diaz2019soft}, which is similar to the cross-entropy used by traditional logistic regression against unordered labels, but where instead of calculating cross-entropy with respect to a Kronecker delta function representing the ground-truth label, that delta function is convolved with a Laplacian distribution. This encourages the model to make predictions that are as close as possible to the ground-truth, while using traditional cross-entropy would incorrectly model any prediction other than the ground-truth (even those immediately adjacent) as being equally costly.

For training, we use Adam \cite{AdamICLR2015} with default parameters (initial lr $=1e-3$, beta1 $= 0.5$, beta2 $= 0.999$), with a batchsize of 128 and for 20k global steps. For the ordinal regression loss, we use L2 cost metric of \cite{diaz2019soft}  with a coefficient of 1.
\section{Results}

\begin{table}[h!]
\centering
\resizebox{\linewidth}{!}{
\Large
\begin{tabular}{c@{\,\,}|l||cccc|cc}
&& \multicolumn{4}{c}{higher is better} & \multicolumn{2}{|c}{lower is better} \\
& Algorithm & $ = 0$ & $\leq 1$ & $\leq 2$ & $\leq 4$ & MAE & RMSE  \\ \hline\hline
\contrastall & DCT Reduced Energy Ratio \cite{Lee2009} & 0.034  & 0.082  & 0.122  & 0.186  & 18.673  & 22.855  \\
\contrastall & Total Variation (L1) \cite{Nanda01,rudin1992} & 0.048  & 0.136  & 0.208  & 0.316  & 15.817  & 21.013  \\
\contrastall & Histogram Entropy \cite{krotkov1988} & 0.087  & 0.230  & 0.326  & 0.432  & 14.013  & 20.223  \\
\contrastall & Modified DCT \cite{Lee2008}         & 0.033  & 0.091  & 0.142  & 0.235  & 15.713  & 20.197  \\
\contrastall & Gradient Count ($t = 3$) \cite{krotkov1988} & 0.109  & 0.312  & 0.453  & 0.612  & 9.543  & 16.448  \\
\contrastall & Gradient Count ($t = 10$) \cite{krotkov1988} & 0.126  & 0.347  & 0.493  & 0.645  & 9.103  & 16.218  \\
\contrastall & DCT Energy Ratio \cite{shen2006}    & 0.110  & 0.286  & 0.410  & 0.554  & 9.556  & 15.286  \\
\contrastall & Eigenvalue Trace \cite{Wee2007}     & 0.116  & 0.303  & 0.434  & 0.580  & 8.827  & 14.594  \\
\contrastall & Intensity Variance \cite{krotkov1988} & 0.116  & 0.303  & 0.434  & 0.580  & 8.825  & 14.593  \\
\contrastall & Intensity Coefficient of Variation  & 0.125  & 0.327  & 0.469  & 0.624  & 8.068  & 13.808  \\
\contrastall & Percentile Range ($p=3$) \cite{santos1997} & 0.110  & 0.293  & 0.422  & 0.570  & 8.404  & 13.761  \\
\contrastall & Percentile Range ($p=1$) \cite{santos1997} & 0.123  & 0.326  & 0.470  & 0.633  & 7.126  & 12.312  \\
\contrastall & Percentile Range ($p=0.3$) \cite{santos1997} & 0.134  & 0.347  & 0.502  & 0.672  & 6.372  & 11.456  \\
\contrastall & Total Variation (L2) \cite{rudin1992} & 0.167  & 0.442  & 0.611  & 0.770  & 5.488  & 11.409  \\
\contrastall & Sum of Modified Laplacian \cite{Nayar1994} & 0.209  & 0.524  & 0.706  & 0.852  & 4.169  & 9.781  \\
\contrastall & Diagonal Laplacian \cite{Thelen2009} & 0.210  & 0.528  & 0.709  & 0.857  & 4.006  & 9.467  \\
\contrastall & Laplacian Energy \cite{Subbarao1993} & 0.208  & 0.520  & 0.701  & 0.852  & 3.917  & 9.062  \\
\contrastall & Laplacian Variance \cite{PechPacheco2000} & 0.195  & 0.496  & 0.672  & 0.832  & 3.795  & 8.239  \\
\contrastall & Mean Local Log-Ratio ($\sigma=1$)   & 0.220\cellcolor{yellow}  & 0.559  & 0.751  & 0.906  & 2.652  & 6.396  \\
\contrastall & Mean Local Ratio ($\sigma=1$) \cite{Helmli2001} & 0.220\cellcolor{yellow} & 0.559  & 0.751  & 0.906  & 2.645  & 6.374  \\
\contrastall & Mean Local Norm-Dist-Sq ($\sigma=1$) & 0.219  & 0.562  & 0.752  & 0.907  & 2.526  & 5.924  \\
\contrastall & Wavelet Sum ($\ell=2$) \cite{Yang2003} & 0.210  & 0.547  & 0.752  & 0.918  & 2.392  & 5.650  \\
\contrastall & Mean Gradient Magnitude \cite{Tenenbaum1971} & 0.210  & 0.545  & 0.747  & 0.915  & 2.359  & 5.284  \\
\contrastall & Wavelet Variance ($\ell=2$) \cite{Yang2003} & 0.198  & 0.522  & 0.731  & 0.906  & 2.398  & 5.105  \\
\contrastall & Gradient Magnitude Variance \cite{PechPacheco2000} & 0.205  & 0.536  & 0.739  & 0.909  & 2.374  & 5.103  \\
\contrastall & Wavelet Variance ($\ell=3$) \cite{Yang2003} & 0.162  & 0.429  & 0.636  & 0.854  & 2.761  & 5.006  \\
\contrastall & Wavelet Ratio ($\ell=3$) \cite{Xie2006} & 0.161  & 0.430  & 0.640  & 0.862  & 2.706  & 4.856  \\
\contrastall & Mean Wavelet Log-Ratio ($\ell=2$)   & 0.208  & 0.544  & 0.753  & 0.927  & 2.191  & 4.843  \\
\contrastall & Mean Local Ratio ($\sigma=2$) \cite{Helmli2001} & 0.221\cellcolor{orange} & 0.570  & 0.772\cellcolor{orange}     & 0.931\cellcolor{orange}     & 2.072  & 4.569  \\
\contrastall & Wavelet Ratio ($\ell=2$) \cite{Xie2006} & 0.199  & 0.527  & 0.734  & 0.911  & 2.265  & 4.559  \\
\contrastall & Mean Local Log-Ratio ($\sigma=2$)   & 0.221\cellcolor{orange}     & 0.571\cellcolor{yellow}  & 0.772\cellcolor{orange}     & 0.931\cellcolor{orange}     & 2.067  & 4.554  \\
\contrastall & Wavelet Sum ($\ell=3$) \cite{Yang2003} & 0.170  & 0.458  & 0.672  & 0.888  & 2.446  & 4.531  \\
\contrastall & Mean Local Norm-Dist-Sq ($\sigma=2$) & 0.221\cellcolor{orange}     & 0.572\cellcolor{orange}     & 0.770\cellcolor{yellow}  & 0.929\cellcolor{yellow}  & 2.056\cellcolor{orange}     & 4.395  \\
\contrastall & Mean Local Ratio ($\sigma=4$) \cite{Helmli2001} & 0.210  & 0.550  & 0.755  & 0.927  & 2.085  & 4.309  \\
\contrastall & Mean Local Log-Ratio ($\sigma=4$)   & 0.211  & 0.551  & 0.755  & 0.927  & 2.083  & 4.305  \\
\contrastall & Mean Wavelet Log-Ratio ($\ell=3$)   & 0.169  & 0.458  & 0.672  & 0.891  & 2.358  & 4.174\cellcolor{yellow}  \\
\contrastall & Mean Local Norm-Dist-Sq ($\sigma=4$) & 0.212  & 0.555  & 0.760  & 0.928  & 2.059\cellcolor{yellow}  & 4.164\cellcolor{orange}     \\

\contrastall & {\bf Our Model} & \bf0.233\cellcolor{red}     & \bf0.600\cellcolor{red}     & \bf0.798\cellcolor{red}     & \bf0.957\cellcolor{red}     & \bf1.600\cellcolor{red}     & \bf2.446\cellcolor{red}     \\
\hline
\hline
\dualpixelall & Normalized SAD \cite{Hannah1974}   & 0.166  & 0.443  & 0.636  & 0.819  & 4.280  & 8.981  \\
\dualpixelall & Ternary Census (L1, $\epsilon=30$) \cite{stein2004efficient} & 0.171  & 0.450  & 0.633  & 0.802  & 4.347  & 8.794  \\
\dualpixelall & Normalized Cross-Correlation \cite{Barnea1972ACO, Hannah1974} & 0.168  & 0.446  & 0.639  & 0.824  & 4.149  & 8.740  \\
\dualpixelall & Rank Transform (L1) \cite{zabih1994} & 0.172  & 0.451  & 0.633  & 0.811  & 4.138  & 8.558  \\
\dualpixelall & Census Transform (Hamming) \cite{zabih1994} & 0.179\cellcolor{orange}  & 0.473\cellcolor{orange}  & 0.663\cellcolor{orange}  & 0.842\cellcolor{orange} & 3.737  & 8.126  \\
\dualpixelall & Ternary Census (L1, $\epsilon=10$) \cite{stein2004efficient} & 0.178\cellcolor{yellow}  & 0.472\cellcolor{yellow}  & 0.664  & 0.841  & 3.645  & 7.804  \\
\dualpixelall & Normalized Envelope (L2) \cite{birchfield1998} & 0.155  & 0.432  & 0.633  & 0.856  & 2.945  & 5.665  \\
\dualpixelall & Normalized Envelope (L1) \cite{birchfield1998} & 0.165  & 0.448  & 0.653  & 0.870  & 2.731\cellcolor{orange}  & 5.218\cellcolor{orange}  \\
\dualpixelall & {\bf Our Model} & \bf0.241\cellcolor{red} & \bf0.606\cellcolor{red} & \bf0.807\cellcolor{red} &\bf 0.955\cellcolor{red} &\bf 1.611\cellcolor{red} &\bf2.674\cellcolor{red}     \\
\hline \hline

\dualpixelone & ZNCC Disparity with Calibration & 0.064 & 0.181 & 0.286 & 0.448 & 8.879 & 12.911 \\
\dualpixelone & SSD Disparity$^\dagger$ \cite{WadhwaSIGGRAPH2018} & 0.097  & 0.262  & 0.393  & 0.547 & 7.537 & 11.374  \\
\dualpixelone & Learned Depth$^\dagger$ \cite{GargICCV2019} & 0.108\cellcolor{yellow} & 0.289\cellcolor{yellow} & 0.428\cellcolor{yellow} & 0.586\cellcolor{yellow} & 7.176\cellcolor{yellow} & 11.351\cellcolor{yellow}    \\

 \dualpixelone & {\bf Our Model} & \bf0.164\cellcolor{red} & \bf0.455\cellcolor{red}	& \bf0.653\cellcolor{red} & \bf0.885\cellcolor{red} & \bf2.235\cellcolor{red} & \bf3.112\cellcolor{red} \\
 \hline
 \contrastone & {\bf Our Model}  & 0.115\cellcolor{orange}  & 0.318\cellcolor{orange}  & 0.597\cellcolor{orange}  & 0.691\cellcolor{orange}  & 4.321\cellcolor{orange}  & 6.737\cellcolor{orange}  \\
\hline
\hline
\end{tabular}
}
\caption{Results of our model and baselines on the test set for four different versions of the autofocus problem. The leftmost column indicates problem type with \contrastall meaning the full focal stack of green-channel images is passed to the algorithm. In \dualpixelall, the full focal stack of dual-pixel data is passed to the algorithm. In \dualpixelone, a randomly chosen dual-pixel focal slice is passed to the algorithm and in \contrastone, a randomly chosen green-channel slice is passed. Results are sorted by RMSE independently for each input type. The top three techniques for each metric are highlighted with single slice techniques clubbed together. A $\dagger$ indicates that the results were computed on patches inside a 1.5x crop of the entire image.} 
\label{table:big_table}
\end{table}

\newcommand{\rulesep}{\unskip\ {\vrule width 1.2pt}\ }

\begin{figure}
    \centering
    
    {\includegraphics[width=0.09\textwidth]{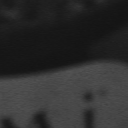}\label{subfig:qual_input_left}
    \includegraphics[width=0.09\textwidth]{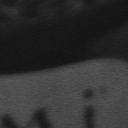}\label{subfig:qual_input_right}}
    {\includegraphics[width=0.09\textwidth]{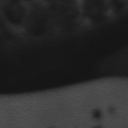}\label{subfig:qual_output_baseline}} 
    {\includegraphics[width=0.09\textwidth]{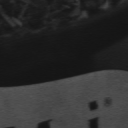}\label{subfig:qual_output}} 
    {\includegraphics[width=0.09\textwidth]{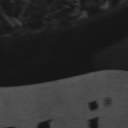}\label{subfig:qual_gt}}     
    
    \subfigure[Dual-pixel input] {\includegraphics[width=0.09\textwidth]{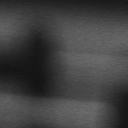}\label{subfig:qual_input_left}
    \includegraphics[width=0.09\textwidth]{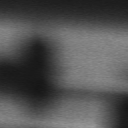}\label{subfig:qual_input_right}}
    \subfigure[Baseline] {\includegraphics[width=0.09\textwidth]{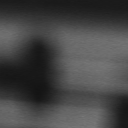}\label{subfig:qual_output_baseline}} 
    \subfigure[Ours] {\includegraphics[width=0.09\textwidth]{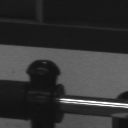}\label{subfig:qual_output}} 
    \subfigure[GT] {\includegraphics[width=0.09\textwidth]{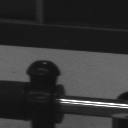}\label{subfig:qual_gt}} 
    \caption{Qualitative results using Learned Depth$^\dagger$ \cite{GargICCV2019} and our \dualpixelone model. Given a defocused dual-pixel patch~\subref{subfig:qual_input_left}, the baseline predicts out-of-focus slices~\subref{subfig:qual_output_baseline}; our model predicts in-focus slices~\subref{subfig:qual_output} that are similar to the ground truth~\subref{subfig:qual_gt}. }
    \label{fig:my_label}
\end{figure}

\begin{figure}
\setlength{\tabcolsep}{1pt}
    \renewcommand{\arraystretch}{1.5}
    \centering
    \resizebox{\linewidth}{!}{
    \begin{tabular}{ccccc}
    \includegraphics[width=0.12\textwidth]{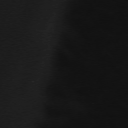} &
    \includegraphics[width=0.12\textwidth]{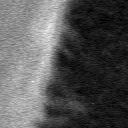} &
    \includegraphics[width=0.12\textwidth]{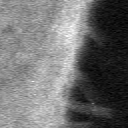}\label{subfig:qual_output_baseline} &
    \includegraphics[width=0.12\textwidth]{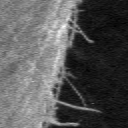}\label{subfig:qual_output} &
    \includegraphics[width=0.12\textwidth]{images/dark1/dark_adjust_modelpredict.png}\label{subfig:qual_gt} \\
    \includegraphics[width=0.12\textwidth]{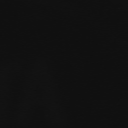} &
    \includegraphics[width=0.12\textwidth]{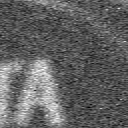} &
\includegraphics[width=0.12\textwidth]{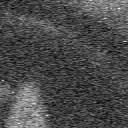}\label{subfig:qual_output_baseline} &
\includegraphics[width=0.12\textwidth]{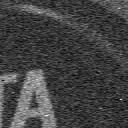}\label{subfig:qual_output} &
\includegraphics[width=0.12\textwidth]{images/dark2/dark_adjust_modelpredict.png}\label{subfig:qual_gt}  \\ 
    (a) Input original & (b) Input  brightened  &  (c) Baseline &  (d) Ours &  (e) GT \\
    \end{tabular}}
    \caption{Qualitative results on low-light examples using ZNCC disparity as baseline and our \dualpixelone model on an example patch for a dark scene. The images have been brightened for visualization.}
    \label{fig:dark_examples}
\end{figure}

\newcommand{\stackwidth}{0.115\textwidth}
\begin{figure}
\centering
\subfigure[Input stack \contrastall] {\includegraphics[width=\stackwidth]{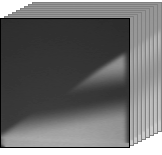}\label{subfig:i1_input}}
\subfigure[Baseline]{\includegraphics[width=\stackwidth]{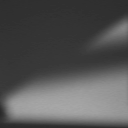}\label{subfig:i1_contrast_output}}
\subfigure[Ours]{\includegraphics[width=\stackwidth]{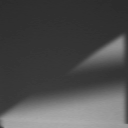}\label{subfig:i1_model_output}}
\subfigure[GT]{\includegraphics[width=\stackwidth]{images/real_breathing_example_model.png}\label{subfig:i1_gt}}
\caption{Qualitative result on an example patch \subref{subfig:i1_input} for \contrastall. All $49$ images are passed as input. The output \subref{subfig:i1_contrast_output} from the \contrastall baseline Mean Local Norm-Dist-Sq ($\sigma=4$)  is out of focus. There is less dark image content in the output due to focal breathing which fools the contrast-based baseline. The output \subref{subfig:i1_model_output} from our \contrastall model is the same as the ground truth \subref{subfig:i1_gt}.}
\end{figure}

We demonstrate that our approach is better than numerous baselines on several variants of the autofocus problem.

We use similar error metrics as the Middlebury stereo dataset~\cite{scharstein2002taxonomy}: the fraction of patches whose predicted focal indices have an error of no more than 0, 1, 2, or 4, as well as the mean absolute error (MAE) and root mean square error (RMSE). For the focal stack problem, all algorithms are run on all elements of the test set and aggregated. For the single-slice problem, an algorithm will be run on $I_k$ for all $k$. For the multi-step problem, each patch in the test set will be evaluated 49 different times, with different focal indices acting as the starting position.

We compare our model's performance against a wide range of baselines.
For the baselines labeled as \contrastall, we take all $49$ images (\ie, the sum of the two dual-pixel images) from the input focal stack, evaluate a sharpness metric for each image, and then take the top-scoring image as the predicted focal depth for the stack. This is basically contrast-based depth-from-defocus. We take the top performing techniques from a recent survey paper~\cite{Pertuz2012}.

The baselines labeled as \dualpixelall use  dual-pixel images as input. Instead of maximizing contrast, they instead attempt to identify which slice in the dual-pixel focal stack has the most similar-looking left and right sub-images, under the assumption that the two sub-images of an in-focus image are identical. Because there is little prior work on dual-pixel autofocus or depth-from-focus using the entire focus stack, we use classical techniques in stereo image-matching to produce a similarity metric between the left and right images that we maximize. 

Finally, the \dualpixelone baselines try to predict the in-focus index given only one dual-pixel image pair. These baselines compute a disparity between the left and right views. As these baselines lack the global knowledge of the entire focal stack, they require calibration mapping this disparity to focus distances in the physical world. This calibration is spatially-varying and typically less accurate in the periphery of the field-of-view \cite{WadhwaSIGGRAPH2018}. Two of the baselines based on prior work only work in the center 1.5x crop of the image. We evaluate these baselines only in the crop region. This only helps those baselines, as issues like focal breathing and irregular PSFs are worse at the periphery. Please see the supplemental material for a description of the baselines.

\subsection{Performance}

Table \ref{table:big_table} presents our model's performance for the full-focal green (\contrastall), full-focal dual pixel (\dualpixelall), single-slice green (\contrastone), and single-slice dual pixel (\dualpixelone) problems. Our \dualpixelone model significantly out-performs other single-slice algorithms, with a RMSE of 3.11 compared to the closest baseline value of 11.351, and MAE of 2.235 compared to 7.176. In other words, baselines were wrong on average by 14.6\% of the focal sweep, whereas our learned model was wrong by only 4.5\%. We also demonstrate improved performance for the full-focal sweep problem, with a MAE of 1.60 compared to 2.06 of Mean Local Norm-Dist. Our \dualpixelall model also outperforms the baselines in its category but performs about the same as our \contrastall model; despite having better within-0, within-1, and within-2 scores, it has slightly lower MAE and MSQE. In a visual comparison, we observed that both of our full-focal models produced patches which were visually very similar to the ground truth and were rarely blatantly incorrect. This suggests that both \contrastall and \dualpixelall have enough information to make an accurate prediction; as such, the additional information in \dualpixelall does not provide a significant advantage.

\subsection{Multi-step}

Table \ref{table:multi-step} presents the results for the multi-step problem. Two \dualpixelone baselines were extended into multi-step algorithms by re-evaluating them on the results of the previous run's output. Both improve substantially from the additional step. In particular, these algorithms are more accurate on indices with less defocus blur (indices close to the ground truth). The first step serves to move the algorithm from a high blur slice to a lower blur slice and the second step then fine-tunes. We see similar behavior from our \contrastone model, which also improves substantially in the second step. We attribute this gain to the model solving the focus-blur ambiguity which we discuss more in Section \ref{sec:ambiguity}. Our \dualpixelone model improves but by a smaller amount than other techniques, likely because it already has high performance in the first step. It also gains much less information from the second slice than the \contrastone model since there is no ambiguity to resolve.

\begin{table}[h!]
\centering
\resizebox{\linewidth}{!}{
\Large
\begin{tabular}{c|l@{\,\,}|c||cccc|cc}
&& \multicolumn{4}{c}{higher is better} & \multicolumn{2}{c}{lower is better} \\
& Algorithm & \# of steps & $ = 0$ & $\leq 1$ & $\leq 2$ & $\leq 4$ & MAE & RMSE  \\ \hline
\dualpixelone & ZNCC Disparity with Calibration & 1 & 0.064  & 0.181  & 0.286  & 0.448  & 8.879  & 12.911  \\ 
& & 2 & 0.100  & 0.278  & 0.426  & 0.617  & 6.662  & 10.993  \\ 
\hline
\dualpixelone & Learned Depth$^\dagger$ \cite{GargICCV2019} & 1 & 0.108  & 0.289  & 0.428  & 0.586  & 7.176  & 11.351  \\
&  & 2 & 0.172\cellcolor{orange}  & 0.433\cellcolor{yellow}  & 0.618\cellcolor{yellow}  & 0.802  & 3.876  & 7.410  \\

\hline
\dualpixelone & Our model & 1 & 0.164\cellcolor{yellow} & 0.455\cellcolor{orange} & 0.653\cellcolor{orange} & 0.885\cellcolor{orange}  & 2.235\cellcolor{orange} \cellcolor{orange} & 3.112 \cellcolor{orange} \\
&  & 2 & \bf 0.201 \cellcolor{red} & \bf 0.519 \cellcolor{red} & \bf0.723 \cellcolor{red}& \bf0.916\cellcolor{red} & \bf1.931\cellcolor{red} & \bf2.772\cellcolor{red} \\

\hline
\contrastone & Our model & 1 & 0.115  & 0.318  & 0.597  & 0.691  & 4.321  & 6.737  \\
&  & 2 & 0.138 & 0.377 & 0.567 & 0.807\cellcolor{yellow} & 2.855\cellcolor{yellow} & 4.088\cellcolor{yellow} \\

\hline
\end{tabular}
}
\caption{
Multi-step problem. Note that the \dualpixelone Learned Depth model uses a 1.5x center crop on the images it evaluates; it evaluates on a subset of the test set which has generally fewer artifacts (eg. focal breathing, radial distortion, etc.).}
\label{table:multi-step}
\end{table}

\subsection{Performance with Registration}

As stated in Section~\ref{sec:challenges}, focal breathing can cause errors in contrast-based techniques. Here, we estimate the magnitude of this problem by registering the focal stack to compensate for focal breathing and then re-evaluating the algorithms on the registered focal stack.

\begin{table}[h!]
\centering
\resizebox{\linewidth}{!}{
\Large
\begin{tabular}{r@{\,\,}|l||cccc|cc}
&& \multicolumn{4}{c}{higher is better} & \multicolumn{2}{|c}{lower is better} \\
 & Algorithm & $ = 0$ & $\leq 1$ & $\leq 2$ & $\leq 4$ & MAE & RMSE  \\ \hline

\contrastall & Mean Local Ratio ($\sigma=2$) \cite{Helmli2001} & 0.222\cellcolor{orange}     & 0.578\cellcolor{yellow}  & 0.776\cellcolor{orange}     & 0.932  & 2.181  & 5.184  \\
\contrastall & Mean Local Log-Ratio ($\sigma=2$)   & 0.222\cellcolor{orange}     & 0.579\cellcolor{orange}     & 0.776\cellcolor{orange}     & 0.932  & 2.176  & 5.178  \\
\contrastall & Mean Local Norm-Dist-Sq ($\sigma=2$) & 0.221\cellcolor{yellow}  & 0.576  & 0.773  & 0.928  & 2.202  & 5.097  \\
\contrastall & Mean Local Ratio ($\sigma=4$) \cite{Helmli2001} & 0.212  & 0.565  & 0.773 & 0.940\cellcolor{yellow}  & 1.923\cellcolor{yellow}  & 3.920  \\
\contrastall & Mean Local Log-Ratio ($\sigma=4$)   & 0.213  & 0.566  & 0.774\cellcolor{yellow}  & 0.941\cellcolor{orange}     & 1.916\cellcolor{orange}  & 3.917\cellcolor{yellow}  \\
\contrastall & Wavelet Sum ($\ell=3$) \cite{Yang2003} & 0.194  & 0.520  & 0.731  & 0.922  & 2.019  & 3.558\cellcolor{yellow}  \\
\contrastall & Mean Wavelet Log-Ratio ($\ell=3$)   & 0.185  & 0.504  & 0.718  & 0.922  & 2.003  & 3.239\cellcolor{orange}  \\

\contrastall & \bf{Our Model} & \bf0.251\cellcolor{red}     & \bf0.610\cellcolor{red}     & \bf0.809\cellcolor{red}     & \bf0.957\cellcolor{red}     & \bf1.570\cellcolor{red}     & \bf2.529\cellcolor{red}     \\
\hline
\end{tabular}%
}
\vspace*{.01in}
\caption{Ablation study with regards to registrations. Existing techniques perform better when the focal stack has undergone a simple registration. However, our model trained on the registered data still performs better than the baselines. }
\label{table:register}
\end{table}

Theoretically, the change in FoV due to focal breathing can be removed using a zoom-and-crop registration calibrated by the camera's focal distance. However, in practice, this registration is far from perfect and can introduce artifacts into the scene. Additionally, any noise in the measurement of focal distance means that a calibration-based registration may be imperfect. To evaluate this approach, we tested two different registrations: a zoom-and-crop registration calibrated by reported focal distance, and a grid search over zoom-and-crop registration parameters to minimize the L2 difference between the images. We note that both of these techniques led to registrations that eliminated some but not all of the change in FOV.

Table~\ref{table:register} shows the performance of a model we trained and the best contrast techniques on the registered data. Most of the contrast algorithms improved when run on the registered focal stack, gaining approximately 0.1 MAE. This suggests that focal breathing affects their performance. In addition, our model trained and evaluated on registered data outperforms our model trained and evaluated on non-registered data.

\subsection{Single-slice Focus-blur Ambiguity}\label{sec:ambiguity}
\begin{figure}\label{fig:ambig}
\subfigure[]{\includegraphics[width=.48\linewidth]{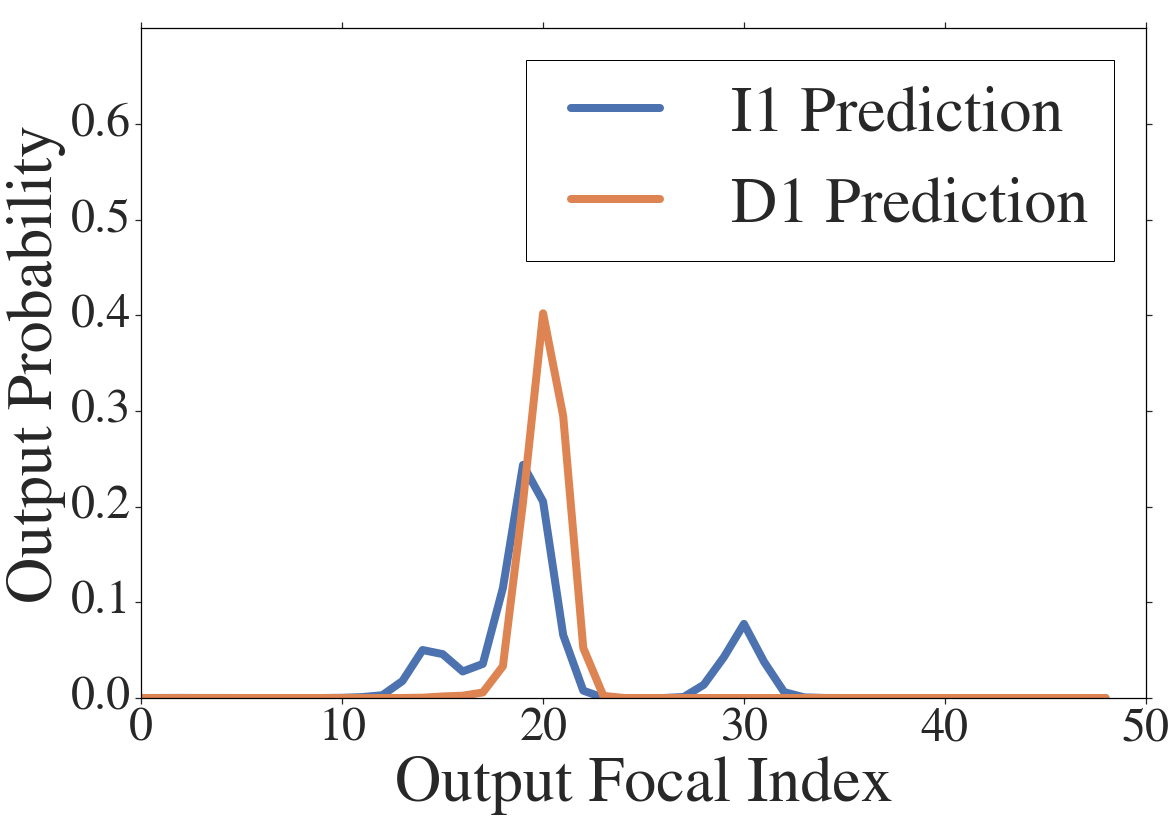}\label{subfig:logits}}
\subfigure[]{\includegraphics[width=.48\linewidth]{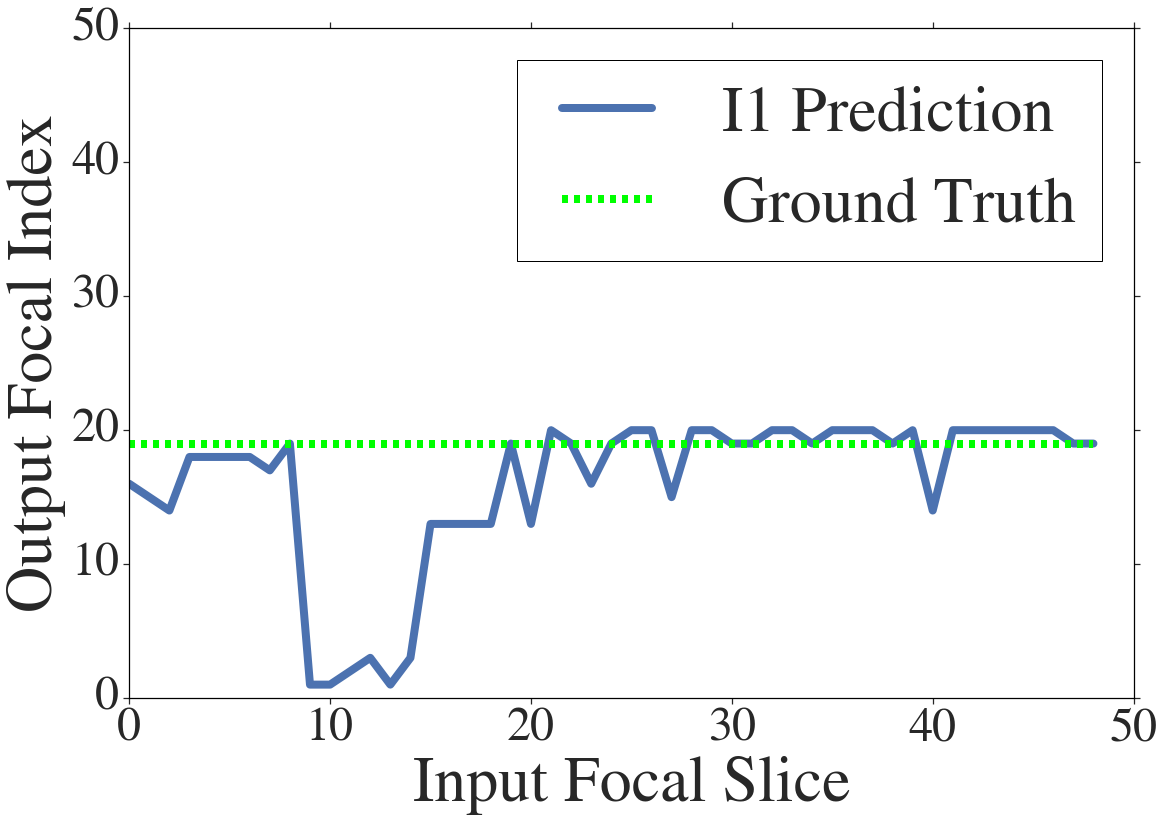}\label{subfig:mirror}}
\caption{\subref{subfig:logits} \contrastone and \dualpixelone model predictions for a patch given focal slice 25 as input. \contrastone model outputs a bimodal distribution as it struggles to disambiguate between the focal indices in-front of and behind the current slice that can generate the same focus-blur. \dualpixelone distribution is unimodal as dual-pixel data helps disambiguate between the two. For the same patch, \contrastone model's prediction for different input slices is visualized in \subref{subfig:mirror}. For focal slices that are towards the near or the far end, the model predicts correctly as one of the two candidate indices lie outside the range while the ambiguity is problematic for input slices in the middle. Inside this problematic range, the model tends to predict focal indices corresponding to depths which, while on the wrong side of the in-focus plane, would produce the same size circle of confusion.}
\end{figure}

In the single-slice problem, an algorithm given only the green-channel faces a fundamental ambiguity: out-of-focus image content may be on either side of the in-focus plane, due to the absolute value in equation~\ref{eq:blur_size}. On the other hand, the model with dual-pixel data can resolve this ambiguity since dual-pixel disparity is signed (Equation~\ref{eq:disparity}). This can be seen from I1 vs D1 results in Table~\ref{table:multi-step} where I1 single step results are significantly worse than single step D1 results, but the difference narrows down for the two step case where the ambiguity can be resolved by looking at two slices.

The ambiguity is also visualized in Figure~\ref{subfig:logits} for a particular patch where the \contrastone model outputs a bimodal distribution while the \dualpixelone model's output probability is unimodal. Interestingly, this ambiguity is only problematic for focal-slices where both the candidate indices are plausible, i.e., lie between 0 and 49, as shown in Figure~\ref{subfig:mirror}.

\clearpage
\newpage
{\small
\bibliographystyle{ieee_fullname}
\bibliography{references}

\begin{thebibliography}{10}\itemsep=-1pt

\bibitem{ansari2019wireless}
Sameer {Ansari}, Neal {Wadhwa}, Rahul {Garg}, and Jiawen {Chen}.
\newblock Wireless software synchronization of multiple distributed cameras.
\newblock {\em ICCP}, 2019.

\bibitem{Barnea1972ACO}
Daniel~I. Barnea and Harvey~F. Silverman.
\newblock A class of algorithms for fast digital image registration.
\newblock {\em Transactions on Computers}, 1972.

\bibitem{birchfield1998}
Stan Birchfield and Carlo Tomasi.
\newblock A pixel dissimilarity measure that is insensitive to image sampling.
\newblock {\em TPAMI}, 1998.

\bibitem{carvalho2018deep}
Marcela Carvalho, Bertrand Le~Saux, Pauline Trouv{\'e}-Peloux, Andr{\'e}s
  Almansa, and Fr{\'e}d{\'e}ric Champagnat.
\newblock Deep depth from defocus: how can defocus blur improve 3d estimation
  using dense neural networks?
\newblock {\em ECCV}, 2018.

\bibitem{Chan:ICIP17}
C. {Chan}, S. {Huang}, and H.~H. {Chen}.
\newblock Enhancement of phase detection for autofocus.
\newblock {\em ICIP}, pages 41--45, Sep. 2017.

\bibitem{shen2006}
{Chun-Hung Shen} and H.~H. {Chen}.
\newblock Robust focus measure for low-contrast images.
\newblock {\em International Conference on Consumer Electronics}, 2006.

\bibitem{cohen1992biorthogonal}
Albert Cohen, Ingrid Daubechies, and J-C Feauveau.
\newblock Biorthogonal bases of compactly supported wavelets.
\newblock {\em Communications on pure and applied mathematics}, 1992.

\bibitem{diaz2019soft}
Raul Diaz and Amit Marathe.
\newblock Soft labels for ordinal regression.
\newblock {\em CVPR}, 2019.

\bibitem{eigen2014depth}
David Eigen, Christian Puhrsch, and Rob Fergus.
\newblock Depth map prediction from a single image using a multi-scale deep
  network.
\newblock {\em NIPS}, 2014.

\bibitem{GargICCV2019}
Rahul Garg, Neal Wadhwa, Sameer Ansari, and Jonathan~T. Barron.
\newblock Learning single camera depth estimation using dual-pixels.
\newblock {\em ICCV}, 2019.

\bibitem{grossmann1987depth}
Paul Grossmann.
\newblock Depth from focus.
\newblock {\em Pattern recognition letters}, 1987.

\bibitem{Hannah1974}
Marsha~Jo Hannah.
\newblock {\em Computer Matching of Areas in Stereo Images.}
\newblock PhD thesis, 1974.

\bibitem{hartley2003multiple}
Richard Hartley and Andrew Zisserman.
\newblock {\em Multiple view geometry in computer vision}.
\newblock Cambridge university press, 2003.

\bibitem{Hasinoff2016}
Samuel~W. Hasinoff, Dillon Sharlet, Ryan Geiss, Andrew Adams, Jonathan~T.
  Barron, Florian Kainz, Jiawen Chen, and Marc Levoy.
\newblock Burst photography for high dynamic range and low-light imaging on
  mobile cameras.
\newblock {\em SIGGRAPH Asia}, 2016.

\bibitem{DeepDDF}
Caner Hazirbas, Sebastian~Georg Soyer, Maximilian~Christian Staab, Laura
  Leal-Taix{\'e}, and Daniel Cremers.
\newblock Deep depth from focus.
\newblock {\em ACCV}, 2018.

\bibitem{Helmli2001}
Franz Helmli and Stefan Scherer.
\newblock Adaptive shape from focus with an error estimation in light
  microscopy.
\newblock {\em International Symposium on Image and Signal Processing and
  Analysis}, 2001.

\bibitem{horn1968focusing}
Berthold~K.P. Horn.
\newblock Focusing.
\newblock 1968.

\bibitem{AdamICLR2015}
Diederick~P Kingma and Jimmy Ba.
\newblock Adam: A method for stochastic optimization.
\newblock {\em ICLR}, 2015.

\bibitem{krotkov1988}
Eric Krotkov.
\newblock Focusing.
\newblock {\em IJCV}, 1988.

\bibitem{Lee2008}
{Sang Yong} Lee, Yogendera Kumar, {Ji Man} Cho, {Sang Won} Lee, and Soo-Won
  Kim.
\newblock Enhanced autofocus algorithm using robust focus measure and fuzzy
  reasoning.
\newblock {\em Transactions on Circuits and Systems for Video Technology},
  2008.

\bibitem{Lee2009}
{Sang Yong} Lee, {Jae Tack} Yoo, and Soo-Won Kim.
\newblock Reduced energy-ratio measure for robust autofocusing in digital
  camera.
\newblock {\em Signal Processing Letters}, 2009.

\bibitem{liba2019handheld}
Orly Liba, Kiran Murthy, Yun-Ta Tsai, Tim Brooks, Tianfan Xue, Nikhil Karnad,
  Qiurui He, Jonathan~T. Barron, Dillon Sharlet, Ryan Geiss, Samuel~W.
  Hasinoff, Yael Pritch, and Marc Levoy.
\newblock Handheld mobile photography in very low light.
\newblock {\em SIGGRAPH Asia}, 2019.

\bibitem{lindeberg1990scale}
Tony Lindeberg.
\newblock Scale-space for discrete signals.
\newblock {\em TPAMI}, 1990.

\bibitem{mir2015autofocus}
Hashim Mir, Peter Xu, Rudi Chen, and Peter van Beek.
\newblock An autofocus heuristic for digital cameras based on supervised
  machine learning.
\newblock {\em Journal of Heuristics}, 2015.

\bibitem{Nanda01}
Harsh Nanda and Ross Cutler.
\newblock Practical calibrations for a real-time digital omnidirectional
  camera.
\newblock Technical report, In Technical Sketches, Computer Vision and Pattern
  Recognition, 2001.

\bibitem{Nayar1994}
Shree~K. Nayar and Yasuo Nakagawa.
\newblock Shape from focus.
\newblock {\em TPAMI}, 1994.

\bibitem{Ng2005}
Ren Ng, Marc Levoy, Mathieu Br{\'e}dif, Gene Duval, Mark~Eden Horowitz, and Pat
  Hanrahan.
\newblock Light field photography with a hand-held plenoptic camera.
\newblock Technical report, Stanford University, 2005.

\bibitem{PechPacheco2000}
Jos{\'e}~Luis Pech-Pacheco, Gabriel Crist{\'o}bal, Jes{\'u}s
  Chamorro-Mart{\'i}nez, and Joaqu{\'i}n Fern{\'a}ndez-Valdivia.
\newblock Diatom autofocusing in brightfield microscopy: a comparative study.
\newblock {\em ICPR}, 2000.

\bibitem{pentland1987new}
Alex~Paul Pentland.
\newblock A new sense for depth of field.
\newblock {\em TPAMI}, 1987.

\bibitem{Pertuz2012}
Said Pertuz, Domenec Puig, and Miguel García.
\newblock Analysis of focus measure operators in shape-from-focus.
\newblock {\em Pattern Recognition}, 2012.

\bibitem{rudin1992}
Leonid~I Rudin, Stanley Osher, and Emad Fatemi.
\newblock Nonlinear total variation based noise removal algorithms.
\newblock {\em Physica D: nonlinear phenomena}, 1992.

\bibitem{sandler2018mobilenetv2}
Mark Sandler, Andrew Howard, Menglong Zhu, Andrey Zhmoginov, and Liang-Chieh
  Chen.
\newblock Mobile{N}et{V}2: Inverted residuals and linear bottlenecks.
\newblock {\em CVPR}, 2018.

\bibitem{santos1997}
Andr{\'e}s~Benavides Santos, Carlos~Ortiz de Solorzano, Juan~J. Vaquero,
  Javier~M{\'a}rquez Pe{\~n}a, Norberto Malpica, and Francisco del Pozo.
\newblock Evaluation of autofocus functions in molecular cytogenetic analysis.
\newblock {\em Journal of microscopy}, 1997.

\bibitem{saxena2008make3d}
Ashutosh Saxena, Min Sun, and Andrew~Y Ng.
\newblock Make3d: Learning 3d scene structure from a single still image.
\newblock {\em TPAMI}, 2008.

\bibitem{scharstein2002taxonomy}
Daniel Scharstein and Richard Szeliski.
\newblock A taxonomy and evaluation of dense two-frame stereo correspondence
  algorithms.
\newblock {\em IJCV}, 2002.

\bibitem{Srinivasan2018}
Pratul~P. Srinivasan, Rahul Garg, Neal Wadhwa, Ren Ng, and Jonathan~T. Barron.
\newblock Aperture supervision for monocular depth estimation.
\newblock {\em CVPR}, 2018.

\bibitem{stein2004efficient}
Fridtjof Stein.
\newblock Efficient computation of optical flow using the census transform.
\newblock {\em Pattern Recognition}, 2004.

\bibitem{Subbarao1993}
Murali Subbarao, Tae-Sun Choi, and Arman Nikzad.
\newblock Focusing techniques.
\newblock {\em Optical Engineering}, 1993.

\bibitem{Suwajanakorn2015}
Supasorn {Suwajanakorn}, Carlos {Hernandez}, and Steven~M. {Seitz}.
\newblock Depth from focus with your mobile phone.
\newblock {\em CVPR}, 2015.

\bibitem{Tang_2017_CVPR}
Huixuan Tang, Scott Cohen, Brian Price, Stephen Schiller, and Kiriakos~N.
  Kutulakos.
\newblock Depth from defocus in the wild.
\newblock In {\em The IEEE Conference on Computer Vision and Pattern
  Recognition (CVPR)}, July 2017.

\bibitem{Tenenbaum1971}
Jay~Martin Tenenbaum.
\newblock {\em Accommodation in Computer Vision}.
\newblock PhD thesis, Stanford University, 1971.

\bibitem{Thelen2009}
Andrea Thelen, Susanne Frey, Sven Hirsch, and Peter Hering.
\newblock Improvements in shape-from-focus for holographic reconstructions with
  regard to focus operators, neighborhood-size, and height value interpolation.
\newblock {\em TIP}, 2009.

\bibitem{WadhwaSIGGRAPH2018}
Neal Wadhwa, Rahul Garg, David~E. Jacobs, Bryan~E. Feldman, Nori Kanazawa,
  Robert Carroll, Yair Movshovitz{-}Attias, Jonathan~T. Barron, Yael Pritch,
  and Marc Levoy.
\newblock Synthetic depth-of-field with a single-camera mobile phone.
\newblock {\em SIGGRAPH}, 2018.

\bibitem{Wee2007}
Chong-Yaw Wee and Raveendran Paramesran.
\newblock Measure of image sharpness using eigenvalues.
\newblock {\em Information Sciences}, 2007.

\bibitem{Xie2006}
Hui Xie, Weibin Rong, and Lining Sun.
\newblock Wavelet-based focus measure and 3-d surface reconstruction method for
  microscopy images.
\newblock {\em IROS}, 2006.

\bibitem{Yang:ICIP16}
C. {Yang} and H.~H. {Chen}.
\newblock Gaussian noise approximation for disparity-based autofocus.
\newblock {\em ICIP}, Sep. 2016.

\bibitem{Yang:TIP18}
C. {Yang}, S. {Huang}, K. {Shih}, and H.~H. {Chen}.
\newblock Analysis of disparity error for stereo autofocus.
\newblock {\em IEEE Transactions on Image Processing}, 2018.

\bibitem{Yang2003}
Ge Yang and B.~J. {Nelson}.
\newblock Wavelet-based autofocusing and unsupervised segmentation of
  microscopic images.
\newblock {\em IROS}, 2003.

\bibitem{yuille1986scaling}
Alan~L Yuille and Tomaso~A Poggio.
\newblock Scaling theorems for zero crossings.
\newblock {\em TPAMI}, 1986.

\bibitem{zabih1994}
Ramin Zabih and John Woodfill.
\newblock Non-parametric local transforms for computing visual correspondence.
\newblock {\em ECCV}, 1994.

\end{thebibliography}
}

\clearpage
\newpage
\appendix

\section{Baseline Algorithms}

Here we document the algorithms taken from prior work that we use as baselines for our proposed model.

\subsection{Contrast-Based Baseline Algorithms}

\newcommand{\contrastscore}{\phi}

As a point of comparison for our proposed model, we implemented a number of contrast-based autofocus algorithms (or equivalently, patch-based depth-from-defocus algorithms) and evaluated them as baselines on our task. When selecting what baselines to implement, we prioritized top-performing techniques according to a relatively recent survey paper \cite{Pertuz2012}. Given a focal stack of images $\{ I \}$ we compute a contrast score $\contrastscore$ for each $I$, and we return the index into the focal stack that maximizes $\contrastscore$.

\paragraph{Intensity Variance \cite{krotkov1988}:} The variance of the intensity values of the entire image.
\begin{equation}
    \contrastscore = \operatorname{Var}(I)
\end{equation}

\paragraph{Intensity Coefficient of Variation \cite{krotkov1988}:} The coefficient of variation of the intensity values of the entire image, which is the standard deviation of the intensity values divided by their mean. Similar metrics are sometimes referred to in past work as ``normalized variance''.
\begin{equation}
    \contrastscore = \frac{\sqrt{\operatorname{Var}(I)}}{\mu(I)}
\end{equation}

\paragraph{Total Variation (L1) \cite{Nanda01,rudin1992}:} The total absolute difference between the intensity value of all pixels and their (4-connected) neighbors:
\begin{equation}
    \contrastscore = \sum_{x, y} \left|I[x, y] - I[x+1, y]\right| + \left|I[x, y] - I[x, y+1]\right|
\end{equation}

\paragraph{Total Variation (L2) \cite{rudin1992}:} The total squared difference between the intensity value of all pixels and their (4-connected) neighbors. This is sometimes referred to as ``gradient energy'':
\begin{equation}
    \contrastscore = \sum_{x, y} \left(I[x, y] - I[x+1, y] \right)^2 + \left( I[x, y] - I[x, y+1] \right)^2
\end{equation}

\paragraph{Energy of Laplacian \cite{Subbarao1993}:} The image is convolved by a discrete Laplace operator, and the response are squared and summed.
\begin{equation}
    \contrastscore = \sum_{x, y} \Delta[x, y]^2, \quad \Delta = I * \begin{bmatrix}0 & \phantom{-}1 & 0\\1 & -4 & 1\\0 & \phantom{-}1 & 0\end{bmatrix}
\end{equation}

\paragraph{Laplacian Variance \cite{PechPacheco2000}:} The image is convolved by a discrete Laplace operator, and the global variance of the response is computed.
\begin{equation}
    \contrastscore = \operatorname{Var}(\Delta[x, y]), \quad \Delta = I * \begin{bmatrix}0 & \phantom{-}1 & 0\\1 & -4 & 1\\0 & \phantom{-}1 &0\end{bmatrix}
\end{equation}

\paragraph{Sum of Modified Laplacian \cite{Nayar1994}:} The image is convolved by a 1D discrete Laplace operator in x and y, and the absolute values of each filter response are summed.
\begin{gather}
    \contrastscore =  \sum_{x, y} \Delta[x, y], \quad \Delta = \left| I * L_x \right| + \left| I * L_y \right| \nonumber  \\
    L_x = \begin{bmatrix} 0 & \phantom{-}0 & 0 \\ 1 & -2 & -1 \\ 0 & \phantom{-}0 & 0 \end{bmatrix}, \quad L_y = L_x^\mathrm{T}
\end{gather}

\paragraph{Diagonal Laplacian \cite{Thelen2009}:} This is the same as the ``sum of modified Laplacian'' approach, but augmented with diagonal Laplacian filters as well.
\begin{gather}
    \contrastscore =  \sum_{x, y} \Delta[x, y] \nonumber \\ 
    \Delta = \left| I * L_x \right| + \left| I * L_y  \right| + \left| I * L_{xy} \right| + \left| I * L_{yx}  \right| \nonumber \\
    L_{xy} = \frac{1}{\sqrt{2}}\begin{bmatrix} 0 & \phantom{-}0 & 1 \\ 0 & -2 & 0 \\ 1 & \phantom{-}0 & 0 \end{bmatrix}, \quad L_{yx} = L_{xy}^\mathrm{T}
\end{gather}

\paragraph{Mean Gradient Magnitude \cite{Tenenbaum1971}:} The mean gradient magnitude, where the gradient is computed using the norm of the response of Sobel filters. This is sometimes referred to as ``Tenengrad''.
\begin{gather}
  \contrastscore = \frac{1}{n} \sum_{x, y} \sqrt{\nabla_x[x, y]^2 + \nabla_y[x, y]^2} \\
  \nabla_x = I * \begin{bmatrix}-1&0&+1\\-2&0&+2\\-1&0&+1\end{bmatrix}, \,\, \nabla_y = I * \begin{bmatrix}-1&-2&-1\\\phantom{-}0 & \phantom{-}0 & \phantom{-}0\\+1&+2&+1\end{bmatrix} \nonumber
\end{gather}

\paragraph{Gradient Count \cite{krotkov1988}:} The total number of edges in the image whose magnitude is above some threshold $t$, where the gradient magnitude is again computed using Sobel filters.
\begin{align}
  \contrastscore = \frac{1}{n} \sum_{x, y} \left[ \left| \nabla_x[x, y] \right| > t \right] + \left[ \left| \nabla_y[x, y] \right| > t \right]
\end{align}

\paragraph{Gradient Magnitude Variance \cite{PechPacheco2000}:} The global variance of gradient magnitudes, where gradients are again computed using Sobel filters.
\begin{align}
  \contrastscore = \operatorname{Var}\left(\sqrt{\nabla_x[x, y]^2 + \nabla_y[x, y]^2} \right)
\end{align}

\paragraph{Percentile Range:} The difference between the $100-p$'th percentile and the $p$'th percentile of intensity values in the image. When $p=0$, this is the difference between the maximum and minimum pixel intensities in the image.
\begin{equation}
    \contrastscore = \operatorname{percentile}(I, 100 - p) - \operatorname{percentile}(I, p)
\end{equation}

\paragraph{Histogram Entropy \cite{krotkov1988}:} The Shannon entropy of all pixel intensities in the image.
\begin{align}
    \contrastscore = -\sum_i \mathbf{n}_i \log(\mathbf{n}_i),\quad \mathbf{n} = \operatorname{hist}(I)
\end{align}

\paragraph{DCT Energy Ratio \cite{shen2006}:} The squared sum of all DCT coefficients of the image without the DC component, divided by the squared DC component.
\begin{equation}
\resizebox{0.95\linewidth}{!}{%
$
    \contrastscore = \frac{\left(\sum_{u, v} D[u, v]^2 \right) - D[0,0]^2}{D[0,0]^2}, \quad D = \operatorname{DCT}(I)
$}
\end{equation}

\paragraph{DCT Reduced Energy Ratio \cite{Lee2009}:} The squared sum of the 5 lowest order DCT coefficients (excluding the DC component) divided by the squared DC component.
\begin{align}
    \contrastscore = \frac{D[0,1]^2 + D[1,0]^2 + D[0,2]^2 + D[1,1]^2 + D[2,0]^2}{D[0,0]^2}
\end{align}

\paragraph{Modified DCT \cite{Lee2008}:} The total filter response of the image convolved with a checkerboard-like filter, which is somewhat related to the DCT of the image.
\begin{align}
    \contrastscore = \sum_{x, y} \left( I * \begin{bmatrix} +1 & +1 & -1 & -1 \\  +1 & +1 & -1 & -1 \\ -1 & -1 & +1 & +1 \\   -1 & -1 & +1 & +1 \end{bmatrix} \right)[x, y]
\end{align}

\paragraph{Wavelet Sum \cite{Yang2003}:} The sum of the absolute value of the high-frequency components of level $\ell$ of the wavelet decomposition of the image. In our experiments, we use CDF9/7 wavelets \cite{cohen1992biorthogonal}.
\begin{align}
    \contrastscore &= \sum_{x, y} \left| W_{LH}^{(\ell)}[x, y] \right| +  \left| W_{HL}^{(\ell)}[x, y] \right| +  \left| W_{HH}^{(\ell)}[x, y] \right| \nonumber \\
    &\left(W_{LL}^{(\ell)}, W_{LH}^{(\ell)}, W_{HL}^{(\ell)}, W_{HH}^{(\ell)} \right) = \operatorname{CDF9/7}(I, \ell)
\end{align}

\paragraph{Wavelet Variance \cite{Yang2003}:} The variance of the high-frequency components of level $\ell$ of the wavelet decomposition of the image.
\begin{equation}
\resizebox{0.95\linewidth}{!}{%
$
    \contrastscore = \operatorname{Var}\left(W_{LH}^{(\ell)}[x, y]\right) + 
    \operatorname{Var}\left(W_{HL}^{(\ell)}[x, y] \right) +
    \operatorname{Var}\left(W_{HH}^{(\ell)}[x, y]\right)
$}
\end{equation}

\paragraph{Wavelet Ratio \cite{Xie2006}:} The ratio of the squared norm of the high-frequency components of level $\ell$ of the wavelet decomposition of the image to the squared norm of the low-frequency components.
\begin{equation}
\contrastscore = \frac{\sum_{x, y} W_{LH}^{(\ell)}[x, y]^2 + W_{HL}^{(\ell)}[x, y]^2 +  W_{HH}^{(\ell)}[x, y]^2}{\sum_{x, y} W_{LL}^{(\ell)}[x, y]^2}
\end{equation}

\paragraph{Mean Wavelet Log-Ratio}: This is a baseline of our own design in which we modify the ``Wavelet Ratio'' model to compute a local log-ratio between the high-frequency and low-frequency energy at each spatial location in one level of a wavelet decomposition, and then compute the mean of those log-ratios. We add $1$ to the denominator to prevent numerical issues.
\begin{equation}
\resizebox{0.99\linewidth}{!}{%
$
\displaystyle \contrastscore = \frac{1}{n} \sum_{x, y} \log \left( \frac{W_{LH}^{(\ell)}[x, y]^2 + W_{HL}^{(\ell)}[x, y]^2 +  W_{HH}^{(\ell)}[x, y]^2}{W_{LL}^{(\ell)}[x, y]^2 + 1} \right) 
$
}
\end{equation}

\paragraph{Eigenvalue Trace \cite{Wee2007}:} The image is reduced to a matrix where each column is a vector containing the intensity values of each non-overlapping patch (here, of size $4 \times 4$) in the image. The trace of the sample covariance of that matrix is then used as a measure of sharpness.
\begin{equation}
\contrastscore = \operatorname{trace}(\operatorname{cov}(\operatorname{im2col}(I, 4)))
\end{equation}

\paragraph{Mean Local Ratio \cite{Helmli2001}:} A local measure of contrast is computed at each pixel by considering the ratio of each pixel intensity to a local average, and the overall contrast is computed as the average of those ratios (rectified if they are below $1$) across the image. The numerator and denominator of each ratio are incremented by $1$ to avoid numerical issues.
\begin{equation}
\resizebox{0.99\linewidth}{!}{%
$
\displaystyle \contrastscore = \frac{1}{n} \sum_{x, y} \operatorname{max} \left( \frac{\operatorname{blur}(I, \sigma)[x, y] + 1}{I[x,y] + 1}, \frac{I[x,y] + 1}{\operatorname{blur}(I, \sigma)[x, y] + 1} \right)
$
}
\end{equation}
Where $\operatorname{blur}(I, \sigma)$ applies a Gaussian blur of standard deviation $\sigma$ to image $I$.

\paragraph{Mean Local Log-Ratio:} This is a baseline of our own design in which we modify the ``Mean Local Ratio'' technique above, by using the geometric mean of ratios instead of the arithmetic mean.
\begin{equation}
\contrastscore = \exp \left( \frac{1}{n} \sum_{x, y} \left| \log \left(  \frac{I[x,y] + 1}{\operatorname{blur}(I, \sigma)[x, y] + 1 }  \right) \right| \right)
\end{equation}

\paragraph{Mean Local Norm-Dist-Sq:} This is another baseline of our own design, in which we modify the ``Mean Local Ratio'' technique to use normalized squared distance (similar to a Coefficient of Variation) instead of ratios, which improves performance.
\begin{equation}
\contrastscore = \frac{1}{n} \sum_{x, y} \frac{ (I[x,y] - \operatorname{blur}(I, \sigma)[x, y])^2 }{\operatorname{blur}(I, \sigma)[x, y]^2 + 1 }
\end{equation}

\subsection{Dual-Pixel / Stereo Baseline Algorithms}

\newcommand{\disparityloss}{f}

Because our images are taken from a dual pixel (DP) sensor, our focal stack can be thought of as a stack of left and right images in a stereo pair $\{(L, R)\}$. When a patch is in focus, the left and right DP images should resemble each other. It is therefore possible to construct simple autofocus algorithms by taking each left/right image pair $(L, R)$ in a DP focal stack, compute some measure of mismatch between those two images $\disparityloss$, and return the focal index that minimizes that loss. In this section, we describe the baseline algorithms we use for this approach. Because patches of the the left and right DP images may have drastically different global brightnesses due to lens shading (especially when the patches are taken from the periphery of the entire image frame), these stereo-like algorithms must be invariant to global transformations of the input images. For this reason, we center each image by its mean and divide by its standard deviation before computing all stereo measures:
\begin{equation}
{\hat {L}} = \frac{L - \mu(L)}{\operatorname{Var}(L)}, \quad {\hat {R}} = \frac{R - \mu(R)}{\operatorname{Var}(R)}
\label{eq:zero_normalize}
\end{equation}
This has no effect on some models (such as census and rank transformations) but is critical for other models.

\paragraph{Census Transform (Hamming) \cite{zabih1994}:} We apply the census transformation to the left and right DP images, wherein each pixel is represented by an 8-length binary vector representing whether or not the pixel is greater than each of its 8 neighbors. We score each pair according to the total Hamming distance between the two census-transformed images.
\begin{gather}
    \disparityloss = \sum_{x, y} \norm{ \operatorname{census}(L)[x,y] - \operatorname{census}(R)[x,y] }_0 \nonumber \\
    \operatorname{census}(I)[x,y] = \bigg[ I[x + \Delta_x,y + \Delta_y] > I[x,y] \nonumber \\
    \big|\, \Delta_x \in [-1, 0, 1], \Delta_y \in [-1, 0, 1], \Delta_x \neq \Delta_y \neq 0 \bigg] 
\end{gather}

\paragraph{Rank Transform (L1) \cite{zabih1994}:} We apply the rank transformation (the $0$-norm of the census transformation) to the left and right DP images, and score each pair according to the L1 distance between the two rank-transformed images.
\begin{gather}
\disparityloss = \sum_{x, y} \norm{ \operatorname{rank}(L)[x,y] - \operatorname{rank}(R)[x,y] }_1 \\
\operatorname{rank}(I)[x,y] = \norm{\operatorname{census}(I)[x,y]}_0
\end{gather}

\paragraph{Ternary Census \cite{stein2004efficient}:} We apply the ternary census transformation to the left and right DP images, wherein each pixel is represented by an 8-length ternary vector representing if the pixel is greater than, less than, or close to (according to some threshold $\epsilon$) each of its 8 neighbors. We then score each pair according to the total L1 distance between the two census-transformed images.
\begin{gather}
    \disparityloss = \sum_{x, y} \norm{ \operatorname{census^3}(L)[x,y] - \operatorname{census^3}(R)[x,y] }_1 \nonumber \\
    \operatorname{census^3}(I)[x,y] = \bigg[ \operatorname {tsgn} \left( I[x + \Delta_x,y + \Delta_y] - I[x,y] \right) \nonumber \\
    \big|\, \Delta_x \in [-1, 0, 1], \Delta_y \in [-1, 0, 1], \Delta_x \neq \Delta_y \neq 0 \bigg] \nonumber \\
    \operatorname{tsng}(x, \epsilon) = \operatorname {sgn}(x) \left[ \left|  x \right| > \epsilon \right] 
\end{gather}

\paragraph{Normalized Cross-Correlation \cite{Barnea1972ACO, Hannah1974}:} 
NCC is just the inner product of these two normalized images, with its sign flipped such that minimization results in maximum cross-correlation. This is equivalent to minimizing the normalized sum of squared distances between the two images.
\begin{equation}
\disparityloss = -\left\langle \hat{L},  \hat{R} \right\rangle
\end{equation}

\paragraph{Normalized SAD \cite{Hannah1974}:} The sum of absolute deviations between the two normalized images.
\begin{equation}
\disparityloss = \sum_{x, y} \left| \hat{L}[x, y] - \hat{R}[x, y] \right|
\end{equation}

\paragraph{Normalized Envelope (L1) \cite{birchfield1998}: } Pixel matching techniques can be made invariant to the discrete sampling of the sensor by adapting them to operate on smooth upper and lower envelopes of image intensities. Here we compute an upper and lower envelope of the left and right images, and from them compute the total L1 distance between the extents of the left and right envelopes.
\begin{align}
\disparityloss = \textstyle \sum_{x, y} \displaystyle & \left| \max\left(0, \hat{L}_\mathrm{lo}[x,y] - \hat{R}_\mathrm{hi}[x,y]\right) \right| \\
+ & \left| \max\left(0, \hat{R}_\mathrm{lo}[x,y] - \hat{L}_\mathrm{hi}[x,y]\right) \right| \nonumber \\
\hat{L}_\mathrm{lo} = \mathrm{min2} &\left(\mathrm{blur2}\left(\hat L\right)\right), \quad L_\mathrm{hi} = \mathrm{max2}\left(\mathrm{blur2}\left(\hat L\right)\right)  \nonumber
\end{align}
where $\mathrm{max2}(\cdot)$ is a $2 \times 2$ ``max'' filter (i.e.\ max pooling), $\mathrm{min2}(\cdot)$ is a $2 \times 2$ ``min'' filter (i.e.\ min pooling), and $\mathrm{blur2}(\cdot)$ is a $2 \times 2$ box filter (i.e.\ average pooling). $R_\mathrm{lo}$ and $\hat{R}_\mathrm{hi}$ are defined similarly.

\paragraph{Normalized Envelope (L2) \cite{birchfield1998}: } Similarly, we can compute the total squared distance between the extents of the left and right envelopes.
\begin{align}
\disparityloss = \textstyle \sum_{x, y} \displaystyle & \max\left(0, \hat{L}_\mathrm{lo}[x,y] - \hat{R}_\mathrm{hi}[x,y]\right)^2 \nonumber \\
+  & \max\left(0, \hat{R}_\mathrm{lo}[x,y] - \hat{L}_\mathrm{hi}[x,y]\right)^2
\end{align}

\subsection{Single-Slice Baseline Algorithms}

The baseline methods above infer the in-focus index by either maximizing contrast $\contrastscore$ (for contrast-based methods) or minimizing stereo mismatch $\disparityloss$ (for dual-pixel methods). Hence, they all require the knowledge of the entire focal stack before making a prediction.

However, the DP algorithms can be extended to predict the in-focus index with just one input DP image pair, if we can establish the relationship between left/right disparity $d$ and ideal focus distance $z^*$. We list a few such algorithms below.

\paragraph{SSD Disparity: }
We use the block matching approach of \cite{WadhwaSIGGRAPH2018} to estimate disparity. In order to convert the disparity of a patch to a focal depth, we fit a linear model that estimates focal depth from the median patch disparity. The linear model is robustly estimated from all training patches using RANSAC. This methods computes depth over $1.5\times$ reduced field of view and we report results only on patches contained within that field of view. A narrower field of view is not unfair to the baseline as PSF variations and focal breathing are worse near the periphery.

\paragraph{Learned Depth: }
We use the neural network based approach of \cite{GargICCV2019} to predict depth from dual-pixel images.  The model from \cite{GargICCV2019} predicts depth maps up to an unknown affine transform, which we estimate by solving a least squares problem that minimizes the $L2$ distance between the affine transformed depth map and the disparity from \cite{WadhwaSIGGRAPH2018} that are known to be linearly related. We use the same fitting described in SSD Disparity and restrict evaluation to the same $1.5\times$ reduced field of view.

\paragraph{ZNCC Disparity with Calibration: }
We compute the zero-normalized cross correlation between the input DP image pair $(L, R)$ (using Equation~\ref{eq:zero_normalize}) to get $(\hat{L}, \hat{R})$. Then, we compute disparity between $\hat{L}$ and $\hat{R}$ \cite{Barnea1972ACO, Hannah1974} and apply a precomputed calibration to convert disparity to focal distance. Specifically, to compute disparity $d$, we do the following
\begin{equation}
    d = \argmax_{\delta} \left\langle \hat{L}[x, y], \hat{R}[x + \delta, y] \right\rangle
\end{equation}
 for integer $\delta$ in a small range around zero. We then refine $d$ to get sub-pixel resolution by fitting a quadratic near the peak and finding its supremum.

Under paraxial and thin-lens approximations, and assuming constant aperture and focal length, signed disparity $d$ and ideal focus distance $z^*$ are related by an affine transform \cite{WadhwaSIGGRAPH2018}:
\begin{equation}
    d = C \left( \frac{1}{z} - \frac{1}{z^*} \right)
    \label{eq:zncc_with_calibration}
\end{equation}
where $C$ is a calibration constant and $z$ is the lens's current focus distance.

The assumption that $C$ is a constant breaks down for real lenses as they do not satisfy the paraxial and thin-lens approximations. In fact, the value of $C$ varies significantly across the field of view, due to optical aberration, vignetting, changes in optical blur kernels, etc., as shown in \cite{WadhwaSIGGRAPH2018}. The camera device we use embeds a factory calibration table that specifies the measured $C$ values sparsely across the field of view. We obtain the value of $C$ for each input patch by bilinearly interpolating the low-resolution calibration table.

With the knowledge of disparity $d$, calibration coefficient $C$, and current focus distance $z$, we can easily solve for $z^*$ in Equation~\ref{eq:zncc_with_calibration}.

\section{Generalization to other phones}

To show that our technique generalizes, we use the data captured in the paper to create a new test set using the ``left'' camera, which has a different calibration and PSF.

This left test set contains the same scenes as the test set in the original paper; however, the overall attributes of the set may be different. The ``left'' phone is positioned in front of the ``center'' phone by 1.1 cm (on the z-axis, it is +1.1cm closer to objects in the scene). In addition, the computed depth has an overall lower confidence than that of the center camera since fewer cameras see all the pixels captured by the left camera. This problem is particularly apparent on the left side of the capture. In addition, because we keep the same confidence threshold as used for the center camera, fewer patches will be generated. In general, it may be difficult to compare the raw numbers from the test set using the center camera and the test set using the left camera.

As shown in Table \ref{table:left}, all techniques report slightly lower numbers. This indicates that the ``left'' test-set may be more difficult than the ``center'' test-set due to the aforementioned changes. Despite this, our model still outperforms the baselines. Additionally, several simple techniques, like adding calibration data to the model or a brief fine-tuning stage for each camera, could be easily added to our approach and potentially lead to improved per-device performance.

For this run, ZNCC Disparity uses calibration for the ``left'' camera, and linear models to convert to focal depths for SSD Disparity and Learned Depth were estimated using the training data patches from the ``left'' camera.

\begin{table}
\resizebox{\linewidth}{!}{
\begin{tabular}{cc|cccc|cc}
&& \multicolumn{4}{c}{higher is better} & \multicolumn{2}{|c}{lower is better} \\
& Algorithm & $ = 0$ & $\leq 1$ & $\leq 2$ & $\leq 4$ & MAE & RMSE  \\ \hline\hline

\dualpixelone & Learned Depth$^\dagger$ \cite{GargICCV2019} & 0.070\cellcolor{orange} & 0.206\cellcolor{orange} & 0.340\cellcolor{orange} & 0.564\cellcolor{orange} & 7.224\cellcolor{orange} & 11.010\cellcolor{yellow}  \\
\dualpixelone & SSD Disparity$^\dagger$ \cite{WadhwaSIGGRAPH2018} & 0.068\cellcolor{yellow}  & 0.200\cellcolor{yellow}  & 0.333\cellcolor{yellow}  & 0.550\cellcolor{yellow}  & 7.377\cellcolor{yellow}  & 10.951\cellcolor{orange}     \\
\dualpixelone & ZNCC Disparity & 0.046 & 0.136  & 0.224  & 0.379 & 9.436 & 13.138 \\
\dualpixelone & Our model & 0.105\cellcolor{red} & 0.322\cellcolor{red} & 0.513\cellcolor{red} & 0.807\cellcolor{red} & 2.912\cellcolor{red} & 3.867\cellcolor{red} \\ \hline
\end{tabular}}
\caption{Evaluating techniques on the ``left'' version of the test set. This tests whether the technique generalizes to other phones. Note our model still outperforms the baselines and that the performance went down for all techniques indicating that the ``left'' version of the test set is harder. See text for explanation. A $\dagger$ indicates that patches within a $1.5\times$ reduced field of view were used.}\label{table:left}
\end{table}

\section{Light and Dark Scenes}

In Figure 8 in the main paper, we presented examples on particularly dark images. In Table \ref{fig:lightdark}, we present the full numeric breakdowns of the performance of single-index algorithms on scenes with a normal amounts of light versus scenes with low light.

To capture these, we placed the rig in a fixed position and then captured two focal stacks: one with the light on and then one with the light turned off. As a result, these captures should be perfectly registered and should be the identical besides the presence or absence of light. We then used the ground truth depth from the light image to eliminate any possible mistakes that the SFM pipeline would have with the darker images.

\begin{table}
\resizebox{\linewidth}{!}{
\begin{tabular}{c|cl@{\,\,}|cccc|cc}
&&& \multicolumn{4}{c}{higher is better} & \multicolumn{2}{|c}{lower is better} \\
Setting & & Algorithm & $ = 0$ & $\leq 1$ & $\leq 2$ & $\leq 4$ & MAE & RMSE  \\ \hline\hline

Light & \dualpixelone & SSD Disparity$^\dagger$ \cite{WadhwaSIGGRAPH2018} & 0.079\cellcolor{yellow}  & 0.228\cellcolor{yellow}  & 0.355\cellcolor{yellow}  & 0.528\cellcolor{yellow}  & 6.732\cellcolor{yellow}  & 9.577\cellcolor{yellow}  \\
 & \dualpixelone & Learned Depth$^\dagger$ \cite{GargICCV2019} & 0.094\cellcolor{orange} & 0.264\cellcolor{orange} & 0.401\cellcolor{orange} & 0.576\cellcolor{orange} & 6.262\cellcolor{orange} & 9.376\cellcolor{orange} \\
 & \dualpixelone & ZNCC Disparity & 0.064 & 0.188 & 0.304 & 0.486 & 7.222 & 10.179  \\
 & \dualpixelone & Our model & 0.126\cellcolor{red}     & 0.369\cellcolor{red}     & 0.578\cellcolor{red}     & 0.832\cellcolor{red}     & 2.654\cellcolor{red}     & 3.563\cellcolor{red}     \\

\hline\hline

Dark & \dualpixelone & Learned Depth$^\dagger$ \cite{GargICCV2019} & 0.061\cellcolor{orange}     & 0.178\cellcolor{orange}     & 0.286\cellcolor{orange}     & 0.442\cellcolor{yellow}  & 9.104\cellcolor{yellow}  & 12.793  \\
 & \dualpixelone & SSD Disparity$^\dagger$ \cite{WadhwaSIGGRAPH2018} & 0.055  & 0.162  & 0.252  & 0.396  & 9.343  & 12.669\cellcolor{yellow}  \\
 & \dualpixelone & ZNCC Disparity & 0.056\cellcolor{yellow}  & 0.167\cellcolor{yellow}  & 0.272\cellcolor{yellow}  & 0.443\cellcolor{orange} & 7.972\cellcolor{orange} & 11.080\cellcolor{orange}     \\
 & \dualpixelone & Our model & 0.112\cellcolor{red}     & 0.323\cellcolor{red}     & 0.497\cellcolor{red}     & 0.729\cellcolor{red}     & 3.479\cellcolor{red}     & 4.957\cellcolor{red}     \\
\hline
\end{tabular}}
\caption{Performance for scenes in high and low light. Note that our technique is the most resistant to dark scenes. A $\dagger$ indicates that patches within a $1.5\times$ reduced field of view were used.}
\label{fig:lightdark}
\end{table}

\section{Example Images}

\subsection{Single slice as input}

\begin{figure*}
\vspace{-.4cm}
\includegraphics[width=\linewidth]{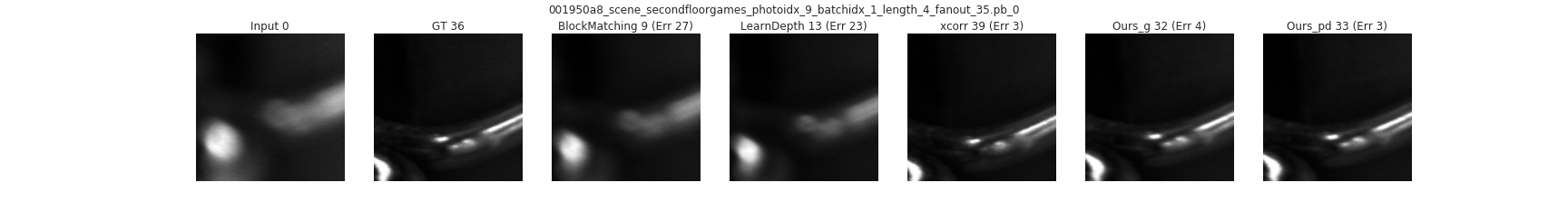}
\vspace{-.4cm}
\includegraphics[width=\linewidth]{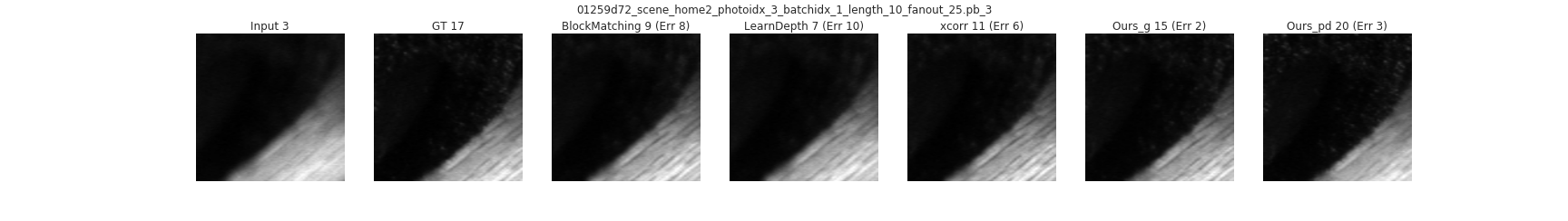}
\vspace{-.4cm}
\includegraphics[width=\linewidth]{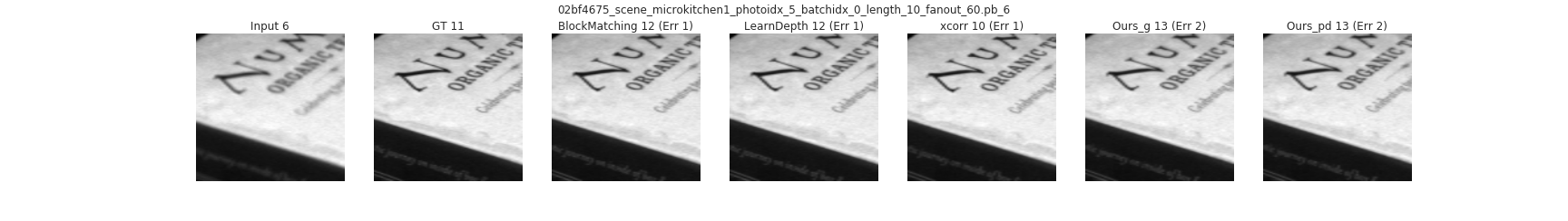}
\vspace{-.4cm}
\includegraphics[width=\linewidth]{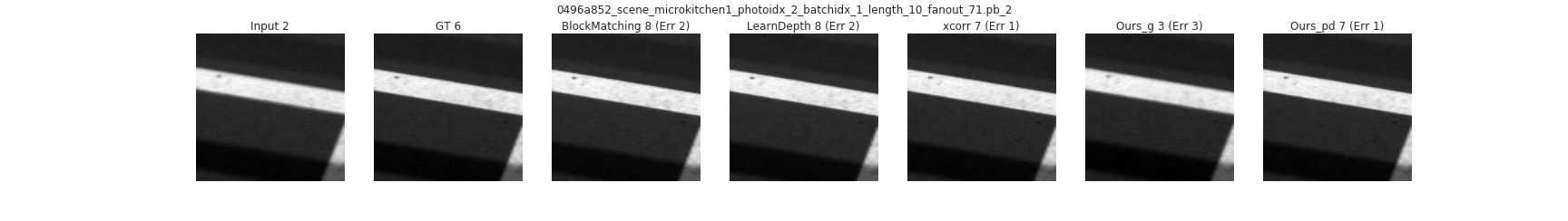}
\vspace{-.4cm}
\includegraphics[width=\linewidth]{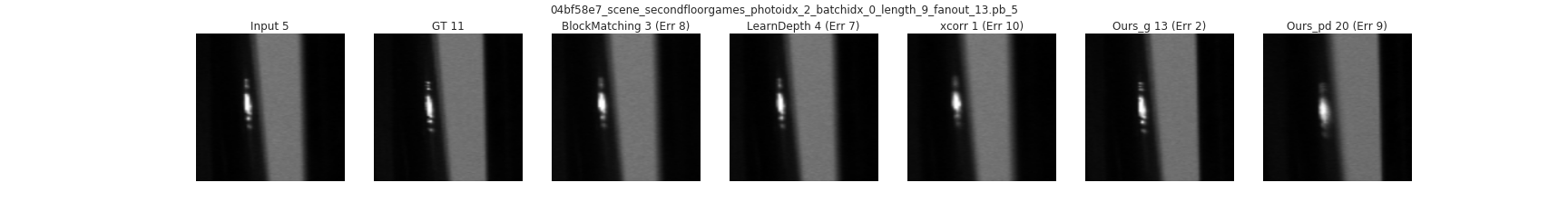}
\vspace{-.4cm}
\includegraphics[width=\linewidth]{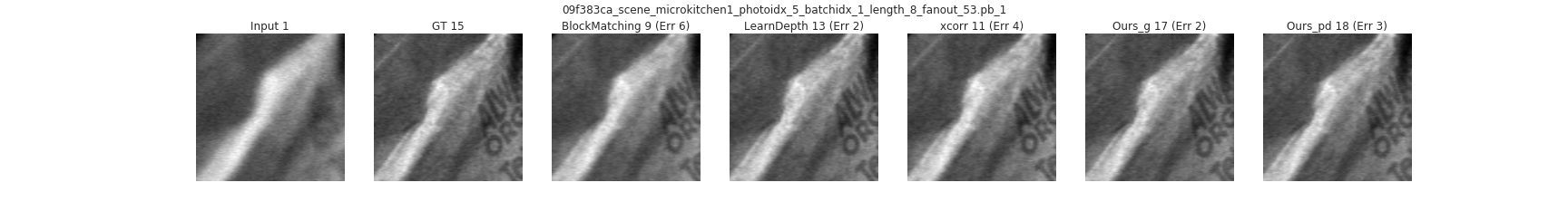}
\vspace{-.4cm}
\includegraphics[width=\linewidth]{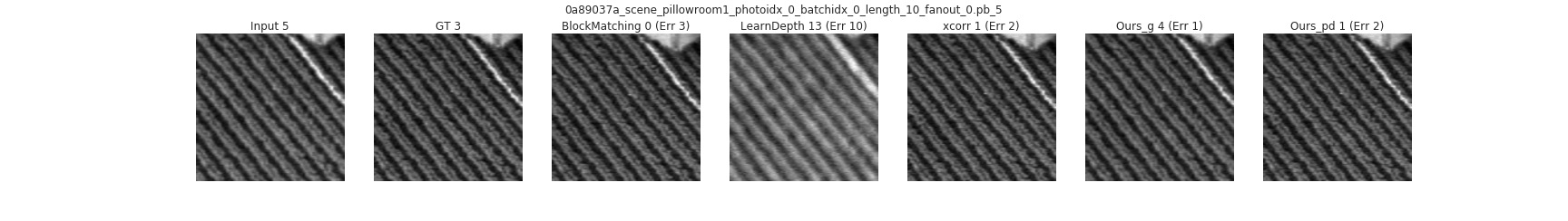}
\vspace{-.4cm}
\includegraphics[width=\linewidth]{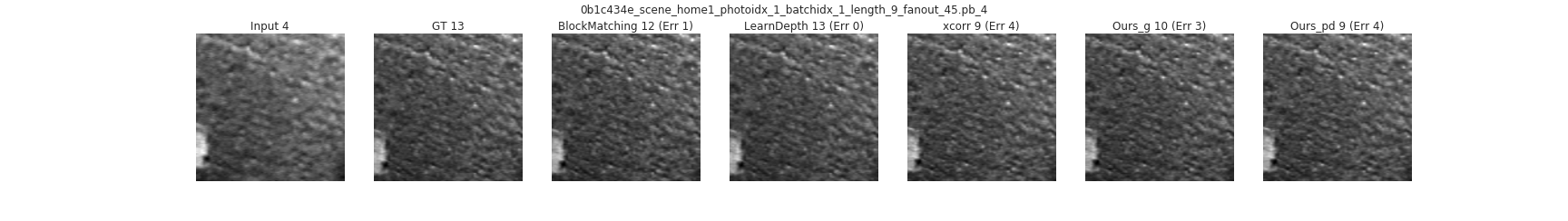}
\vspace{-.4cm}
\includegraphics[width=\linewidth]{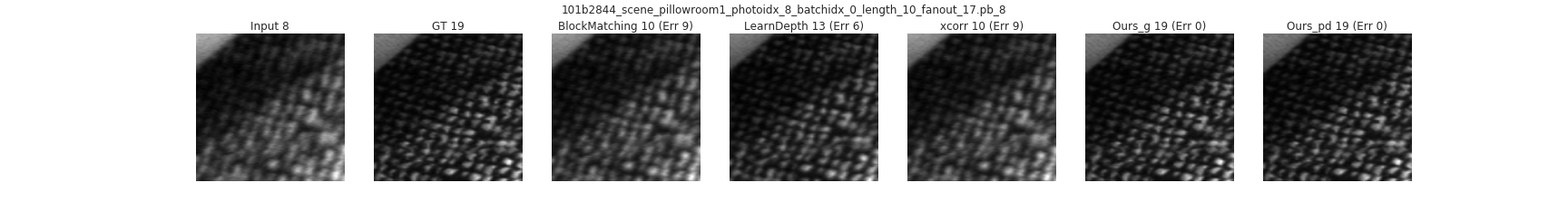}
\vspace{-.4cm}
\includegraphics[width=\linewidth]{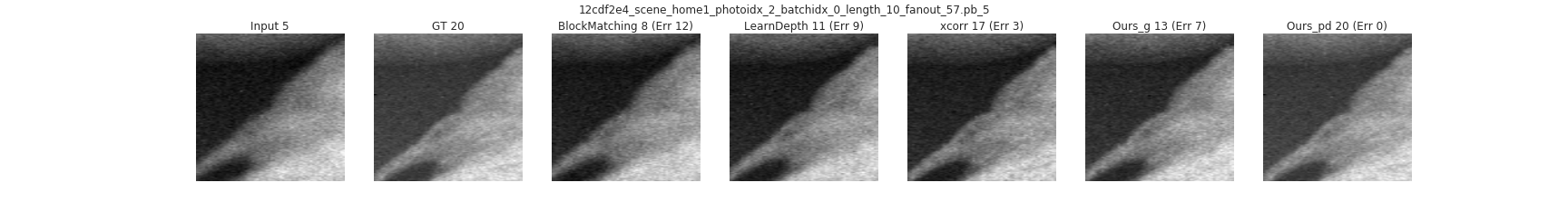}
\caption{Algorithms given singleindex. Example page 1}\label{fig:single1}
\end{figure*}

\begin{figure*}
\vspace{-.4cm}
\includegraphics[width=\linewidth]{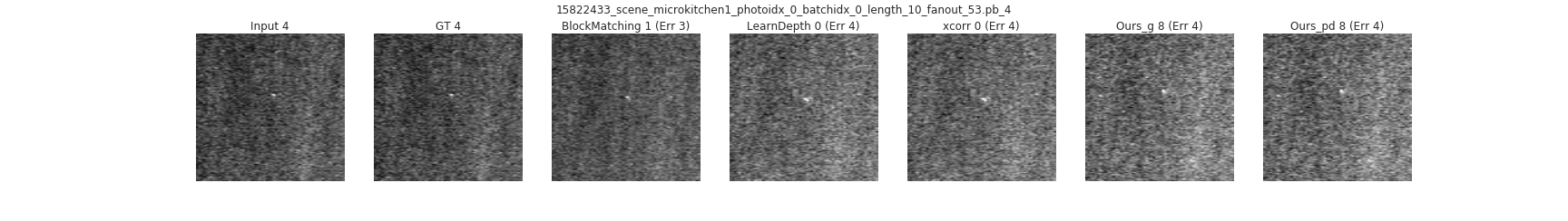}
\vspace{-.4cm}
\includegraphics[width=\linewidth]{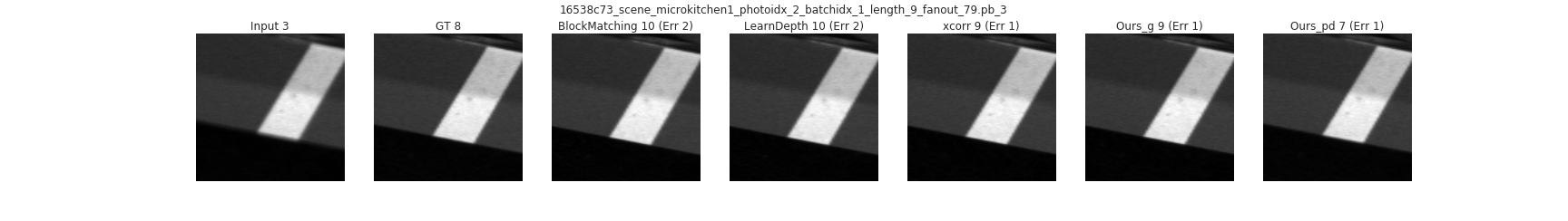}
\vspace{-.4cm}
\includegraphics[width=\linewidth]{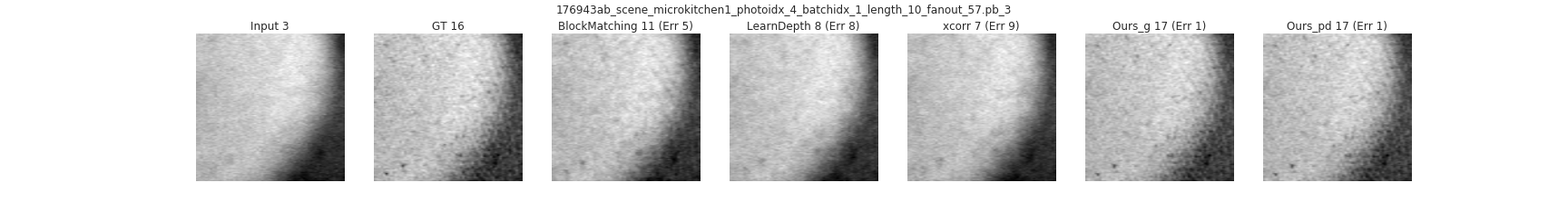}
\vspace{-.4cm}
\includegraphics[width=\linewidth]{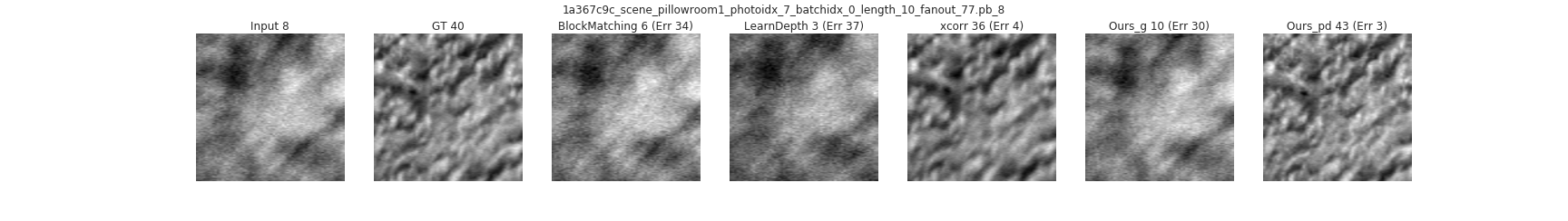}
\vspace{-.4cm}
\includegraphics[width=\linewidth]{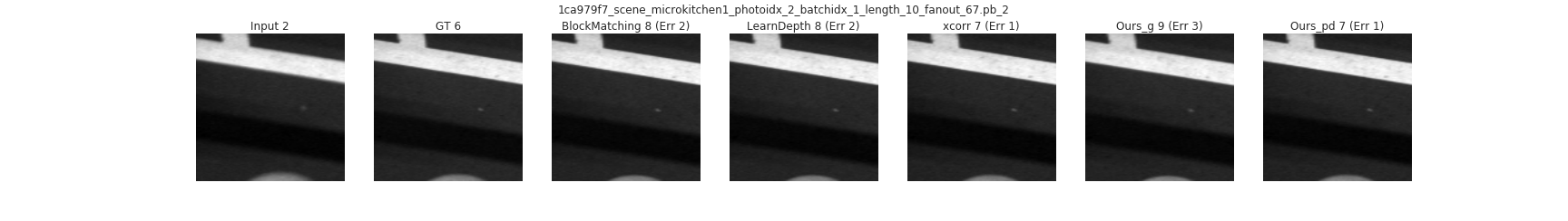}
\vspace{-.4cm}
\includegraphics[width=\linewidth]{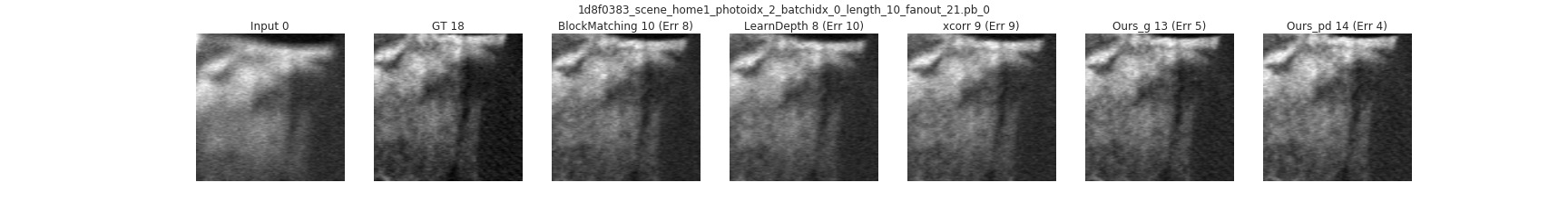}
\vspace{-.4cm}
\includegraphics[width=\linewidth]{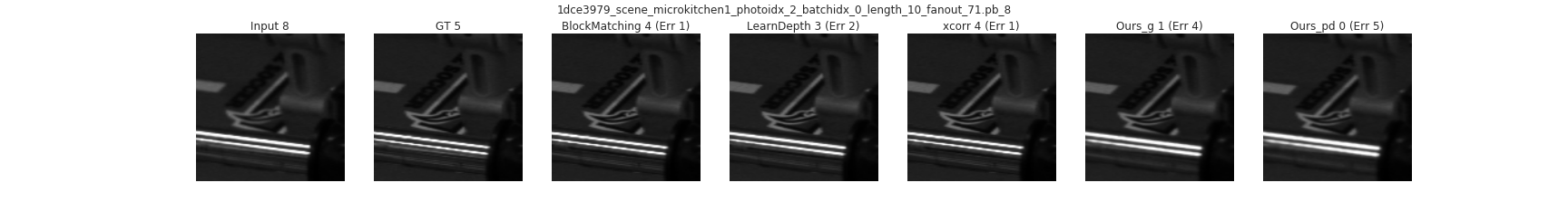}
\vspace{-.4cm}
\includegraphics[width=\linewidth]{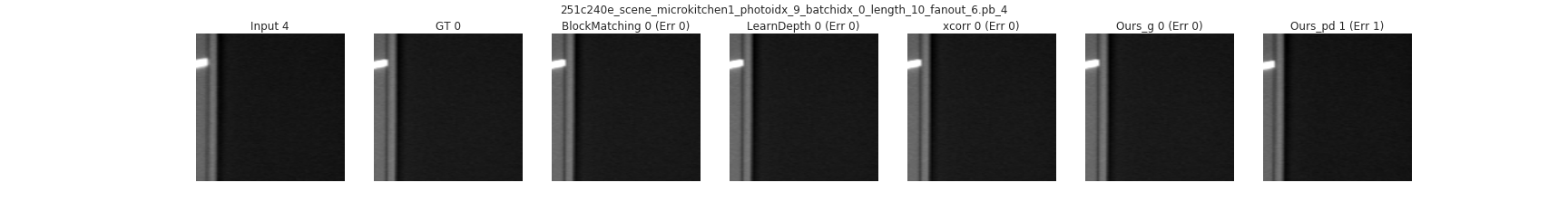}
\vspace{-.4cm}
\includegraphics[width=\linewidth]{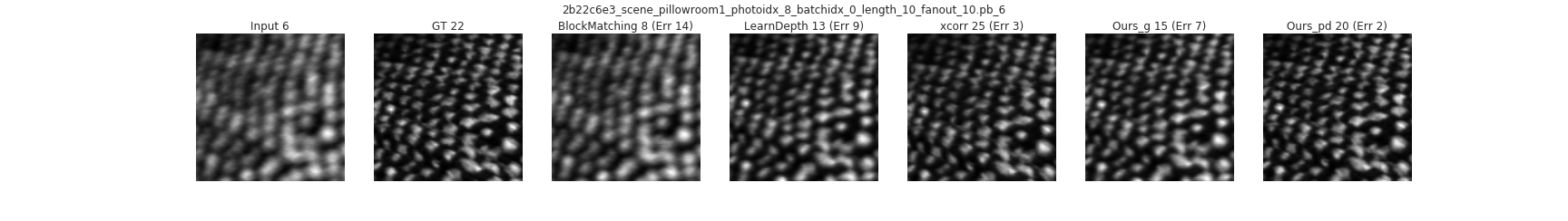}
\vspace{-.4cm}
\includegraphics[width=\linewidth]{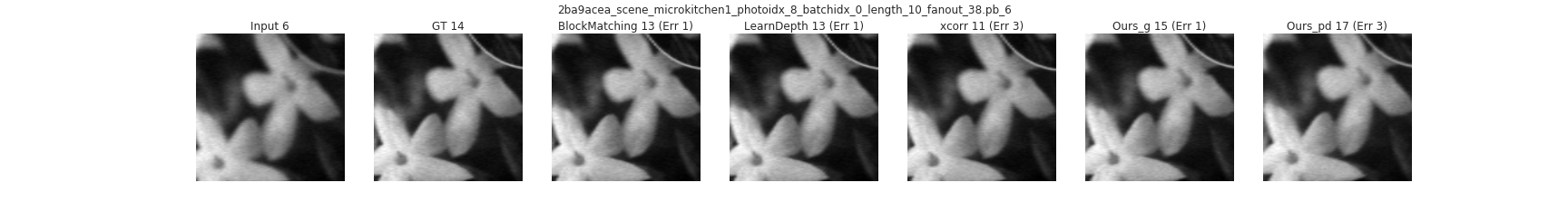}
\caption{Algorithms given singleindex. Example page 2}\label{fig:single2}
\end{figure*}

\begin{figure*}
\vspace{-.4cm}
\includegraphics[width=\linewidth]{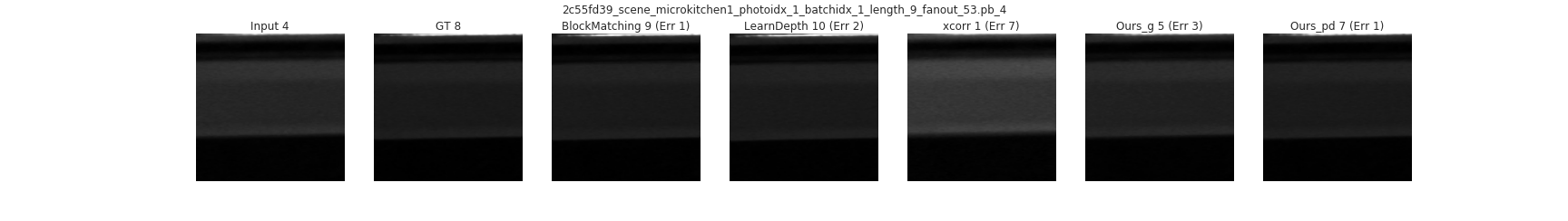}
\vspace{-.4cm}
\includegraphics[width=\linewidth]{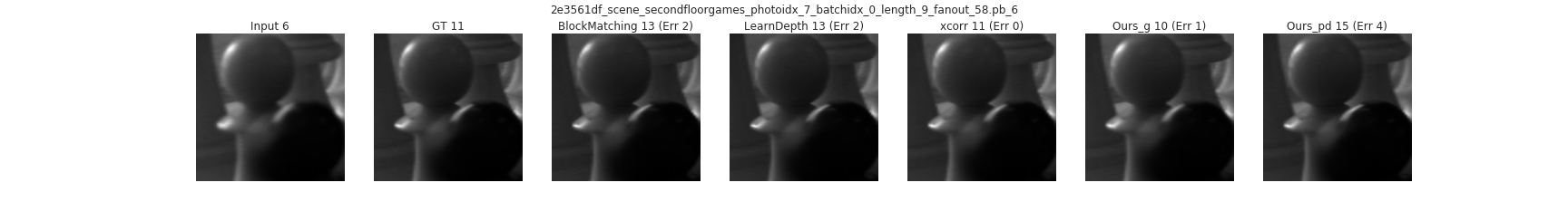}
\vspace{-.4cm}
\includegraphics[width=\linewidth]{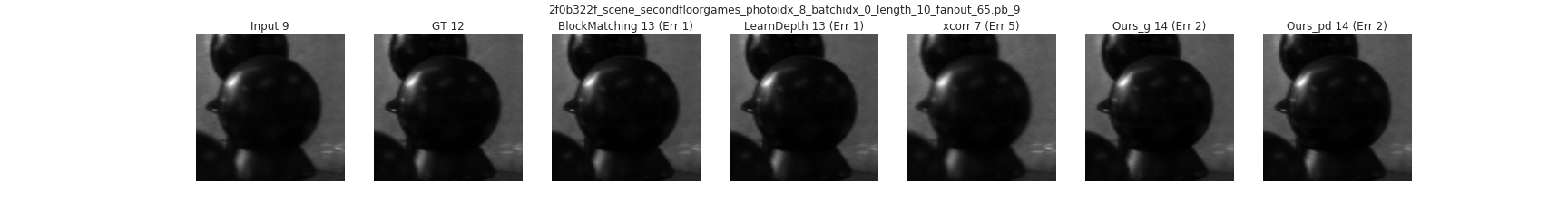}
\vspace{-.4cm}
\includegraphics[width=\linewidth]{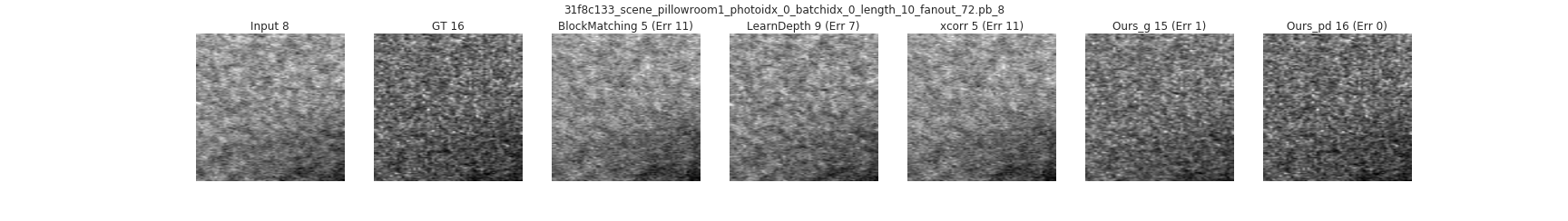}
\vspace{-.4cm}
\includegraphics[width=\linewidth]{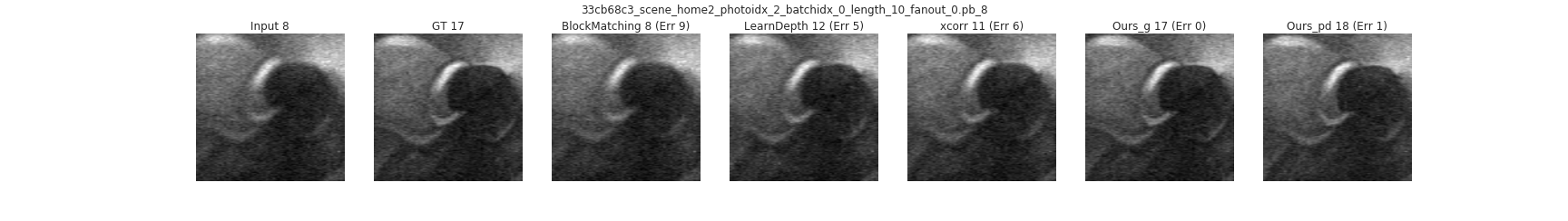}
\vspace{-.4cm}
\includegraphics[width=\linewidth]{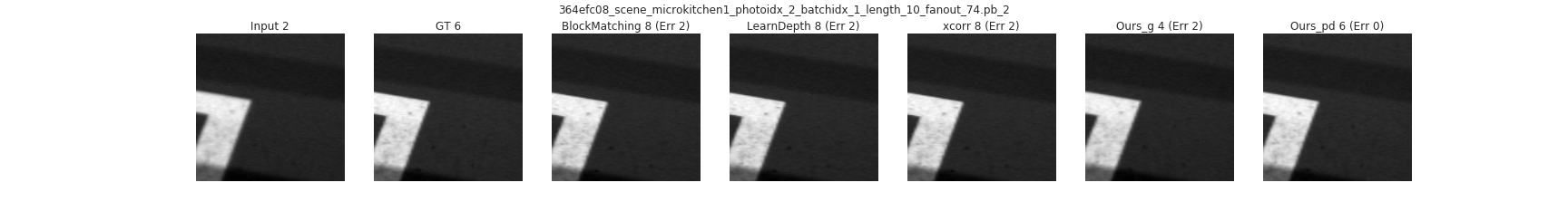}
\vspace{-.4cm}
\includegraphics[width=\linewidth]{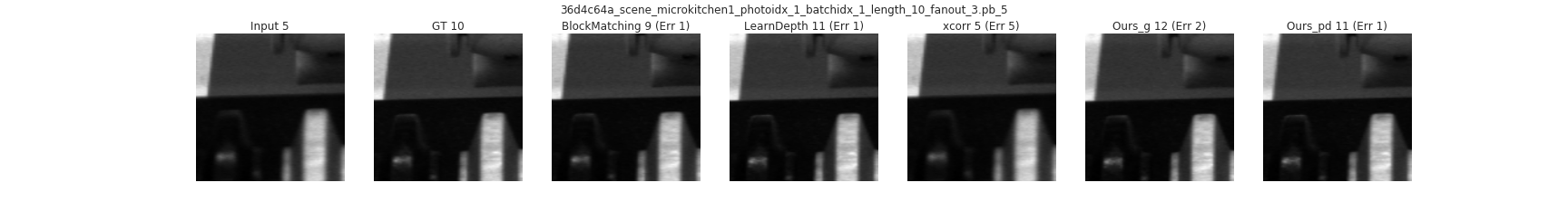}
\vspace{-.4cm}
\includegraphics[width=\linewidth]{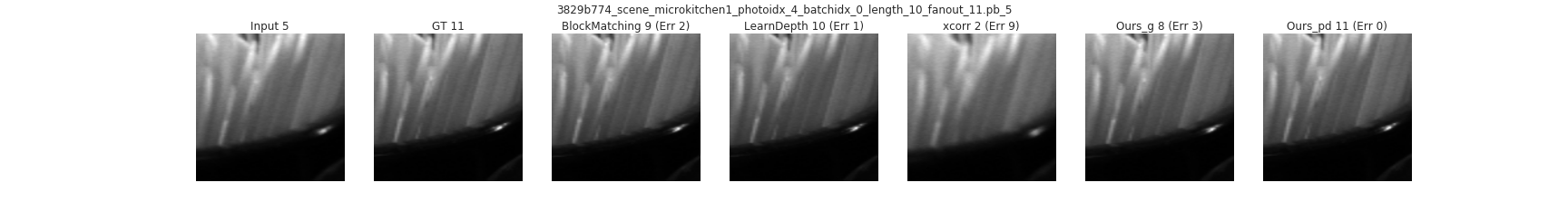}
\vspace{-.4cm}
\includegraphics[width=\linewidth]{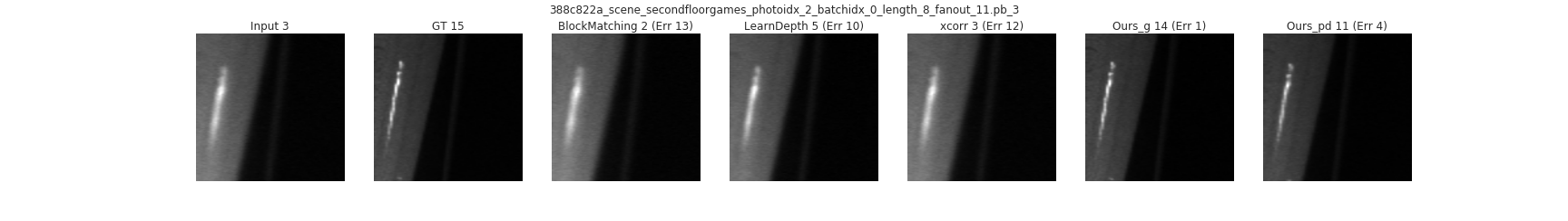}
\vspace{-.4cm}
\includegraphics[width=\linewidth]{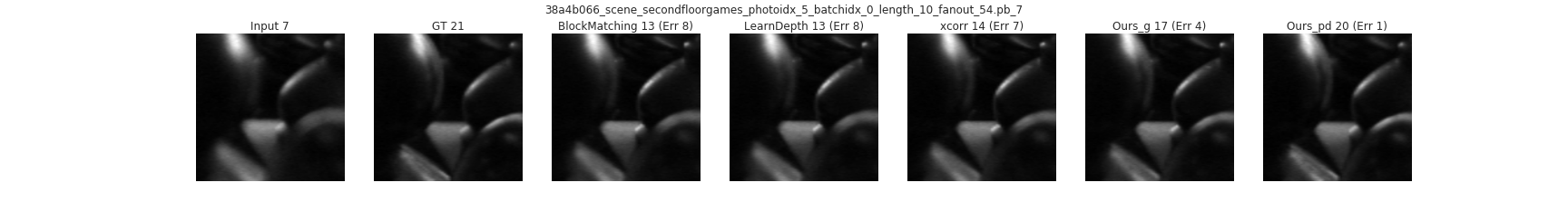}
\caption{Algorithms given singleindex. Example page 3}\label{fig:single3}
\end{figure*}

\begin{figure*}
\vspace{-.4cm}
\includegraphics[width=\linewidth]{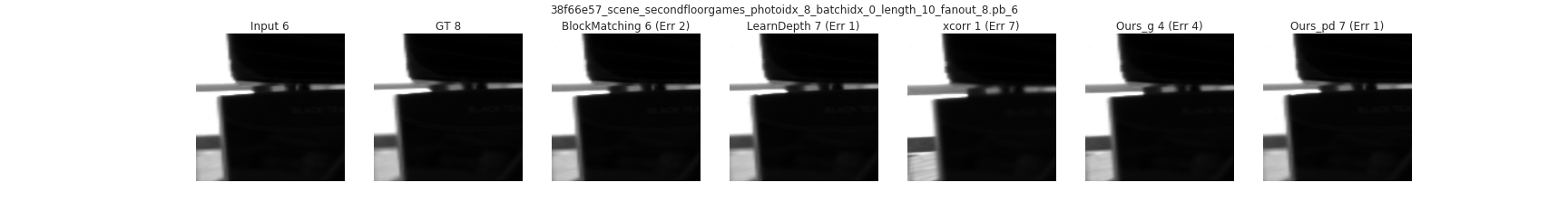}
\vspace{-.4cm}
\includegraphics[width=\linewidth]{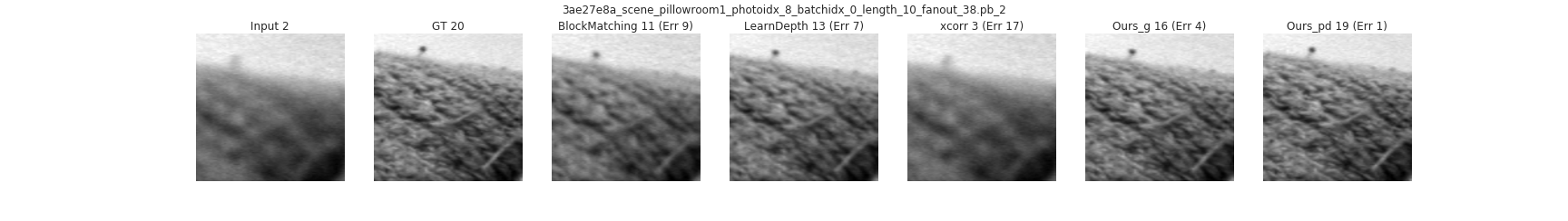}
\vspace{-.4cm}
\includegraphics[width=\linewidth]{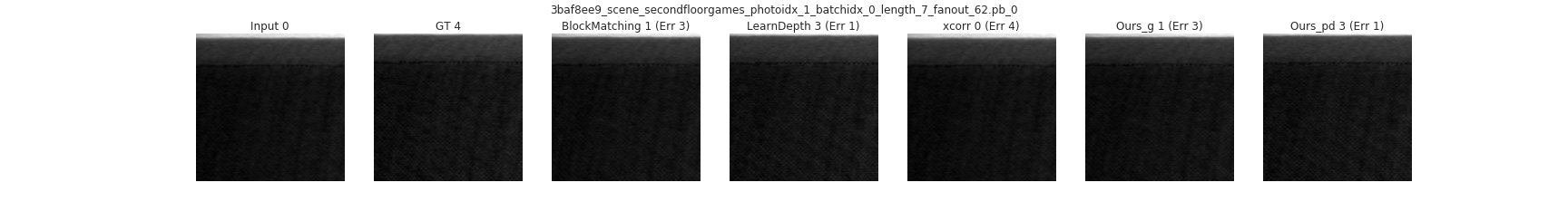}
\vspace{-.4cm}
\includegraphics[width=\linewidth]{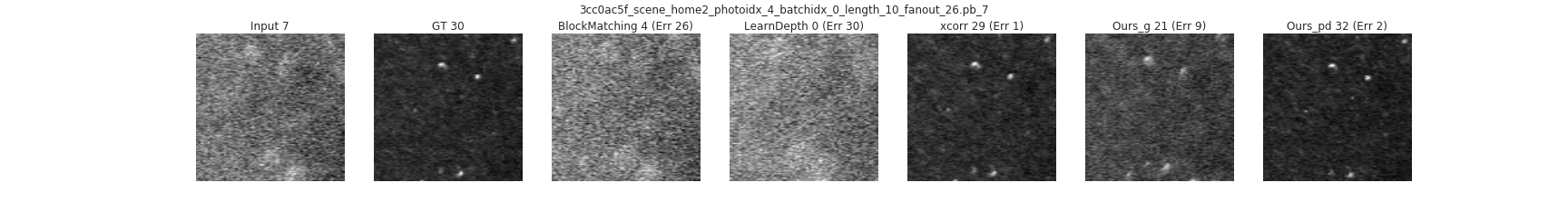}
\vspace{-.4cm}
\includegraphics[width=\linewidth]{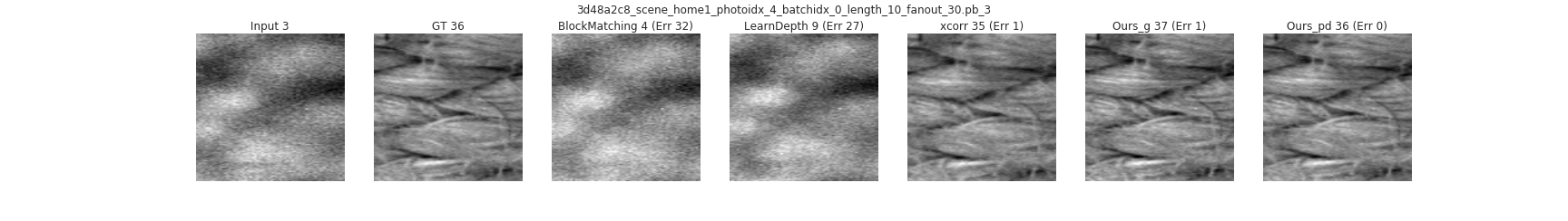}
\vspace{-.4cm}
\includegraphics[width=\linewidth]{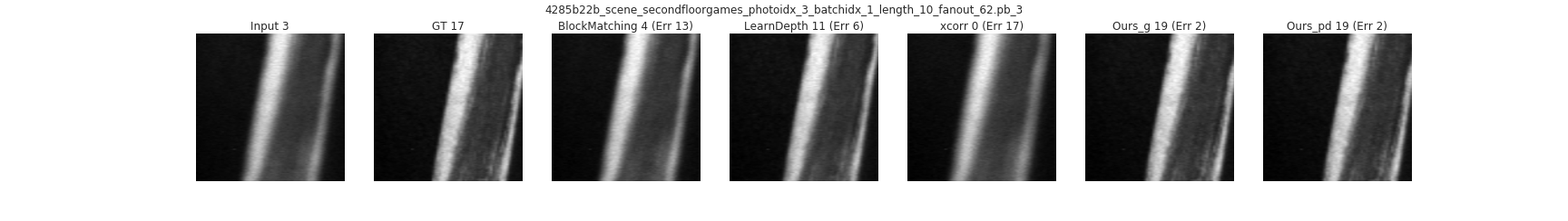}
\vspace{-.4cm}
\includegraphics[width=\linewidth]{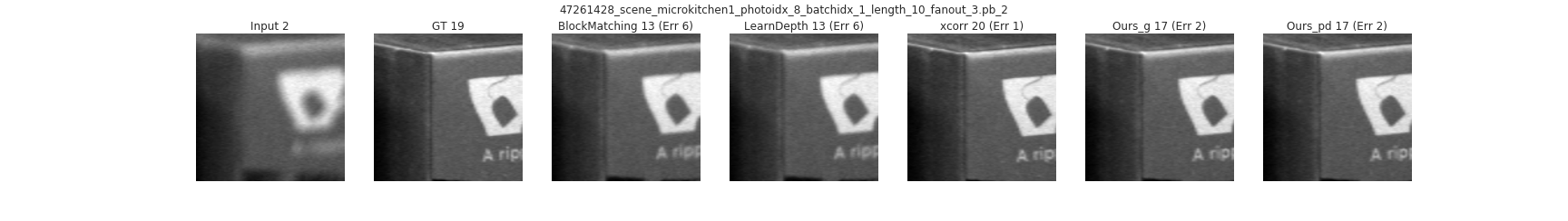}
\vspace{-.4cm}
\includegraphics[width=\linewidth]{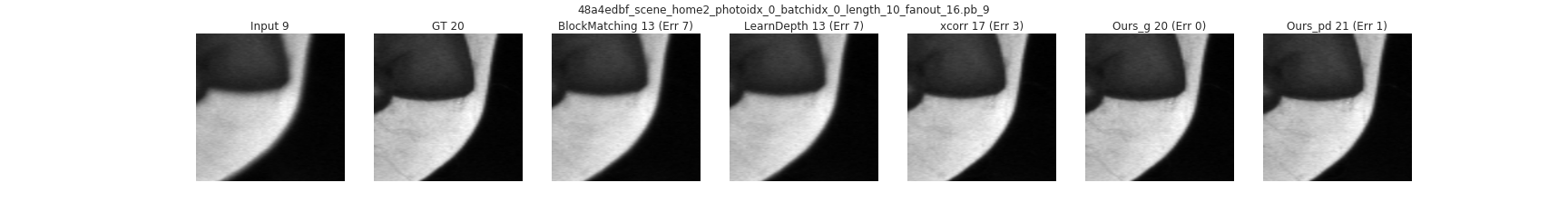}
\vspace{-.4cm}
\includegraphics[width=\linewidth]{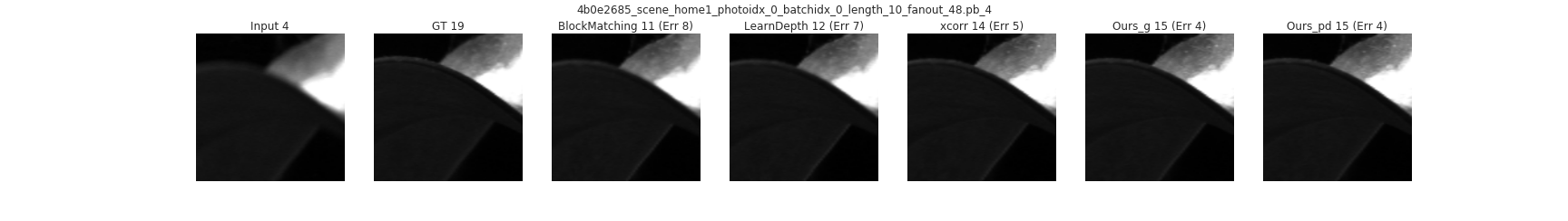}
\vspace{-.4cm}
\includegraphics[width=\linewidth]{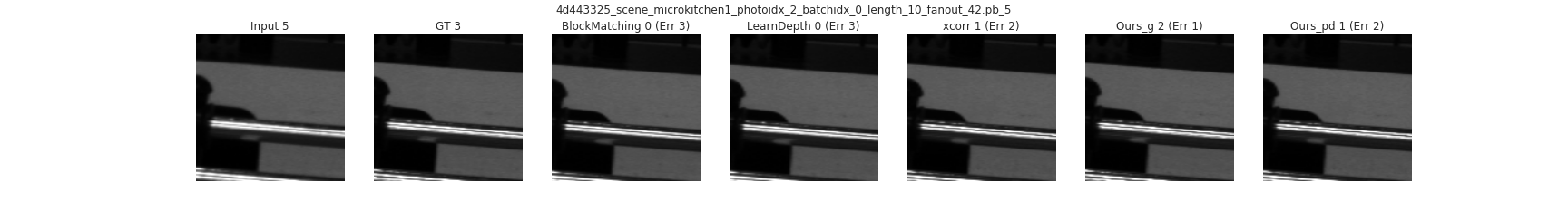}
\caption{Algorithms given singleindex. Example page 4}\label{fig:single4}
\end{figure*}

In Figures \ref{fig:single1}, \ref{fig:single2}, \ref{fig:single3}, \ref{fig:single4}, we provide a random selection of inputs (among those inside the 1.5x crop center, so that the PD baselines are present) and the predictions from all baselines and our models. The ``Input'' is what the algorithm is given. The focal stack identification key is directly above the row. The title of each focal slice is: the name of the algorithm, the index, (``Err'' followed by the number of indices away from the ground truth).

\subsection{Focal stack as input}
\begin{figure*}
\vspace{-.4cm}
\includegraphics[width=\linewidth]{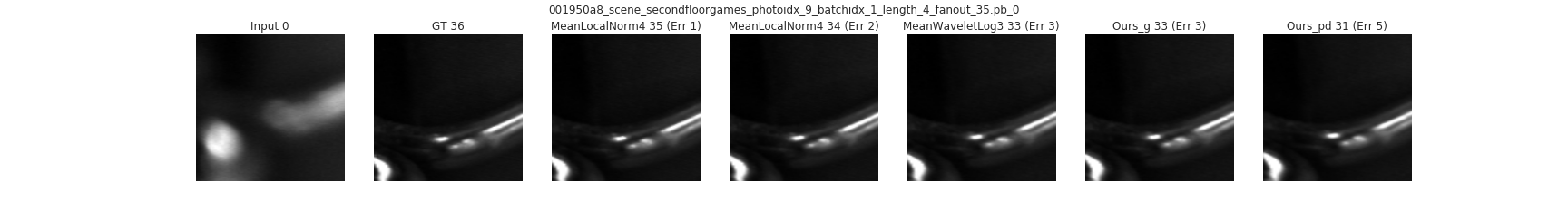}
\vspace{-.4cm}
\includegraphics[width=\linewidth]{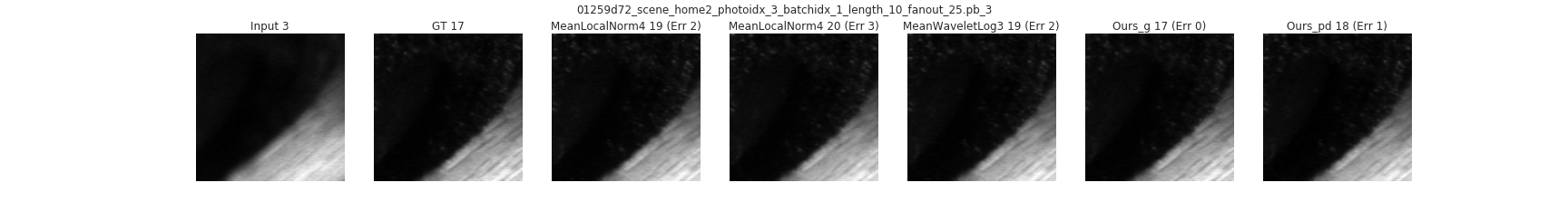}
\vspace{-.4cm}
\includegraphics[width=\linewidth]{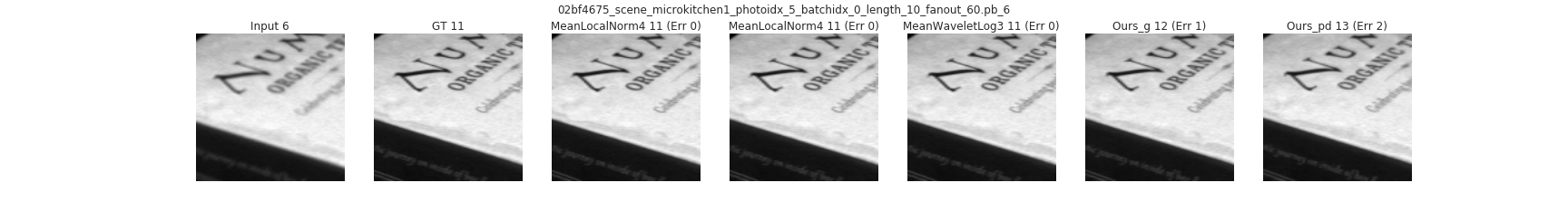}
\vspace{-.4cm}
\includegraphics[width=\linewidth]{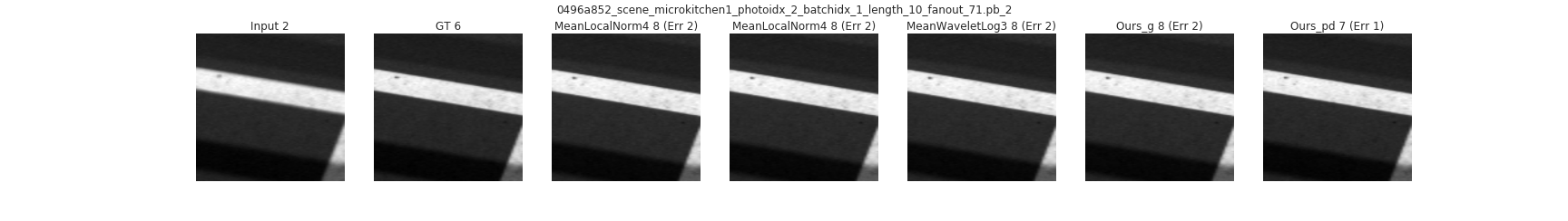}
\vspace{-.4cm}
\includegraphics[width=\linewidth]{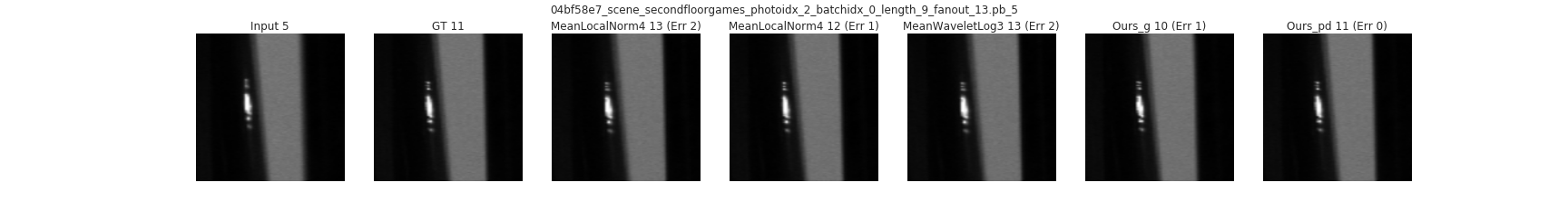}
\vspace{-.4cm}
\includegraphics[width=\linewidth]{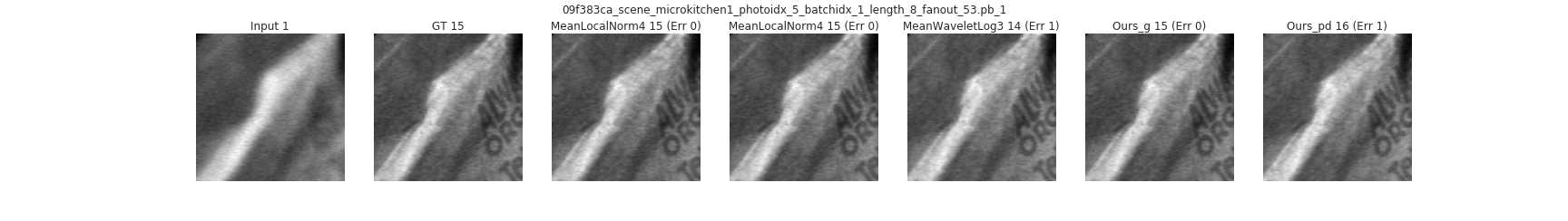}
\vspace{-.4cm}
\includegraphics[width=\linewidth]{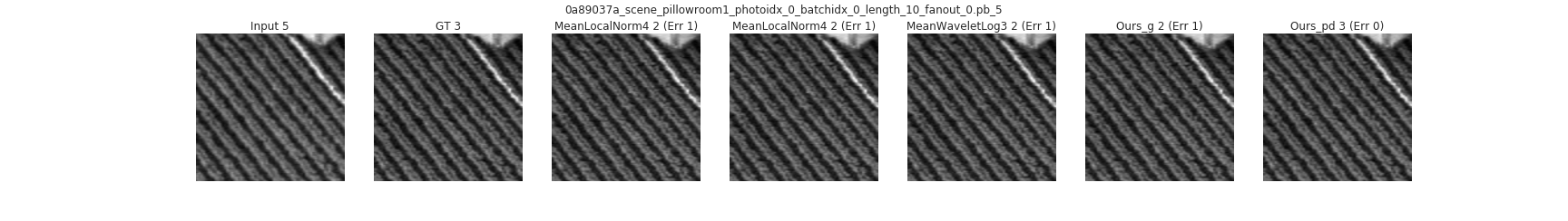}
\vspace{-.4cm}
\includegraphics[width=\linewidth]{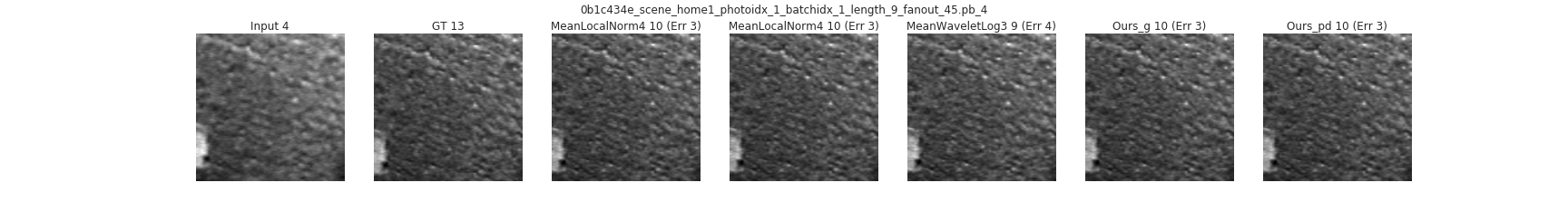}
\vspace{-.4cm}
\includegraphics[width=\linewidth]{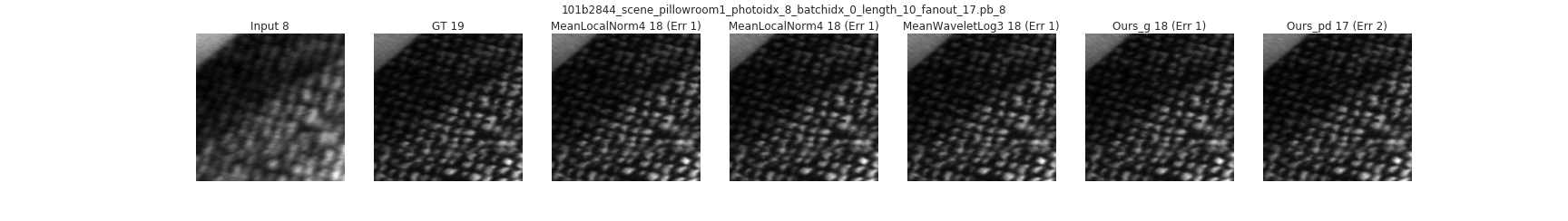}
\vspace{-.4cm}
\includegraphics[width=\linewidth]{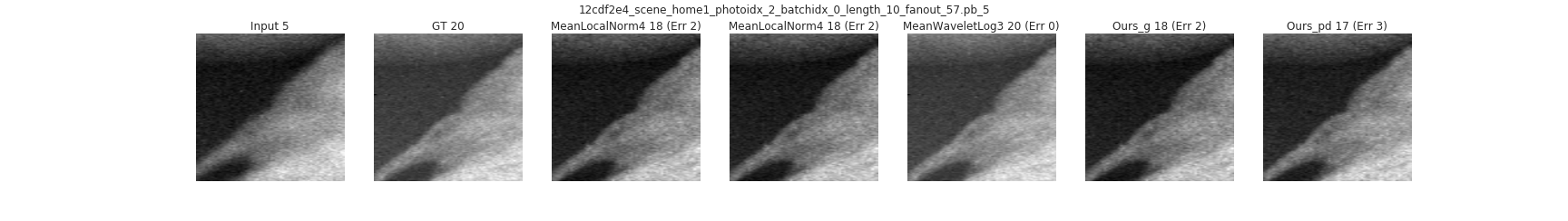}
\caption{Algorithms given fullfocal. Example page 1}\label{fig:full1}
\end{figure*}

\begin{figure*}
\vspace{-.4cm}
\includegraphics[width=\linewidth]{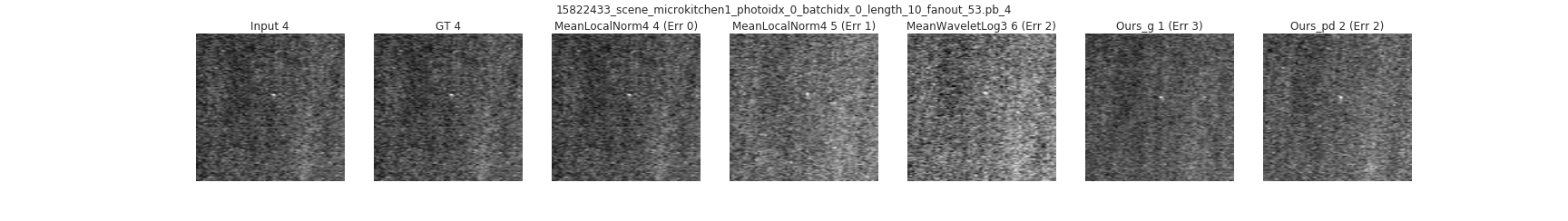}
\vspace{-.4cm}
\includegraphics[width=\linewidth]{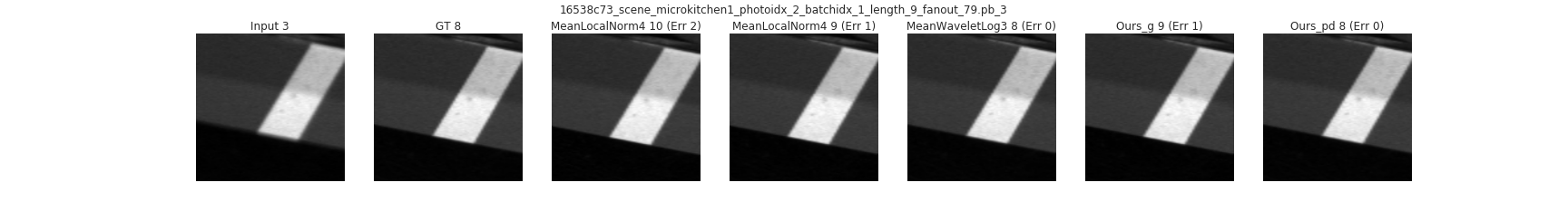}
\vspace{-.4cm}
\includegraphics[width=\linewidth]{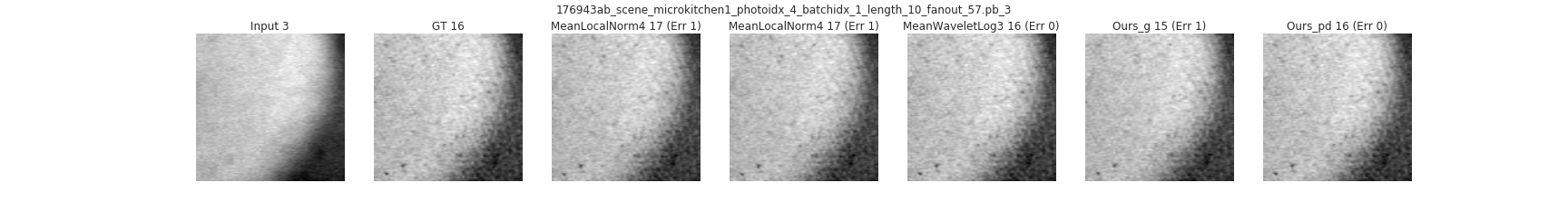}
\vspace{-.4cm}
\includegraphics[width=\linewidth]{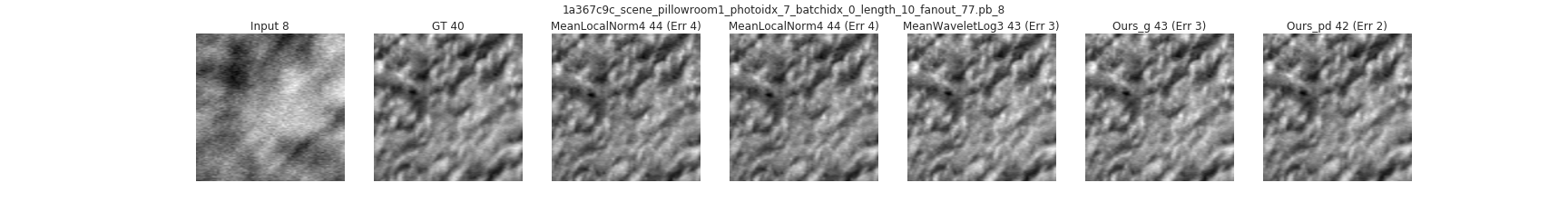}
\vspace{-.4cm}
\includegraphics[width=\linewidth]{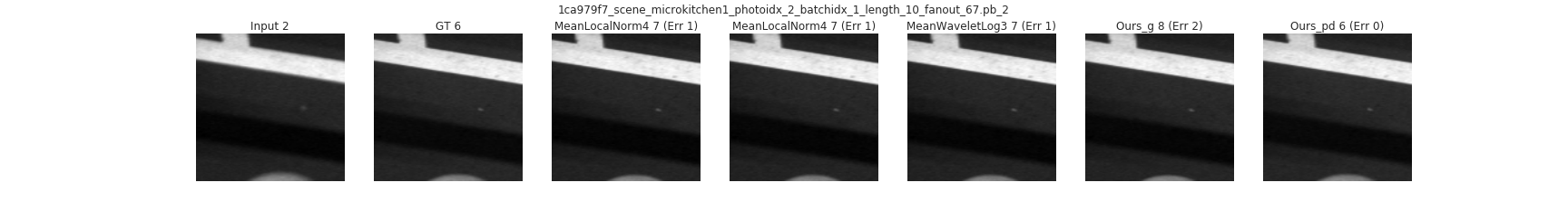}
\vspace{-.4cm}
\includegraphics[width=\linewidth]{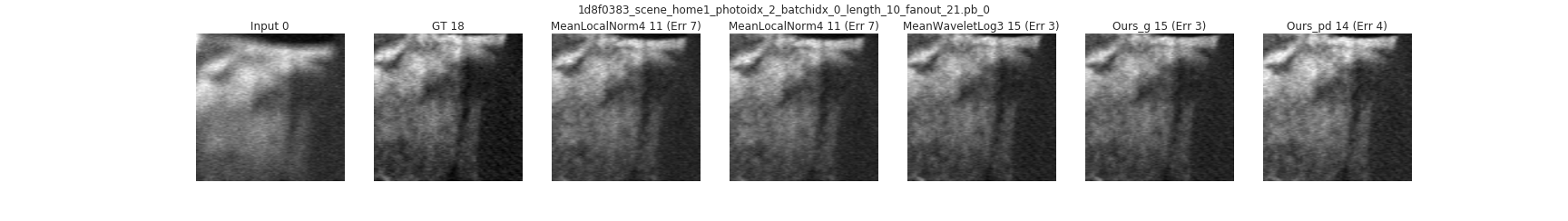}
\vspace{-.4cm}
\includegraphics[width=\linewidth]{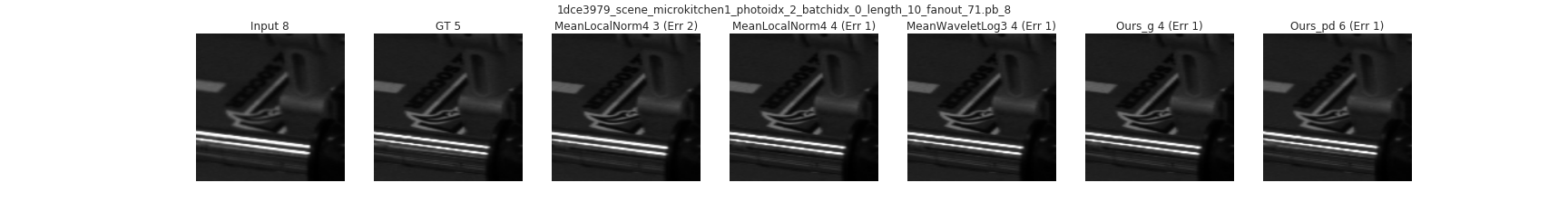}
\vspace{-.4cm}
\includegraphics[width=\linewidth]{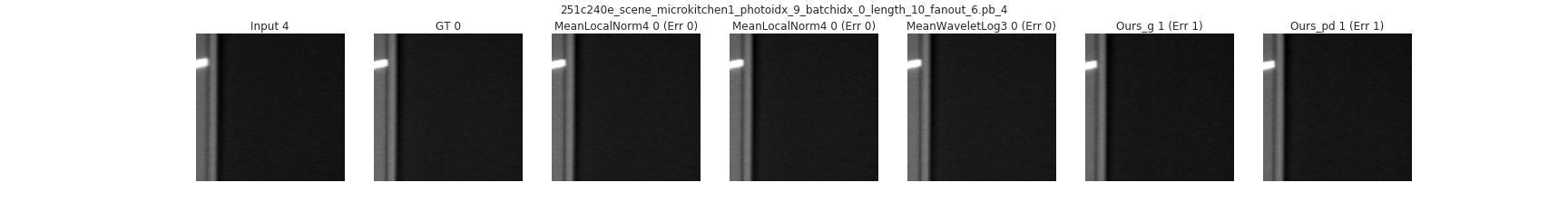}
\vspace{-.4cm}
\includegraphics[width=\linewidth]{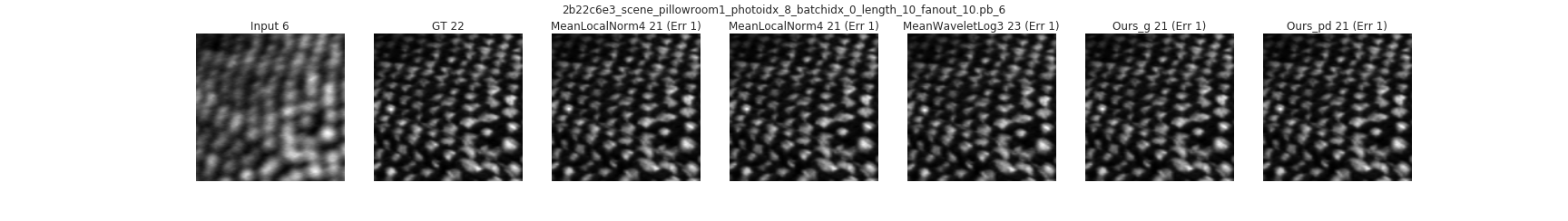}
\vspace{-.4cm}
\includegraphics[width=\linewidth]{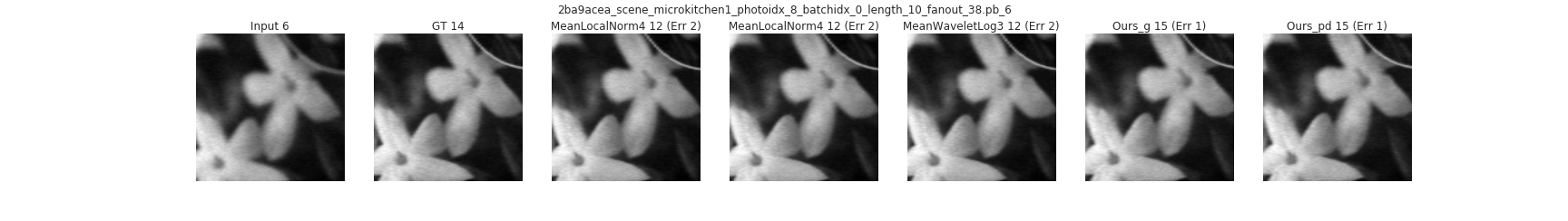}
\caption{Algorithms given fullfocal. Example page 2}\label{fig:full2}
\end{figure*}

\begin{figure*}
\vspace{-.4cm}
\includegraphics[width=\linewidth]{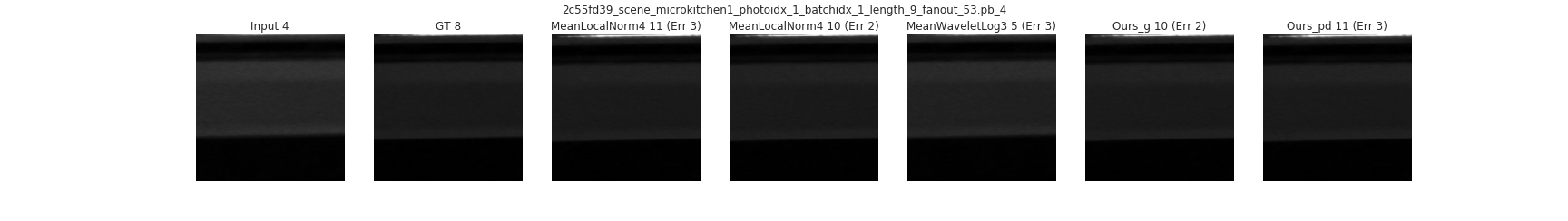}
\vspace{-.4cm}
\includegraphics[width=\linewidth]{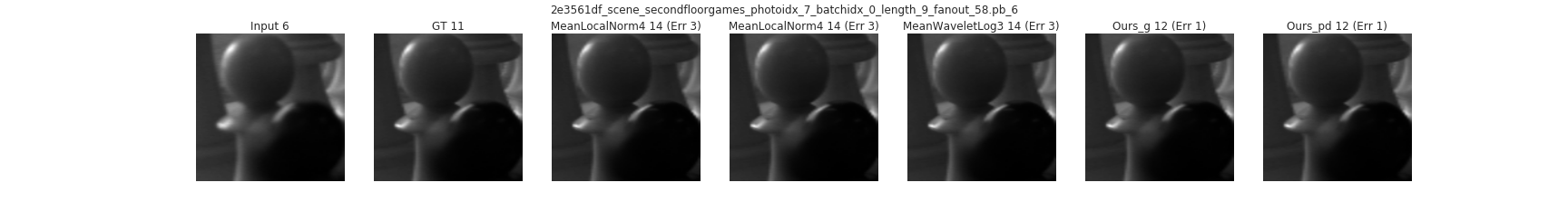}
\vspace{-.4cm}
\includegraphics[width=\linewidth]{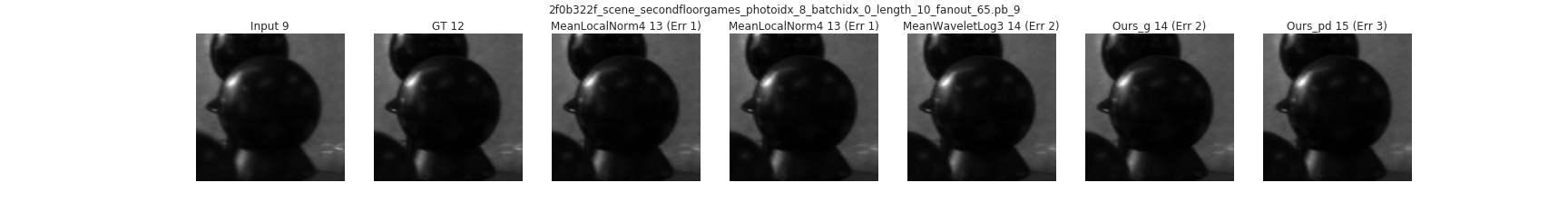}
\vspace{-.4cm}
\includegraphics[width=\linewidth]{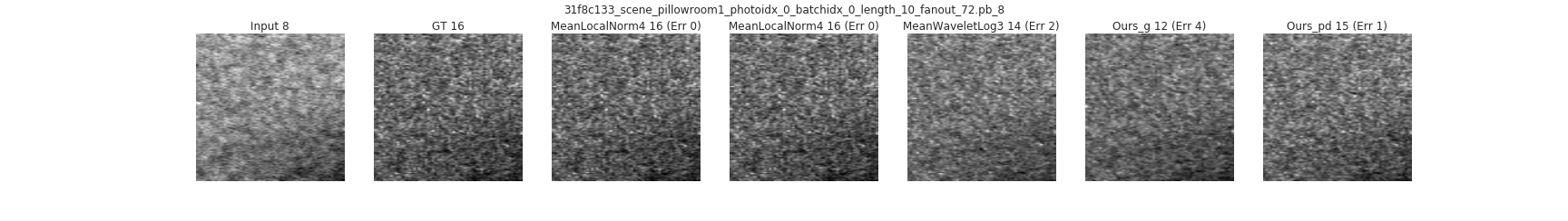}
\vspace{-.4cm}
\includegraphics[width=\linewidth]{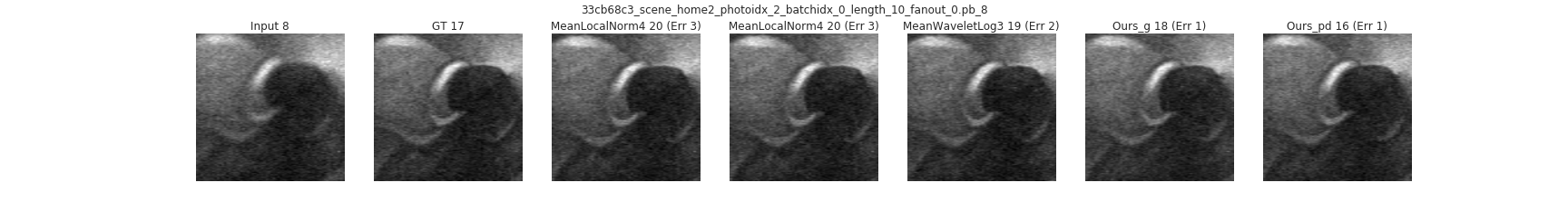}
\vspace{-.4cm}
\includegraphics[width=\linewidth]{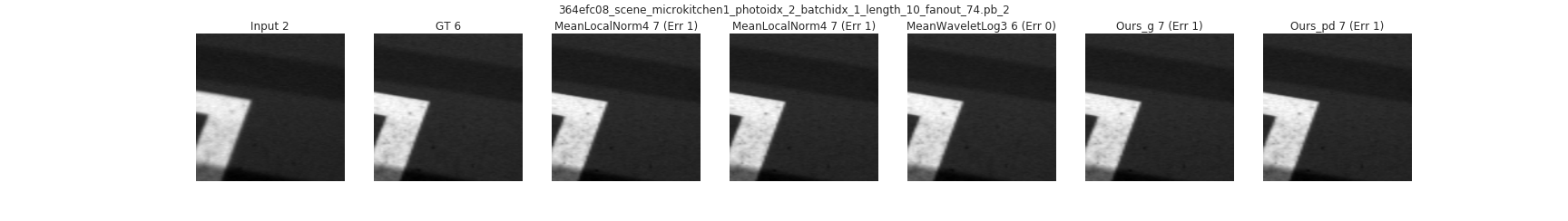}
\vspace{-.4cm}
\includegraphics[width=\linewidth]{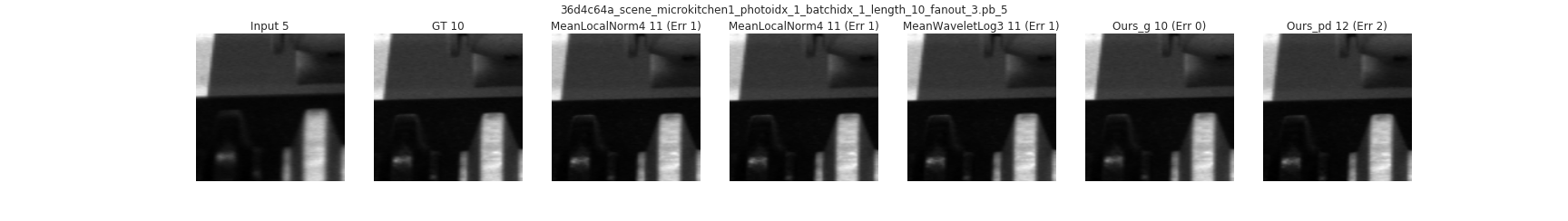}
\vspace{-.4cm}
\includegraphics[width=\linewidth]{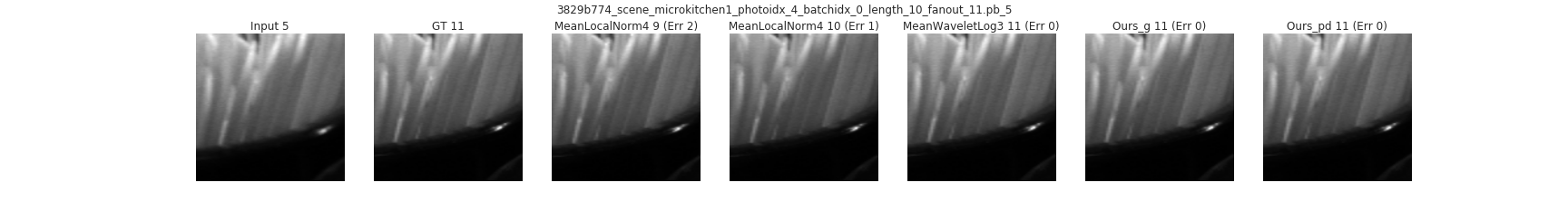}
\vspace{-.4cm}
\includegraphics[width=\linewidth]{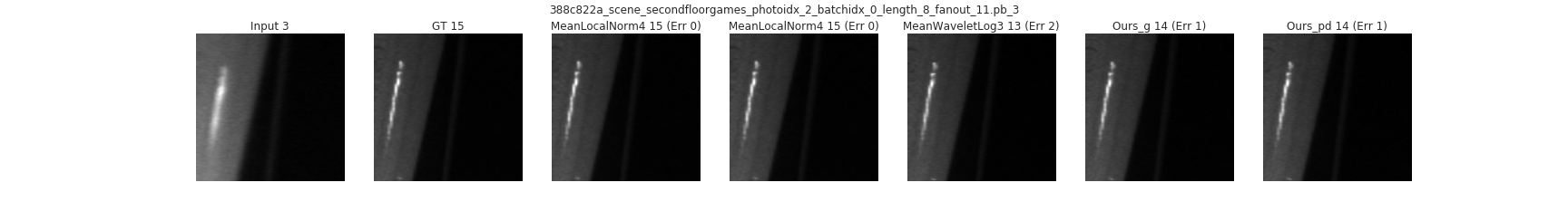}
\vspace{-.4cm}
\includegraphics[width=\linewidth]{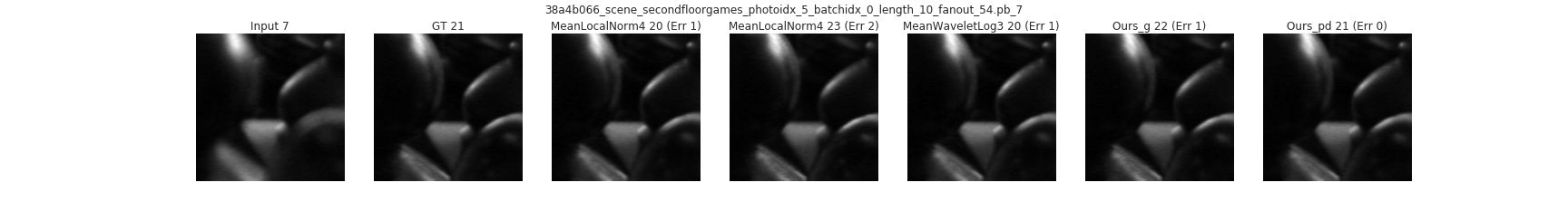}
\caption{Algorithms given fullfocal. Example page 3}\label{fig:full3}
\end{figure*}

\begin{figure*}
\vspace{-.4cm}
\includegraphics[width=\linewidth]{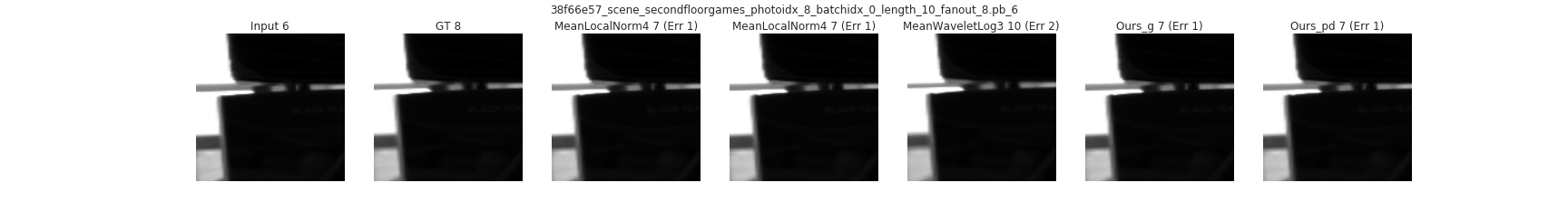}
\vspace{-.4cm}
\includegraphics[width=\linewidth]{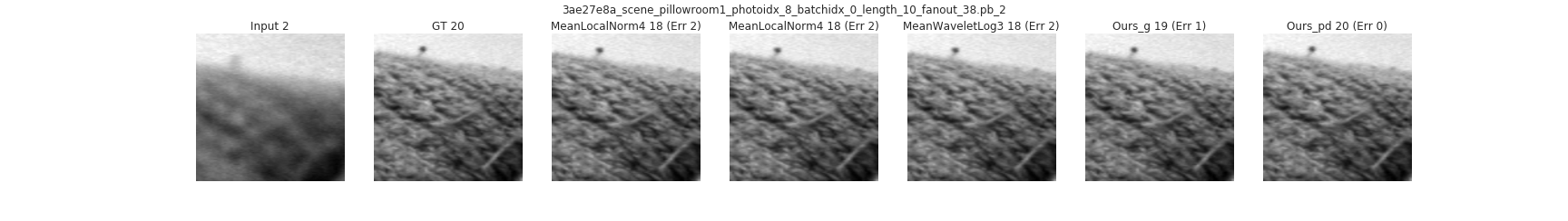}
\vspace{-.4cm}
\includegraphics[width=\linewidth]{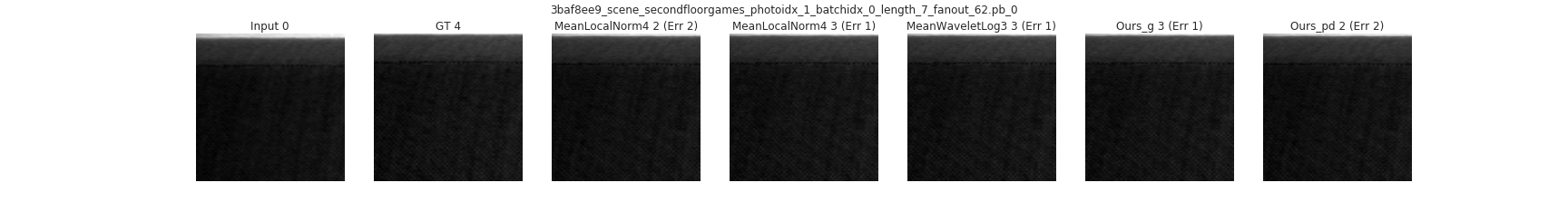}
\vspace{-.4cm}
\includegraphics[width=\linewidth]{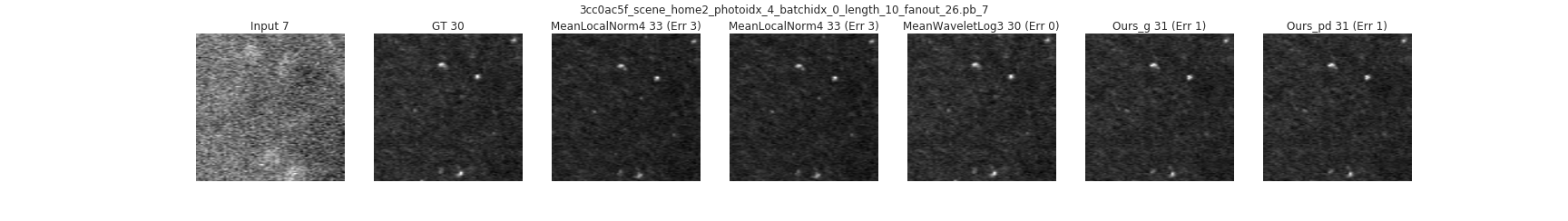}
\vspace{-.4cm}
\includegraphics[width=\linewidth]{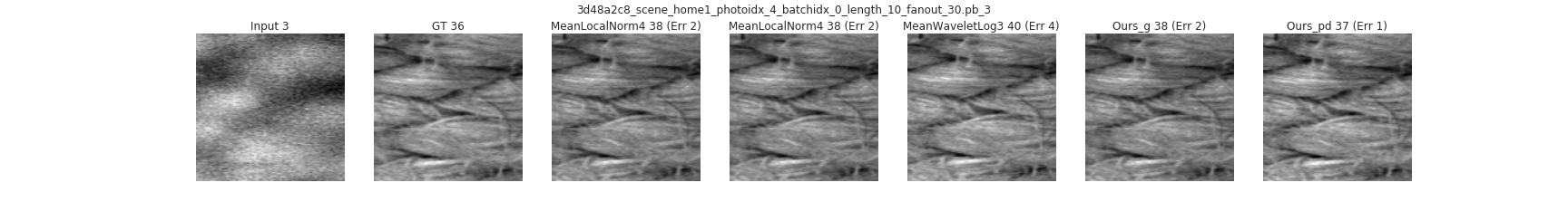}
\vspace{-.4cm}
\includegraphics[width=\linewidth]{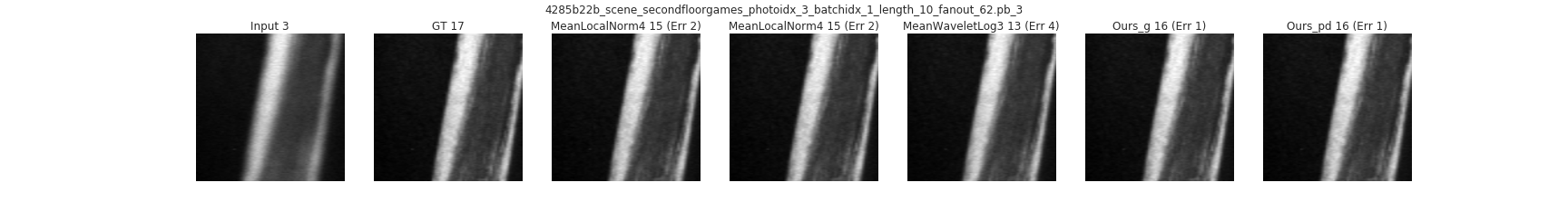}
\vspace{-.4cm}
\includegraphics[width=\linewidth]{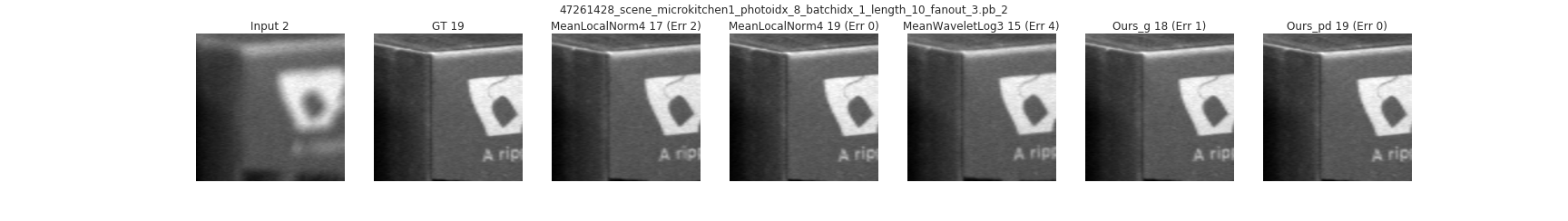}
\vspace{-.4cm}
\includegraphics[width=\linewidth]{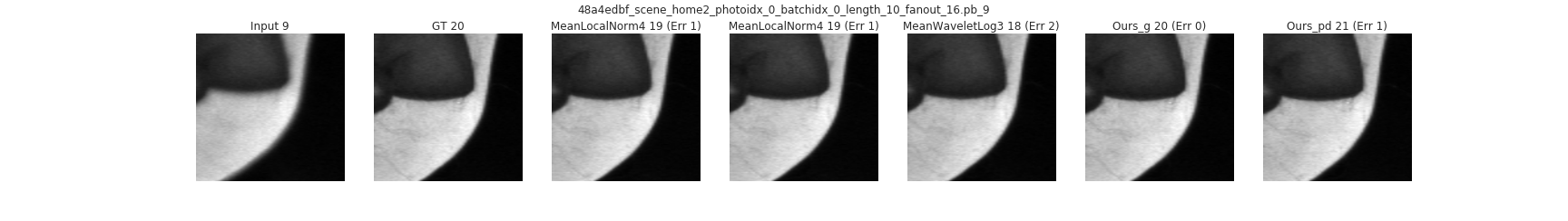}
\vspace{-.4cm}
\includegraphics[width=\linewidth]{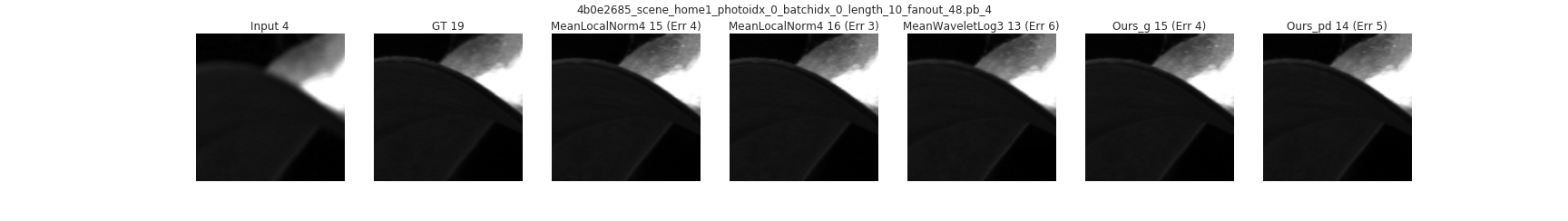}
\vspace{-.4cm}
\includegraphics[width=\linewidth]{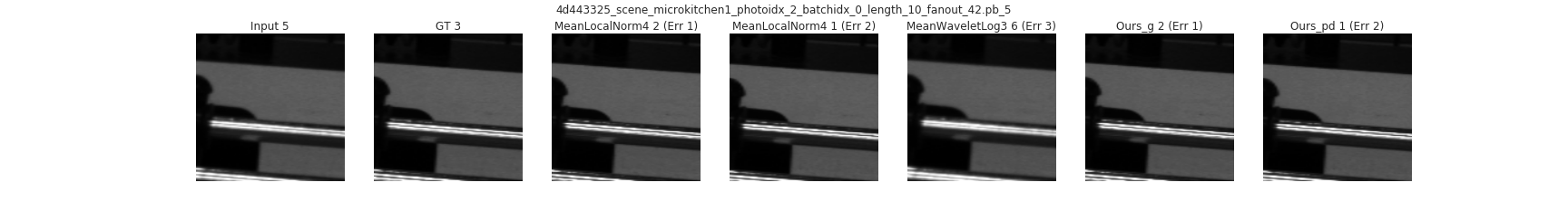}
\caption{Algorithms given fullfocal. Example page 4}\label{fig:full4}
\end{figure*}

In Figures \ref{fig:full1}, \ref{fig:full2}, \ref{fig:full3}, \ref{fig:full4}, we provide a random selection of inputs in the test set. The diagram contains an ``Input'' category; however, this is simply to display another element of the focal stack. All of these algorithms receive the full focal stack as input. The focal stack identification key is directly above the row. The title of each focal slice is: the name of the algorithm, the index, (``Err'' followed by the number of indices away from the ground truth).

\end{document}